\pdfoutput=1
\PassOptionsToPackage{table}{xcolor}
\documentclass[11pt]{article}

\usepackage[final]{acl}

\usepackage{times}
\usepackage{latexsym}

\usepackage[T1]{fontenc}

\usepackage[utf8]{inputenc}

\usepackage{microtype}

\usepackage{inconsolata}

\usepackage{graphicx}

\usepackage{url}
\usepackage{CJKutf8}
\usepackage{booktabs}
\usepackage{hyperref}
\usepackage[table]{xcolor}
\usepackage{bbding}
\usepackage{amsfonts}
\usepackage{amsmath}
\usepackage{multirow}
\usepackage{subcaption}

\title{TaP: A Taxonomy-Guided Framework for Automated and Scalable Preference Data Generation}

\author{
Renren Jin$^{1}$\thanks{Work done during internship at Xiaomi AI Lab.}, Tianhao Shen$^{1}$, Xinwei Wu$^{1}$, Dan Shi$^{1}$, Haoran Sun$^{1}$, Yuqi Ren$^{1}$ \\ \textbf{Wuwei Huang}$^{2}$\textbf{,}
\textbf{Quandong Wang}$^{2}$\textbf{,} \textbf{Wei Liu}$^{2}$\textbf{,} \textbf{Jian Luan}$^{2}$\textbf{,} \textbf{Bin Wang}$^{2}$\textbf{,} \textbf{Deyi Xiong}$^{1}$\thanks{~Corresponding author.}\\
$^1$School of Computer Science and Technology, Tianjin University, Tianjin, China\\
$^{2}$Xiaomi AI Lab, Beijing, China\\
\texttt{\{rrjin, thshen, wuxw2021, shidan, hrsun, ryq20, dyxiong\}@tju.edu.cn}\\
\texttt{\{huangwuwei, wangquandong, liuwei40, luanjian, wangbin11\}@xiaomi.com}\\
}

\begin{document}
\maketitle
\begin{abstract}

Conducting supervised and preference fine-tuning of large language models (LLMs) requires high-quality datasets to improve their ability to follow instructions and align with human preferences and values. However, constructing such datasets is resource-intensive, and most publicly available datasets are in English. To address these challenges, we propose the \underline{\textbf{Ta}}xonomy-Guided \underline{\textbf{P}}reference Data Generation (TaP) framework for automated, scalable preference dataset construction across languages. TaP uses a structured taxonomy to provide fine-grained control over dataset composition, ensuring diversity and broad coverage. We use TaP-generated datasets to perform supervised and preference fine-tuning on multiple LLMs. Experimental results demonstrate that LLMs trained on TaP-generated datasets outperform those trained on existing open-source datasets. Remarkably, LLMs trained on TaP-generated datasets outperform models trained on an open-source dataset that is 180$\times$ larger.

\end{abstract}

\section{Introduction}
Large language models (LLMs) typically undergo three primary training stages: (1) pre-training, during which LLMs are trained on extensive and diverse datasets encompassing multiple languages and modalities \citep{DBLP:journals/corr/abs-2303-18223,DBLP:journals/corr/abs-2412-15115,DBLP:journals/corr/abs-2407-21783,DBLP:journals/corr/abs-2412-19437}; (2) supervised fine-tuning, in which LLMs are fine-tuned using prompt--response pairs authored by humans or LLMs \citep{DBLP:journals/corr/abs-2308-10792}; and (3) preference fine-tuning, wherein preference data annotated by humans or LLMs are used to conduct RLHF or alternative methods such as DPO and KTO \citep{DBLP:conf/nips/Ouyang0JAWMZASR22,DBLP:conf/nips/RafailovSMMEF23,DBLP:journals/corr/abs-2402-01306}. The latter two stages are essential for improving LLMs' ability to follow instructions and generate responses aligned with human preferences and values.

\begin{figure*}[ht]
    \centering
    \includegraphics[width=0.95\textwidth]{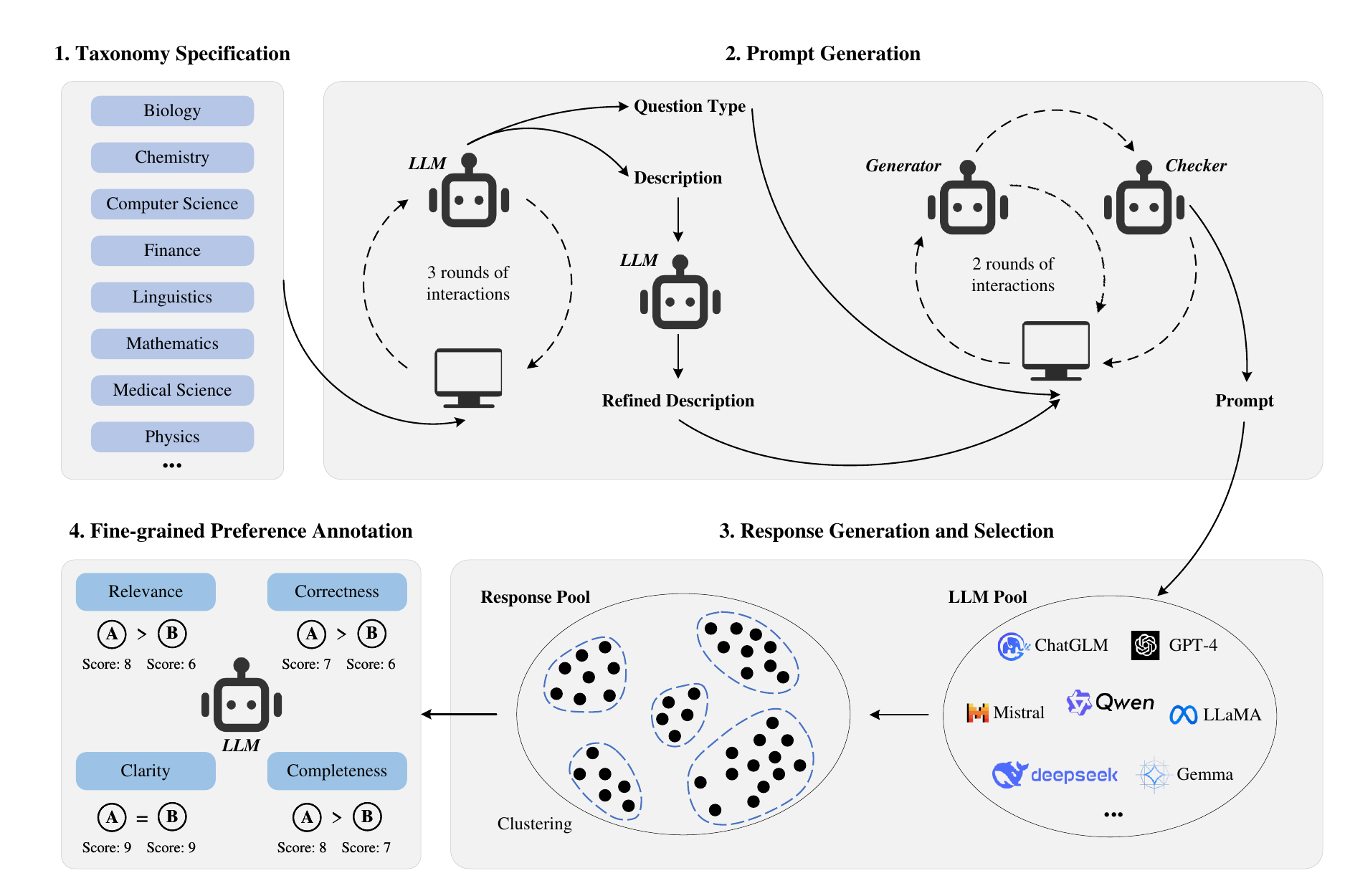}
    \caption{Diagram of the preference dataset construction process using TaP, where the taxonomy can be derived from the undergraduate program catalog.}
    \label{fig:TaP}
\end{figure*}

However, datasets for supervised and preference fine-tuning often require human annotation, which is time-consuming and expensive. These constraints hinder the scalability of dataset expansion, limiting the amount of available training data. A common alternative is to leverage LLMs to generate synthetic data, thereby reducing reliance on human labor and improving scalability. Nevertheless, naive prompting often yields homogeneous data with limited coverage and pronounced task or domain imbalance, where certain domains or tasks are overrepresented while others remain underrepresented \citep{DBLP:conf/acl/WangKMLSKH23,DBLP:journals/corr/abs-2306-11644,DBLP:conf/iclr/XuJNDP0L25}. Such imbalance can cause LLMs to overfit overrepresented domains or tasks, weakening their generalizability in real-world applications. Consequently, ensuring both diversity and comprehensive coverage in synthetic data generation remains a critical challenge.

Furthermore, compared with open-source datasets for supervised fine-tuning, far fewer datasets are available for preference fine-tuning \citep{DBLP:conf/icml/CuiY0YH0NXXL0024,DBLP:conf/naacl/WangDZASEDSKSK24,DBLP:journals/corr/abs-2406-08673}. Additionally, open-source datasets for both supervised and preference fine-tuning are predominantly in English. This language bias limits the global applicability of LLMs and marginalizes non-English-speaking communities, preventing them from fully benefiting from these models. Although translating English datasets into other languages is a potential solution, it can introduce translation errors and produce ``translationese'', thereby compromising dataset quality \citep{DBLP:conf/acl/RileyCFG20}. Given that there are over 7{,}000 languages worldwide, it is imperative to develop novel approaches and release open-source datasets in diverse languages to better support the latter two training stages of LLMs.

To address these challenges, we propose the \underline{\textbf{Ta}}xonomy-Guided \underline{\textbf{P}}reference Data Generation (TaP) framework for automated, scalable preference data generation across languages. TaP leverages a predefined taxonomy to guide LLMs in producing preference data spanning diverse domains and tasks. By generating data according to the taxonomy, TaP provides fine-grained control over dataset composition and facilitates the construction of diverse datasets with broad coverage of the taxonomy categories. The TaP framework comprises four steps: (1) designing a taxonomy aligned with the target application of LLMs (e.g., for general-purpose LLMs, organizing the taxonomy around use cases such as brainstorming, open-ended question answering, and summarization); (2) prompting LLMs to generate prompts for the categories specified in the taxonomy; (3) employing multiple LLMs to generate diverse responses for each prompt; and (4) using LLMs to evaluate and rank responses according to predefined criteria. A diagram of the dataset construction process using the TaP framework is shown in Figure~\ref{fig:TaP}.

Motivated by the fundamental role of knowledge in decision-making, problem-solving, and adaptation to new environments, as well as by empirical findings indicating that LLMs benefit from diverse and knowledge-rich training data \citep{DBLP:journals/corr/abs-2309-05463,lozhkov2024fineweb-edu}, we propose leveraging a comprehensive taxonomy, which encompasses a broad spectrum of knowledge and skills essential to human expertise, to guide the generation of such preference data using LLMs. Through careful manual analysis, we identify undergraduate program catalogs as a valuable resource for constructing such a taxonomy, as they are structured, comprehensive repositories of essential knowledge for human learning. \textbf{Considering that undergraduate education constitutes a foundational component of higher education systems worldwide and is available in most countries,}\footnote{\url{https://en.wikipedia.org/wiki/Undergraduate_education}} \textbf{employing undergraduate program catalogs enhances the adaptability of the TaP framework across various languages}. When such catalogs are unavailable in a given language, we discuss alternative approaches for constructing the taxonomy in Appendix~\ref{sec:appendix_alternative_approaches_for_constructing_the_taxonomy}.

Given the abundance of publicly available preference datasets in English and the relative scarcity of such datasets in other languages, we apply the TaP framework to generate preference data in Chinese. Specifically, we adopt the \textit{Undergraduate Program Catalog of Regular Higher Education Institutions of China}\footnote{\url{http://www.moe.gov.cn/srcsite/A08/moe_1034/s4930/202403/W020240319305498791768.pdf}} as the guiding taxonomy for preference data generation. Manual inspection of the generated data reveals several desirable properties: (1) rich in knowledge, (2) high information density, and (3) broad coverage across domains and tasks. An illustrative example of a prompt and corresponding response is presented in Figure~\ref{fig:illustrative_example} in Appendix~\ref{sec:appendix_an_illustrative_example} to highlight these properties. We conduct both supervised and preference fine-tuning, including PPO and DPO, on LLMs using the TaP-generated data. Experimental results demonstrate that LLMs fine-tuned with TaP-generated data outperform those trained on existing open-source datasets. Our contributions can be summarized as follows:

\begin{itemize}
    \item We propose the TaP framework, which enables automated, scalable preference data generation across languages and provides fine-grained control over dataset composition.

    \item We apply TaP to generate high-quality Chinese preference data for supervised and preference fine-tuning (e.g., PPO, DPO, GRPO). Additionally, the annotations produced by LLMs during data generation can be used to fine-tune open-source LLMs as judges \citep{DBLP:conf/iclr/WangYYZYW0J000024,DBLP:conf/emnlp/KimSLLSWNL0S24,DBLP:conf/iclr/ZhuWW25}, reducing reliance on costly proprietary APIs and the risk of data leakage.

    \item We conduct extensive experiments using TaP-generated preference data to train five open-source LLMs (3B--14B) from three model families: LLaMA-3.1, Qwen2.5, and Gemma-2. The results show that models trained on TaP-generated data outperform those trained on existing open-source datasets.

\end{itemize}

\section{Related Work}

Our research is closely related to studies on synthetic data generation using LLMs. Pioneering studies in this field have leveraged the in-context learning capabilities of LLMs by providing demonstrations to guide the generation of synthetic samples \citep{DBLP:conf/acl/HonovichSLS23,DBLP:conf/acl/WangKMLSKH23}, typically requiring a small set of seed samples. Following this line of research, subsequent work has typically required a small number of seed samples to guide LLMs in producing synthetic samples \citep{DBLP:conf/iclr/XuSZG0FTLJ24,DBLP:journals/corr/abs-2306-02707,DBLP:conf/iclr/LiYZSLZWL24,DBLP:conf/nips/YueZZC24,DBLP:conf/acl/ZhuXW025,DBLP:conf/acl/RenZHLXZSY25}.

In contrast, TaP uses a taxonomy to guide LLMs in autonomously generating synthetic data, eliminating the need for seed samples authored by humans or LLMs. While TaP shares the goal of constructing preference datasets with UltraFeedback \citep{DBLP:conf/icml/CuiY0YH0NXXL0024} and the HelpSteer series \citep{DBLP:conf/naacl/WangDZASEDSKSK24,DBLP:journals/corr/abs-2406-08673}, it differs in two key aspects. First, TaP can produce prompts from scratch without relying on existing instruction datasets, enabling fine-grained control over dataset composition and improving its applicability across languages. By contrast, \citet{DBLP:conf/icml/CuiY0YH0NXXL0024} and \citet{DBLP:conf/naacl/WangDZASEDSKSK24,DBLP:journals/corr/abs-2406-08673} construct preference datasets by collecting prompts from existing instruction datasets. Second, TaP constructs Chinese preference datasets to address the scarcity of such data, whereas \citet{DBLP:conf/icml/CuiY0YH0NXXL0024} and \citet{DBLP:conf/naacl/WangDZASEDSKSK24,DBLP:journals/corr/abs-2406-08673} mainly focus on English.

\section{Preference Dataset Curation}
\label{sec:dataset_construction}
The TaP framework for automated and scalable preference data generation comprises four key steps: (1) \textbf{Taxonomy Specification}, which constructs a structured taxonomy to comprehensively cover common use cases of LLMs; (2) \textbf{Prompt Generation}, which guides LLMs to produce prompts corresponding to the categories defined in the taxonomy; (3) \textbf{Response Generation and Selection}, in which multiple LLMs generate responses to each prompt and a subset is selected to balance annotation cost and response diversity; and (4) \textbf{Fine-grained Preference Annotation}, in which LLMs evaluate and score the selected responses across multiple dimensions. Figure~\ref{fig:TaP} illustrates the entire process of constructing the preference dataset using the TaP framework.

\subsection{Taxonomy Specification}

We adopt the \textit{Undergraduate Program Catalog of Regular Higher Education Institutions of China} as the taxonomy for guiding LLMs to generate Chinese preference data. This catalog lists all undergraduate subjects offered by Chinese universities, spanning a wide range of disciplines across the humanities, social sciences, and STEM fields. It comprises 816 distinct subjects grouped into 93 discipline categories. These categories are further organized into 12 primary degree-conferring disciplines, providing a structured organization of undergraduate knowledge.

To construct preference datasets covering diverse domains and tasks, we incorporate nearly all subjects specified in the catalog into our taxonomy, except those in the Foreign Languages and Literatures discipline, which includes many low-resource languages. Instead, we selectively include 12 subjects: the top nine languages, along with Linguistics, Translation, and Business English. Consequently, our taxonomy contains 724 subjects in total: 12 from Foreign Languages and Literatures and 712 from other disciplines.

\subsection{Prompt Generation}
\label{subsec:prompt_generation}

We adopt a multi-stage prompting strategy, rather than instructing LLMs to generate prompts directly, to improve both the diversity and coverage of the resulting outputs. \textbf{In the first stage}, LLMs are prompted to generate diverse question types and brief descriptions for each subject within a predefined taxonomy through a three-round interaction. Each round produces a distinct set of question types and descriptions, thereby increasing both the quantity and diversity of the generated outputs. The prompt used in this stage is provided in Figure~\ref{fig:question_type_prompt} of Appendix~\ref{subsec:appendix_prompts}. \textbf{In the second stage}, LLMs refine the question-type descriptions from the first stage to improve their clarity and real-world relevance. This refinement aims to improve the quality of the model-generated prompts. The prompt used in this stage is shown in Figure~\ref{fig:description_revise_prompt} of Appendix~\ref{subsec:appendix_prompts}. \textbf{In the third stage}, we employ a two-round interaction to address the issue that LLMs occasionally produce incomplete prompts that lack essential contextual information.\footnote{For instance, LLMs may generate prompts that request a summary of an article without providing the article itself.} In the first round, LLMs generate a candidate prompt based on the question types and refined descriptions. In the second round, they assess whether the prompt contains all necessary information. If a prompt is identified as incomplete, the LLMs regenerate it to ensure completeness. The prompt used for this interaction is illustrated in Figure~\ref{fig:prompt_generation_prompt} of Appendix~\ref{subsec:appendix_prompts}. Additionally, we find that LLMs sometimes generate prompts beyond their capabilities, such as those requiring real-world actions. To mitigate this issue, we employ an LLM as a checker to evaluate whether a given prompt is feasible within its operational constraints. If a prompt is deemed infeasible, it is revised, with up to three regeneration attempts. The evaluation prompt for this step is presented in Figure~\ref{fig:prompt_check_prompt} of Appendix~\ref{subsec:appendix_prompts}. To verify the reliability of LLMs in identifying such infeasible prompts, we manually reviewed a subset of evaluations generated by GPT-4. The results demonstrate that GPT-4 reliably identifies prompts that exceed its capabilities. Further details are provided in Appendix~\ref{subsec:appendix_human_evaluation_on_feasibility_check}.

\begin{table*}[!ht]
  \centering
  \tiny
  \resizebox{\textwidth}{!}{
    \begin{tabular}{lllllcccc}
    \toprule
    \textbf{Dataset} & \textbf{\#Samples} & \textbf{\#Prompts} & \textbf{Prompt Length} & \textbf{Response Length} & \textbf{Fine-grained Annotation?} & \textbf{Annotation Type} & \textbf{Annotator} & \textbf{Languages} \\
    \midrule
    \href{https://github.com/HIT-SCIR/huozi/tree/main/data/huozi-rlhf}{Huozi-RLHF} \citep{huozi} & 16,918   & 3,725   & 69.19 $\pm$ 107.38 & 216.02 $\pm$ 165.44 & \XSolidBrush    & Ranking & Human & Chinese \\
    \href{https://huggingface.co/datasets/wenbopan/Chinese-dpo-pairs}{Chinese-DPO-Pairs} & 10,735   & 10,717 & 220.23 $\pm$ 318.37  & 401.90 $\pm$ 347.63 & \XSolidBrush    & Ranking & AI   & Chinese \\
    \midrule
    \rowcolor[rgb]{ .906,  .902,  .902}
    TaP (GPT-4) & 261,874   & 27,715   & 379.29 $\pm$ 159.11 & 982.89 $\pm$ 919.97 & \Checkmark   & Scalar & AI    & Chinese \\
    \bottomrule
    \end{tabular}%
}
  \caption{Comparison of open-source Chinese preference datasets with the dataset constructed using TaP. ``TaP (GPT-4)'' indicates that the prompts in the dataset were generated by GPT-4. The column labeled ``\#Prompts'' denotes the number of unique prompts in each dataset. ``Prompt Length'' and ``Response Length'' are reported as $XX \pm YY$, where $XX$ is the mean number of tokens and $YY$ is the standard deviation. Tokenization uses the \texttt{cl100k\_base} encoding provided by the tiktoken library.\protect\footnotemark}
  \label{tab:preference_dataset}%
\end{table*}%
\footnotetext{\url{https://github.com/openai/tiktoken}}

\subsection{Response Generation and Selection}

We select 65 widely used LLMs (0.5B--236B parameters) to generate responses for each prompt, increasing response diversity and coverage. The models span dense and MoE architectures and include proprietary and open-source models. The rationale for model selection and detailed model specifications are provided in Appendix~\ref{subsec:appendix_llms_for_response_generation}. Each model uses the system prompt shown in Figure~\ref{fig:response_generation_prompt} in Appendix~\ref{subsec:appendix_prompts}, producing 65 responses per prompt.

Because each prompt yields 65 responses, it produces 2,080 (65 $\times$ 64 / 2) pairwise comparisons for annotation. To reduce annotation cost while maintaining response diversity and coverage, we apply K-means clustering to partition the responses for each prompt into five clusters and select one representative response per cluster. Concretely, we choose the response closest to each cluster centroid, except for the cluster containing the GPT-4 response; in that case, we select the GPT-4 response to ensure that each prompt includes at least one high-quality response. For clustering, we compute response embeddings by mean-pooling token representations from the final layer of Qwen2-7B.
\subsection{Fine-grained Preference Annotation}

\begin{figure*}[!t]
    \centering
    \begin{subfigure}{0.45\textwidth}
        \includegraphics[width=\textwidth]{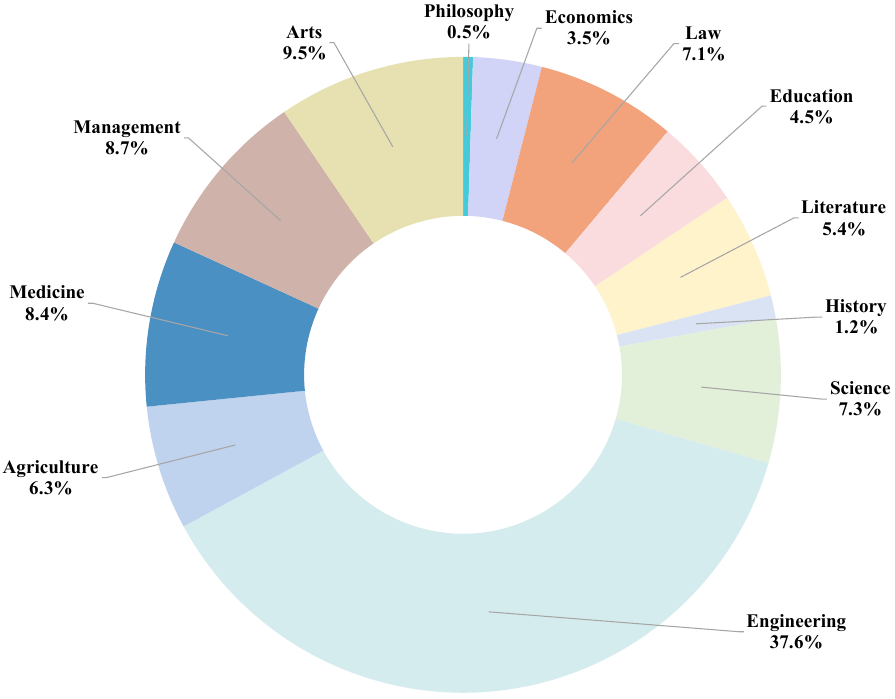}
    \end{subfigure}
    \begin{subfigure}{0.45\textwidth}
        \includegraphics[width=\textwidth]{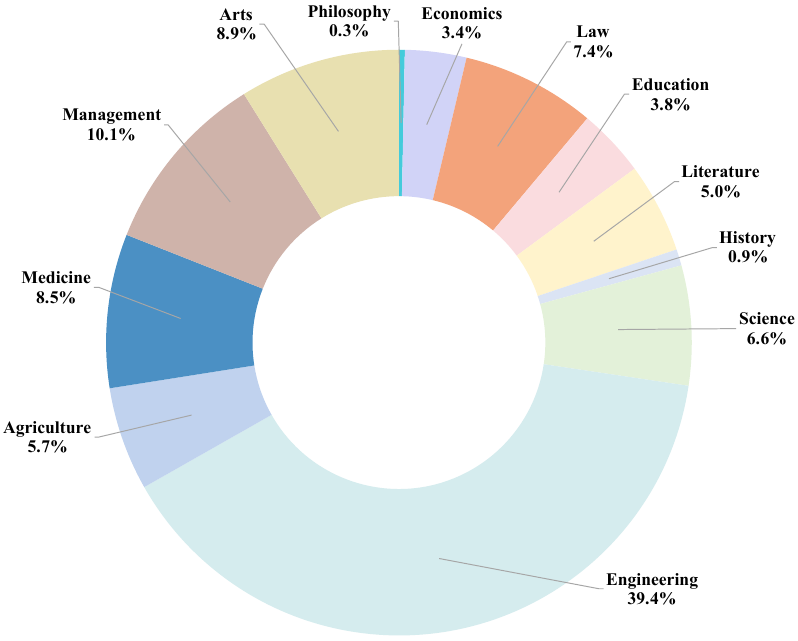}
    \end{subfigure}
    \caption{Distribution of prompts generated by GPT-4 (left) and DeepSeek-V2 (right) across the 12 primary disciplines defined in the taxonomy.}
    \label{fig:distribution}
\end{figure*}

We employ multiple LLMs to perform preference annotations for each pairwise combination of selected responses to a given prompt. Rather than relying on a single model, each response pair is annotated by at least five different LLMs. Specifically, we use GPT-4 together with five high-performing open-source LLMs, including Qwen2-72B-Instruct, Command R+, Mistral-Large-Instruct-2407, Llama-3.1-70B-Instruct, and DeepSeek V2.5, to generate independent preference judgments for all pairwise comparisons.\footnote{Due to the high cost of using GPT-4 for preference annotation, we use it only to annotate response pairs for prompts it generates. In contrast, response pairs for prompts generated by DeepSeek-V2 are annotated by open-source LLMs.}

\begin{table*}[htbp]
  \centering
  \resizebox{\textwidth}{!}{
    \begin{tabular}{lcccccccccc}
    \toprule
          & \multicolumn{1}{c}{\textbf{Human-1}} & \multicolumn{1}{c}{\textbf{Human-2}} & \multicolumn{1}{c}{\textbf{Human-3}} & \multicolumn{1}{c}{\textbf{GPT-4}} & \multicolumn{1}{c}{\textbf{DeepSeek V2.5}} & \multicolumn{1}{c}{\textbf{Command R+}} & \multicolumn{1}{c}{\textbf{Llama-3.1-70B-Instruct}} & \multicolumn{1}{c}{\textbf{Mistral-Large-Instruct-2407}} & \multicolumn{1}{c}{\textbf{Qwen2-72B-Instruct}} & \multicolumn{1}{c}{\textbf{Average}} \\
    \midrule
    \textbf{Human-1} & - & 78.67\% & 76.67\% & 78.93\% & 79.67\% & 77.63\% & 78.93\% & 78.19\% & 74.25\% & 77.87\% \\
    \arrayrulecolor{lightgray}\midrule
    \textbf{Human-2} & 78.67\% &  - & 80.00\% & 88.63\% & 90.33\% & 88.14\% & 88.96\% & 85.91\% & 82.94\% & 85.45\% \\
    \arrayrulecolor{lightgray}\midrule
    \textbf{Human-3} & 76.67\% & 80.00\% &  - & 83.61\% & 82.33\% & 84.07\% & 83.61\% & 84.90\% & 80.94\% & 82.02\% \\
    \arrayrulecolor{lightgray}\midrule
    \textbf{GPT-4} & 78.93\% & 88.63\% & 83.61\% & - & 93.65\% & \textbf{95.25\%} & 92.62\% & \textbf{93.94\%} & 91.61\% & \textbf{89.78\%} \\
    \arrayrulecolor{lightgray}\midrule
    \textbf{DeepSeek V2.5} & \textbf{79.67\%} & \textbf{90.33\%} & 82.33\% & 93.65\% & - & 93.56\% & \textbf{95.99\%} & 92.28\% & 89.63\% & 89.68\% \\
    \arrayrulecolor{lightgray}\midrule
    \textbf{Command R+} & 77.63\% & 88.14\% & 84.07\% & \textbf{95.25\%} & 93.56\% & - & 92.54\% & 93.17\% & \textbf{91.84\%} & 89.52\% \\
    \arrayrulecolor{lightgray}\midrule
    \textbf{Llama-3.1-70B-Instruct} & 78.93\% & 88.96\% & 83.61\% & 92.62\% & \textbf{95.99\%} & 92.54\% & - & 92.93\% & 89.60\% & 89.40\% \\
    \arrayrulecolor{lightgray}\midrule
    \textbf{Mistral-Large-Instruct-2407} & 78.19\% & 85.91\% & \textbf{84.90\%} & 93.94\% & 92.28\% & 93.17\% & 92.93\% & - & 91.58\% & 89.11\% \\
    \arrayrulecolor{lightgray}\midrule
    \textbf{Qwen2-72B-Instruct} & 74.25\% & 82.94\% & 80.94\% & 91.61\% & 89.63\% & 91.84\% & 89.60\% & 91.58\% & - & 86.55\% \\
    \arrayrulecolor{black}\bottomrule
    \end{tabular}%
    }
    \caption{Pairwise preference annotation agreement across annotators. Each cell reports the agreement score between a pair of annotators, and the highest agreement value in each column is highlighted in bold.}
  \label{tab:pairwise_annotation_agreement}%
\end{table*}%

For preference annotation, we adopt a pairwise comparison approach, wherein LLMs choose the preferred response between two alternatives, rather than assigning independent scores to each response as in pointwise grading. Our preliminary experiments indicate that pointwise grading often fails to distinguish subtle differences between similar responses, frequently assigning them identical scores. Despite its higher annotation cost, we employ pairwise comparison to obtain more precise preference annotations. In this approach, we present two responses to LLMs and ask them to score each one across four dimensions: relevance, correctness, clarity, and completeness. These dimensions facilitate a comprehensive assessment of response quality. The resulting scores can be used to train fine-grained reward models, which in turn enhance control over LLM-generated responses by assigning weights to rewards from different fine-grained reward models during RLHF \citep{DBLP:conf/nips/WuHSDSASOH23}. The prompt used for pairwise comparison is provided in Figure~\ref{fig:response_eval_prompt} of Appendix~\ref{subsec:appendix_prompts}. Additionally, to mitigate position bias in pairwise comparison, where LLMs tend to assign higher scores to the response appearing in a particular position (e.g., the first or second slot), we adopt the Balanced Position Calibration strategy \citep{DBLP:conf/acl/WangLCCZLCKLLS24}, which involves conducting an additional comparison with the positions of the two responses swapped.

To derive preference labels from pairwise comparison scores, we compute an overall score for each response and label the higher-scoring response as preferred. Specifically, we average the score for each evaluation dimension across both positions (original vs.\ swapped) to obtain calibrated scores. We then assign equal weights to the calibrated dimension scores and average them to produce a final score for each response. Although domain-specific applications may prioritize certain dimensions (e.g., correctness in medical applications), we use uniform weights across all four dimensions to support general-purpose LLM training.

\section{Dataset Statistics}
\label{sec:dataset_statistics}

We employ GPT-4 and DeepSeek-V2 to generate prompts for the 724 subjects in our taxonomy, derived from the \textit{Undergraduate Program Catalog of Regular Higher Education Institutions of China}. GPT-4 generates 27,715 prompts, whereas DeepSeek-V2 produces 50,929 prompts. The distributions of these prompts across the 12 primary disciplines are presented in Figure~\ref{fig:distribution}. As shown in Figure~\ref{fig:distribution}, despite using different LLMs for prompt generation, the resulting distributions are highly consistent, indicating that TaP enables fine-grained control over dataset composition. In addition, because some prompts must be regenerated after failing the feasibility check, a single question type may correspond to multiple prompts. Consequently, GPT-4 yields 19,840 distinct question types, and DeepSeek-V2 yields 24,839 distinct question types.

For constructing the preference dataset, we rely solely on preference annotations generated by GPT-4, given its pioneering performance.\footnote{While incorporating annotations from multiple LLMs might improve dataset quality, we leave the integration of annotations from other open-source LLMs for future work.} Table~\ref{tab:preference_dataset} presents a comparative summary of the preference dataset constructed by TaP and existing open-source preference datasets. Compared with these datasets, the TaP-constructed dataset contains more samples and prompts, includes longer prompts and responses, and provides fine-grained annotations.

\section{Annotation Agreement}
To assess the reliability of preference annotations produced by LLMs, we examine three aspects of annotation agreement: (i) agreement among human annotators, (ii) agreement between human annotators and LLMs, and (iii) agreement across different LLMs. To this end, we randomly sample 300 prompts, their paired responses generated by two distinct LLMs, and the corresponding model-generated preference annotations. We then recruit three PhD students, all native Chinese speakers with research interests in computer science, to independently label the preferred response for each response pair. Label Studio\footnote{\url{https://github.com/HumanSignal/label-studio}} is employed to provide an annotation interface that facilitates the process. For each prompt, annotators are shown the prompt and the two LLM-generated responses (Response A and Response B) and are instructed to indicate their preference by selecting one of three options: ``Response A Win,'' ``Response B Win,'' or ``Tie.''

\begin{table}[!t]
  \centering
  \tiny
  \resizebox{0.5\textwidth}{!}{
    \begin{tabular}{lrl}
    \toprule
    \textbf{Dataset} & \textbf{\#Samples} & \textbf{Languages} \\
    \midrule
    \href{https://huggingface.co/datasets/YeungNLP/firefly-train-1.1M}{Firefly} & 1,649,399 & Chinese \\
    \href{https://github.com/Instruction-Tuning-with-GPT-4/GPT-4-LLM/blob/main/data/alpaca_gpt4_data_zh.json}{Alpaca-GPT-4-ZH} \citep{DBLP:journals/corr/abs-2304-03277} & 48,818 & Chinese \\
    \href{https://huggingface.co/datasets/BAAI/COIG}{COIG} \citep{DBLP:journals/corr/abs-2304-07987} & 178,246 & Chinese \\
    \href{https://huggingface.co/datasets/fnlp/moss-003-sft-data}{MOSS-SFT} \citep{MIR-2023-12-294} & 1,074,551 & English, Chinese \\
    \href{https://huggingface.co/datasets/m-a-p/COIG-CQIA}{COIG-CQIA} \citep{DBLP:journals/corr/abs-2403-18058} & 44,694 & Chinese \\
    \href{https://huggingface.co/datasets/BAAI/Infinity-Instruct}{Infinity-Instruct} & 757,938 & Chinese \\
    \href{https://huggingface.co/datasets/BelleGroup/train_3.5M_CN}{BELLE-SFT} \citep{BELLE} & 3,606,402 & Chinese \\
    \midrule
    \rowcolor[rgb]{ .906,  .902,  .902}
    TaP-SFT (GPT-4) & 19,840 & Chinese \\
    \rowcolor[rgb]{ .906,  .902,  .902}
    TaP-SFT (DeepSeek-V2) & 24,839 & Chinese \\
    \bottomrule
    \end{tabular}%
}
  \caption{Comparison of open-source datasets containing a substantial number of Chinese samples for supervised fine-tuning with datasets derived from TaP-constructed preference datasets. ``TaP-SFT (GPT-4)'' and ``TaP-SFT (DeepSeek-V2)'' indicate that the prompts were generated by GPT-4 and DeepSeek-V2, respectively. For the multilingual Infinity-Instruct dataset, Chinese samples were selected based on language tags in the data.}
  \label{tab:sft_dataset}%
\end{table}%

Table~\ref{tab:pairwise_annotation_agreement} reports pairwise annotation agreement among different annotators on 300 randomly selected samples. Agreement among human annotators approaches 80\%, exceeding the 60\%--75\% range commonly reported in prior work \citep{DBLP:journals/corr/abs-2204-05862,DBLP:conf/nips/Ouyang0JAWMZASR22,DBLP:conf/icml/CuiY0YH0NXXL0024}. Furthermore, agreement between human annotators and LLM annotators is close to, and in some cases exceeds, 80\%, reaching 90.33\%. This suggests that preference annotations generated by LLMs reliably reflect human preferences. Notably, agreement among LLM annotators consistently surpasses 90\%, substantially outperforming the level of agreement observed among human annotators.

\begin{table*}[!ht]
  \centering
  \tiny
    \begin{tabular}{c|l|ccc|ccc|cc}
    \toprule
    \multirow{3}[6]{*}{\textbf{Model}} & \multicolumn{1}{c|}{\multirow{3}[6]{*}{\textbf{Dataset}}} & \multicolumn{6}{c|}{\textbf{AlignBench}}      & \multicolumn{2}{c}{\textbf{MT-Bench-zh}} \\
\cmidrule{3-10}          &       & \multicolumn{3}{c|}{\textbf{GPT-4o}} & \multicolumn{3}{c|}{\textbf{DeepSeek-V3}} & \multicolumn{1}{c}{\multirow{2}[4]{*}{\textbf{GPT-4o}}} & \multicolumn{1}{c}{\multirow{2}[4]{*}{\textbf{DeepSeek-V3}}} \\
\cmidrule{3-8}          &       & \multicolumn{1}{c}{\textbf{Reasoning}} & \multicolumn{1}{c}{\textbf{Language}} & \multicolumn{1}{c|}{\textbf{Overall}} & \multicolumn{1}{c}{\textbf{Reasoning}} & \multicolumn{1}{c}{\textbf{Language}} & \multicolumn{1}{c|}{\textbf{Overall}} &       &  \\
    \midrule
    \multirow{9}[2]{*}{Qwen2.5-3B} & \href{https://huggingface.co/datasets/YeungNLP/firefly-train-1.1M}{Firefly} & 3.22  & 4.34  & 3.78  & 3.38  & 4.37  & 3.87  & 4.41  & 4.45  \\
          & \href{https://github.com/Instruction-Tuning-with-GPT-4/GPT-4-LLM/blob/main/data/alpaca_gpt4_data_zh.json}{Alpaca-GPT-4-ZH} \citep{DBLP:journals/corr/abs-2304-03277} & 4.16  & 4.87  & 4.51  & 3.99  & 4.75  & 4.37  & 4.83  & 4.68  \\
          & \href{https://huggingface.co/datasets/BAAI/COIG}{COIG} \citep{DBLP:journals/corr/abs-2304-07987}  & 2.73  & 4.12  & 3.42  & 2.84  & 4.03  & 3.43  & 3.35  & 3.44  \\
          & \href{https://huggingface.co/datasets/fnlp/moss-003-sft-data}{MOSS-SFT} \citep{MIR-2023-12-294} & 3.82  & 4.96  & 4.39  & 3.73  & 4.82  & 4.27  & 4.66  & 4.38  \\
          & \href{https://huggingface.co/datasets/m-a-p/COIG-CQIA}{COIG-CQIA} \citep{DBLP:journals/corr/abs-2403-18058} & 3.61  & 4.38  & 3.99  & 3.65  & 4.31  & 3.98  & 4.47  & 4.35  \\
          & \href{https://huggingface.co/datasets/BAAI/Infinity-Instruct}{Infinity-Instruct} & 4.02  & 5.11  & 4.57  & 3.91  & 4.92  & 4.41  & 4.89  & 4.68  \\
          & \href{https://huggingface.co/datasets/BelleGroup/train_3.5M_CN}{BELLE-SFT} \citep{BELLE} & 3.88  & 5.11  & 4.49  & 3.82  & 4.99  & 4.41  & 4.65  & 4.49  \\
          & \cellcolor[rgb]{ .906,  .902,  .902}TaP-SFT (GPT-4) & \cellcolor[rgb]{ .906,  .902,  .902}4.99  & \cellcolor[rgb]{ .906,  .902,  .902}\textbf{5.85} & \cellcolor[rgb]{ .906,  .902,  .902}5.42  & \cellcolor[rgb]{ .906,  .902,  .902}4.57  & \cellcolor[rgb]{ .906,  .902,  .902}\textbf{5.46} & \cellcolor[rgb]{ .906,  .902,  .902}5.02  & \cellcolor[rgb]{ .906,  .902,  .902}\textbf{5.79} & \cellcolor[rgb]{ .906,  .902,  .902}5.53  \\
          & \cellcolor[rgb]{ .906,  .902,  .902}TaP-SFT (DeepSeek-V2) & \cellcolor[rgb]{ .906,  .902,  .902}\textbf{5.25} & \cellcolor[rgb]{ .906,  .902,  .902}5.65  & \cellcolor[rgb]{ .906,  .902,  .902}\textbf{5.45} & \cellcolor[rgb]{ .906,  .902,  .902}\textbf{4.89} & \cellcolor[rgb]{ .906,  .902,  .902}5.39  & \cellcolor[rgb]{ .906,  .902,  .902}\textbf{5.14} & \cellcolor[rgb]{ .906,  .902,  .902}5.67  & \cellcolor[rgb]{ .906,  .902,  .902}\textbf{5.60} \\
    \midrule
    \multirow{9}[2]{*}{Qwen2.5-7B} & \href{https://huggingface.co/datasets/YeungNLP/firefly-train-1.1M}{Firefly} & 3.45  & 4.61  & 4.03  & 3.62  & 4.64  & 4.13  & 4.43  & 4.47  \\
          & \href{https://github.com/Instruction-Tuning-with-GPT-4/GPT-4-LLM/blob/main/data/alpaca_gpt4_data_zh.json}{Alpaca-GPT-4-ZH} \citep{DBLP:journals/corr/abs-2304-03277} & 4.90  & 5.31  & 5.10  & 4.53  & 5.14  & 4.83  & 5.02  & 4.89  \\
          & \href{https://huggingface.co/datasets/BAAI/COIG}{COIG} \citep{DBLP:journals/corr/abs-2304-07987}  & 2.82  & 4.41  & 3.62  & 3.11  & 4.38  & 3.75  & 3.48  & 3.66  \\
          & \href{https://huggingface.co/datasets/fnlp/moss-003-sft-data}{MOSS-SFT} \citep{MIR-2023-12-294} & 4.71  & 5.26  & 4.98  & 4.55  & 5.17  & 4.86  & 5.06  & 4.90  \\
          & \href{https://huggingface.co/datasets/m-a-p/COIG-CQIA}{COIG-CQIA} \citep{DBLP:journals/corr/abs-2403-18058} & 4.72  & 5.13  & 4.92  & 4.51  & 5.02  & 4.76  & 4.73  & 4.51  \\
          & \href{https://huggingface.co/datasets/BAAI/Infinity-Instruct}{Infinity-Instruct} & 4.92  & 5.63  & 5.28  & 4.70  & 5.54  & 5.12  & 4.96  & 4.95  \\
          & \href{https://huggingface.co/datasets/BelleGroup/train_3.5M_CN}{BELLE-SFT} \citep{BELLE} & 4.32  & 5.44  & 4.88  & 4.16  & 5.35  & 4.76  & 4.93  & 4.83  \\
          & \cellcolor[rgb]{ .906,  .902,  .902}TaP-SFT (GPT-4) & \cellcolor[rgb]{ .906,  .902,  .902}\textbf{6.62} & \cellcolor[rgb]{ .906,  .902,  .902}\textbf{6.27} & \cellcolor[rgb]{ .906,  .902,  .902}\textbf{6.45} & \cellcolor[rgb]{ .906,  .902,  .902}\textbf{6.10} & \cellcolor[rgb]{ .906,  .902,  .902}5.83  & \cellcolor[rgb]{ .906,  .902,  .902}\textbf{5.96} & \cellcolor[rgb]{ .906,  .902,  .902}\textbf{6.16} & \cellcolor[rgb]{ .906,  .902,  .902}\textbf{5.99} \\
          & \cellcolor[rgb]{ .906,  .902,  .902}TaP-SFT (DeepSeek-V2) & \cellcolor[rgb]{ .906,  .902,  .902}6.25  & \cellcolor[rgb]{ .906,  .902,  .902}6.07  & \cellcolor[rgb]{ .906,  .902,  .902}6.16  & \cellcolor[rgb]{ .906,  .902,  .902}5.77  & \cellcolor[rgb]{ .906,  .902,  .902}\textbf{5.93} & \cellcolor[rgb]{ .906,  .902,  .902}5.85  & \cellcolor[rgb]{ .906,  .902,  .902}6.10  & \cellcolor[rgb]{ .906,  .902,  .902}5.97  \\
    \midrule
    \multirow{9}[2]{*}{Llama-3.1-8B} & \href{https://huggingface.co/datasets/YeungNLP/firefly-train-1.1M}{Firefly} & 2.40  & 3.97  & 3.19  & 2.40  & 4.07  & 3.24  & 3.72  & 3.69  \\
          & \href{https://github.com/Instruction-Tuning-with-GPT-4/GPT-4-LLM/blob/main/data/alpaca_gpt4_data_zh.json}{Alpaca-GPT-4-ZH} \citep{DBLP:journals/corr/abs-2304-03277} & 2.77  & 4.28  & 3.53  & 2.64  & 4.12  & 3.38  & 4.21  & 4.28  \\
          & \href{https://huggingface.co/datasets/BAAI/COIG}{COIG} \citep{DBLP:journals/corr/abs-2304-07987}  & 1.79  & 2.98  & 2.38  & 1.88  & 3.04  & 2.46  & 2.68  & 2.74  \\
          & \href{https://huggingface.co/datasets/fnlp/moss-003-sft-data}{MOSS-SFT} \citep{MIR-2023-12-294} & 2.84  & 4.41  & 3.63  & 2.69  & 4.32  & 3.50  & 4.21  & 4.06  \\
          & \href{https://huggingface.co/datasets/m-a-p/COIG-CQIA}{COIG-CQIA} \citep{DBLP:journals/corr/abs-2403-18058} & 2.54  & 3.57  & 3.06  & 2.39  & 3.56  & 2.97  & 3.32  & 3.34  \\
          & \href{https://huggingface.co/datasets/BAAI/Infinity-Instruct}{Infinity-Instruct} & 3.15  & 5.07  & 4.11  & 3.01  & \textbf{4.95} & 3.98  & 4.71  & 4.60  \\
          & \href{https://huggingface.co/datasets/BelleGroup/train_3.5M_CN}{BELLE-SFT} \citep{BELLE} & 3.47  & 5.05  & 4.26  & 3.24  & 4.88  & 4.06  & 4.63  & 4.74  \\
          & \cellcolor[rgb]{ .906,  .902,  .902}TaP-SFT (GPT-4) & \cellcolor[rgb]{ .906,  .902,  .902}\textbf{3.87} & \cellcolor[rgb]{ .906,  .902,  .902}\textbf{5.14} & \cellcolor[rgb]{ .906,  .902,  .902}\textbf{4.50} & \cellcolor[rgb]{ .906,  .902,  .902}3.40  & \cellcolor[rgb]{ .906,  .902,  .902}4.73  & \cellcolor[rgb]{ .906,  .902,  .902}\textbf{4.06} & \cellcolor[rgb]{ .906,  .902,  .902}\textbf{5.24} & \cellcolor[rgb]{ .906,  .902,  .902}\textbf{4.95} \\
          & \cellcolor[rgb]{ .906,  .902,  .902}TaP-SFT (DeepSeek-V2) & \cellcolor[rgb]{ .906,  .902,  .902}3.61  & \cellcolor[rgb]{ .906,  .902,  .902}4.75  & \cellcolor[rgb]{ .906,  .902,  .902}4.18  & \cellcolor[rgb]{ .906,  .902,  .902}\textbf{3.47} & \cellcolor[rgb]{ .906,  .902,  .902}4.61  & \cellcolor[rgb]{ .906,  .902,  .902}4.04  & \cellcolor[rgb]{ .906,  .902,  .902}5.11  & \cellcolor[rgb]{ .906,  .902,  .902}4.89  \\
    \midrule
    \multirow{9}[2]{*}{Gemma-2-9B} & \href{https://huggingface.co/datasets/YeungNLP/firefly-train-1.1M}{Firefly} & 2.36  & 3.95  & 3.15  & 2.44  & 4.06  & 3.25  & 3.92  & 3.77  \\
          & \href{https://github.com/Instruction-Tuning-with-GPT-4/GPT-4-LLM/blob/main/data/alpaca_gpt4_data_zh.json}{Alpaca-GPT-4-ZH} \citep{DBLP:journals/corr/abs-2304-03277} & 3.17  & 4.22  & 3.70  & 3.19  & 4.17  & 3.68  & 4.39  & 4.09  \\
          & \href{https://huggingface.co/datasets/BAAI/COIG}{COIG} \citep{DBLP:journals/corr/abs-2304-07987}  & 1.70  & 2.81  & 2.25  & 1.95  & 2.80  & 2.38  & 2.69  & 2.65  \\
          & \href{https://huggingface.co/datasets/fnlp/moss-003-sft-data}{MOSS-SFT} \citep{MIR-2023-12-294} & 2.84  & 4.21  & 3.53  & 2.73  & 4.19  & 3.46  & 4.14  & 4.23  \\
          & \href{https://huggingface.co/datasets/m-a-p/COIG-CQIA}{COIG-CQIA} \citep{DBLP:journals/corr/abs-2403-18058} & 2.69  & 3.58  & 3.14  & 2.64  & 3.56  & 3.10  & 3.36  & 3.13  \\
          & \href{https://huggingface.co/datasets/BAAI/Infinity-Instruct}{Infinity-Instruct} & 3.72  & \textbf{5.21} & 4.47  & 3.57  & 4.99  & 4.28  & 4.78  & 4.45  \\
          & \href{https://huggingface.co/datasets/BelleGroup/train_3.5M_CN}{BELLE-SFT} \citep{BELLE} & 3.72  & 5.16  & 4.44  & \textbf{3.68} & \textbf{5.07} & \textbf{4.37} & 4.69  & 4.87  \\
          & \cellcolor[rgb]{ .906,  .902,  .902}TaP-SFT (GPT-4) & \cellcolor[rgb]{ .906,  .902,  .902}\textbf{4.10} & \cellcolor[rgb]{ .906,  .902,  .902}5.17  & \cellcolor[rgb]{ .906,  .902,  .902}\textbf{4.64} & \cellcolor[rgb]{ .906,  .902,  .902}3.60  & \cellcolor[rgb]{ .906,  .902,  .902}4.88  & \cellcolor[rgb]{ .906,  .902,  .902}4.24  & \cellcolor[rgb]{ .906,  .902,  .902}\textbf{5.37} & \cellcolor[rgb]{ .906,  .902,  .902}\textbf{5.03} \\
          & \cellcolor[rgb]{ .906,  .902,  .902}TaP-SFT (DeepSeek-V2) & \cellcolor[rgb]{ .906,  .902,  .902}3.96  & \cellcolor[rgb]{ .906,  .902,  .902}4.90  & \cellcolor[rgb]{ .906,  .902,  .902}4.43  & \cellcolor[rgb]{ .906,  .902,  .902}3.61  & \cellcolor[rgb]{ .906,  .902,  .902}4.58  & \cellcolor[rgb]{ .906,  .902,  .902}4.10  & \cellcolor[rgb]{ .906,  .902,  .902}4.90  & \cellcolor[rgb]{ .906,  .902,  .902}4.69  \\
    \midrule
    \multirow{9}[2]{*}{Qwen2.5-14B} & \href{https://huggingface.co/datasets/YeungNLP/firefly-train-1.1M}{Firefly} & 3.94  & 4.94  & 4.44  & 4.10  & 4.89  & 4.49  & 4.77  & 4.79  \\
          & \href{https://github.com/Instruction-Tuning-with-GPT-4/GPT-4-LLM/blob/main/data/alpaca_gpt4_data_zh.json}{Alpaca-GPT-4-ZH} \citep{DBLP:journals/corr/abs-2304-03277} & 5.64  & 5.73  & 5.68  & 5.34  & 5.73  & 5.54  & 5.44  & 5.36  \\
          & \href{https://huggingface.co/datasets/BAAI/COIG}{COIG} \citep{DBLP:journals/corr/abs-2304-07987}  & 3.30  & 4.84  & 4.07  & 3.38  & 4.86  & 4.12  & 3.63  & 3.56  \\
          & \href{https://huggingface.co/datasets/fnlp/moss-003-sft-data}{MOSS-SFT} \citep{MIR-2023-12-294} & 5.25  & 5.59  & 5.42  & 4.96  & 5.48  & 5.22  & 5.26  & 5.04  \\
          & \href{https://huggingface.co/datasets/m-a-p/COIG-CQIA}{COIG-CQIA} \citep{DBLP:journals/corr/abs-2403-18058} & 5.14  & 5.27  & 5.20  & 4.94  & 5.26  & 5.10  & 4.92  & 4.78  \\
          & \href{https://huggingface.co/datasets/BAAI/Infinity-Instruct}{Infinity-Instruct} & 5.26  & 5.94  & 5.60  & 5.16  & 5.81  & 5.48  & 5.33  & 5.06  \\
          & \href{https://huggingface.co/datasets/BelleGroup/train_3.5M_CN}{BELLE-SFT} \citep{BELLE} & 4.70  & 5.70  & 5.20  & 4.55  & 5.61  & 5.08  & 4.99  & 4.83  \\
          & \cellcolor[rgb]{ .906,  .902,  .902}TaP-SFT (GPT-4) & \cellcolor[rgb]{ .906,  .902,  .902}\textbf{6.76} & \cellcolor[rgb]{ .906,  .902,  .902}\textbf{6.51} & \cellcolor[rgb]{ .906,  .902,  .902}\textbf{6.64} & \cellcolor[rgb]{ .906,  .902,  .902}6.27  & \cellcolor[rgb]{ .906,  .902,  .902}\textbf{6.27} & \cellcolor[rgb]{ .906,  .902,  .902}6.27  & \cellcolor[rgb]{ .906,  .902,  .902}\textbf{6.34} & \cellcolor[rgb]{ .906,  .902,  .902}6.18  \\
          & \cellcolor[rgb]{ .906,  .902,  .902}TaP-SFT (DeepSeek-V2) & \cellcolor[rgb]{ .906,  .902,  .902}6.34  & \cellcolor[rgb]{ .906,  .902,  .902}5.92  & \cellcolor[rgb]{ .906,  .902,  .902}6.13  & \cellcolor[rgb]{ .906,  .902,  .902}\textbf{6.46} & \cellcolor[rgb]{ .906,  .902,  .902}6.24  & \cellcolor[rgb]{ .906,  .902,  .902}\textbf{6.35} & \cellcolor[rgb]{ .906,  .902,  .902}6.12  & \cellcolor[rgb]{ .906,  .902,  .902}\textbf{6.19} \\
    \bottomrule
    \end{tabular}%
  \caption{Performance of five LLMs on AlignBench and MT-Bench-zh after supervised fine-tuning on different datasets. ``GPT-4o'' and ``DeepSeek-V3'' indicate that responses were evaluated using GPT-4o and DeepSeek-V3.}
  \label{tab:sft_all}
\end{table*}%

When computing pairwise preference annotation agreement, we observe a small number of exceptional cases in which LLMs fail to strictly follow the annotation instructions and therefore do not produce a valid preference score, which prevents extraction of the corresponding preference label. We exclude such samples from the agreement computation. Importantly, these cases are extremely rare: among the 300 randomly selected samples, at most seven LLM-generated annotations fail to yield a valid preference label. Consequently, excluding these samples has a negligible impact on the overall results and does not affect the validity of our experimental conclusions.

\section{Experiments}

To assess TaP’s effectiveness and the quality of the preference datasets it constructs, we conducted supervised fine-tuning and preference fine-tuning (PPO/DPO) on LLMs using TaP-generated data.

\subsection{Supervised Fine-tuning}

To conduct supervised fine-tuning using the preference data generated by TaP, we utilized the prompts from the dataset along with GPT-4-generated responses. To ensure a balanced distribution of question types, we randomly selected one prompt and its corresponding response for each question type with multiple prompts. The resulting dataset comprises 19,840 samples generated by GPT-4 and 24,839 samples produced by DeepSeek-V2. A comparative summary of this supervised fine-tuning dataset, alongside other open-source Chinese datasets, is provided in Table~\ref{tab:sft_dataset}.

We conducted supervised fine-tuning on five open-source LLMs: (1) Qwen2.5-3B, (2) Qwen2.5-7B, (3) Llama-3.1-8B, (4) Gemma-2-9B, and (5) Qwen2.5-14B. These LLMs span three model families, with parameter sizes ranging from 3B to 14B.

\subsection{Preference Fine-tuning}

We adopt two dominant approaches for preference fine-tuning using the dataset constructed by TaP: PPO and DPO. For comparison with other open-source preference datasets, we utilize Huozi-RLHF \citep{huozi} and Chinese-DPO-Pairs\footnote{\url{https://huggingface.co/datasets/wenbopan/Chinese-dpo-pairs}} for preference fine-tuning. Further details on the PPO and DPO setups are provided in Appendix~\ref{subsec:appendix_preference_finetuning}.

\begin{table*}[!ht]
  \centering
  \tiny
    \begin{tabular}{c|l|ccc|ccc|cc}
    \toprule
    \multirow{3}[6]{*}{\textbf{Model}} & \multicolumn{1}{c|}{\multirow{3}[6]{*}{\textbf{Dataset}}} & \multicolumn{6}{c|}{\textbf{AlignBench}}      & \multicolumn{2}{c}{\textbf{MT-Bench-zh}} \\
\cmidrule{3-10}          &       & \multicolumn{3}{c|}{\textbf{GPT-4o}} & \multicolumn{3}{c|}{\textbf{DeepSeek-V3}} & \multicolumn{1}{c}{\multirow{2}[4]{*}{\textbf{GPT-4o}}} & \multicolumn{1}{c}{\multirow{2}[4]{*}{\textbf{DeepSeek-V3}}} \\
\cmidrule{3-8}          &       & \multicolumn{1}{c}{\textbf{Reasoning}} & \multicolumn{1}{c}{\textbf{Language}} & \multicolumn{1}{l|}{\textbf{Overall}} & \multicolumn{1}{c}{\textbf{Reasoning}} & \multicolumn{1}{c}{\textbf{Language}} & \multicolumn{1}{l|}{\textbf{Overall}} &       &  \\
    \midrule
    \multirow{3}[2]{*}{Qwen2.5-3B-SFT-Open} & \href{https://github.com/HIT-SCIR/huozi/tree/main/data/huozi-rlhf}{Huozi-RLHF} \citep{huozi}  & 4.02  & 5.23  & 4.63  & 3.88  & 5.04  & 4.46  & 5.03  & 4.94  \\
          & \href{https://huggingface.co/datasets/wenbopan/Chinese-dpo-pairs}{Chinese-DPO-Pairs} & 4.46  & 5.52  & 4.99  & 4.20  & 5.24  & 4.72  & 5.04  & 4.95  \\
          & \cellcolor[rgb]{ .906,  .902,  .902}TaP (GPT-4) & \cellcolor[rgb]{ .906,  .902,  .902}\textbf{5.19} & \cellcolor[rgb]{ .906,  .902,  .902}\textbf{5.90} & \cellcolor[rgb]{ .906,  .902,  .902}\textbf{5.55} & \cellcolor[rgb]{ .906,  .902,  .902}\textbf{4.79} & \cellcolor[rgb]{ .906,  .902,  .902}\textbf{5.62} & \cellcolor[rgb]{ .906,  .902,  .902}\textbf{5.21} & \cellcolor[rgb]{ .906,  .902,  .902}\textbf{5.88} & \cellcolor[rgb]{ .906,  .902,  .902}\textbf{5.63} \\
    \midrule
    \multirow{3}[2]{*}{Qwen2.5-3B-SFT-TaP} & \href{https://github.com/HIT-SCIR/huozi/tree/main/data/huozi-rlhf}{Huozi-RLHF} \citep{huozi}  & 5.15  & 5.82  & 5.49  & 4.65  & 5.50  & 5.08  & 5.72  & 5.24  \\
          & \href{https://huggingface.co/datasets/wenbopan/Chinese-dpo-pairs}{Chinese-DPO-Pairs} & 5.25  & 6.01  & 5.63  & \textbf{4.90} & 5.51  & 5.20  & 5.91  & 5.68  \\
          & \cellcolor[rgb]{ .906,  .902,  .902}TaP (GPT-4) & \cellcolor[rgb]{ .906,  .902,  .902}\textbf{5.57} & \cellcolor[rgb]{ .906,  .902,  .902}\textbf{6.06} & \cellcolor[rgb]{ .906,  .902,  .902}\textbf{5.81} & \cellcolor[rgb]{ .906,  .902,  .902}\textbf{4.90} & \cellcolor[rgb]{ .906,  .902,  .902}\textbf{5.65} & \cellcolor[rgb]{ .906,  .902,  .902}\textbf{5.28} & \cellcolor[rgb]{ .906,  .902,  .902}\textbf{6.03} & \cellcolor[rgb]{ .906,  .902,  .902}\textbf{5.79} \\
    \midrule
    \multirow{3}[2]{*}{Qwen2.5-7B-SFT-Open} & \href{https://github.com/HIT-SCIR/huozi/tree/main/data/huozi-rlhf}{Huozi-RLHF} \citep{huozi}  & 4.53  & 5.39  & 4.96  & 4.37  & 5.34  & 4.85  & 5.25  & 4.98  \\
          & \href{https://huggingface.co/datasets/wenbopan/Chinese-dpo-pairs}{Chinese-DPO-Pairs} & 4.88  & 5.53  & 5.21  & 4.64  & 5.40  & 5.02  & 5.23  & 5.23  \\
          & \cellcolor[rgb]{ .906,  .902,  .902}TaP (GPT-4) & \cellcolor[rgb]{ .906,  .902,  .902}\textbf{5.11} & \cellcolor[rgb]{ .906,  .902,  .902}\textbf{5.74} & \cellcolor[rgb]{ .906,  .902,  .902}\textbf{5.43} & \cellcolor[rgb]{ .906,  .902,  .902}\textbf{4.87} & \cellcolor[rgb]{ .906,  .902,  .902}\textbf{5.58} & \cellcolor[rgb]{ .906,  .902,  .902}\textbf{5.22} & \cellcolor[rgb]{ .906,  .902,  .902}\textbf{5.54} & \cellcolor[rgb]{ .906,  .902,  .902}\textbf{5.38} \\
    \midrule
    \multirow{3}[2]{*}{Qwen2.5-7B-SFT-TaP} & \href{https://github.com/HIT-SCIR/huozi/tree/main/data/huozi-rlhf}{Huozi-RLHF} \citep{huozi}  & 6.24  & 6.34  & 6.29  & 5.83  & 6.00  & 5.92  & 6.10  & 5.85  \\
          & \href{https://huggingface.co/datasets/wenbopan/Chinese-dpo-pairs}{Chinese-DPO-Pairs} & 6.49  & 6.42  & 6.46  & 5.94  & 6.09  & 6.01  & 6.16  & 5.98  \\
          & \cellcolor[rgb]{ .906,  .902,  .902}TaP (GPT-4) & \cellcolor[rgb]{ .906,  .902,  .902}\textbf{6.59} & \cellcolor[rgb]{ .906,  .902,  .902}\textbf{6.66} & \cellcolor[rgb]{ .906,  .902,  .902}\textbf{6.62} & \cellcolor[rgb]{ .906,  .902,  .902}\textbf{6.03} & \cellcolor[rgb]{ .906,  .902,  .902}\textbf{6.18} & \cellcolor[rgb]{ .906,  .902,  .902}\textbf{6.11} & \cellcolor[rgb]{ .906,  .902,  .902}\textbf{6.36} & \cellcolor[rgb]{ .906,  .902,  .902}\textbf{6.11} \\
    \midrule
    \multirow{3}[2]{*}{Llama-3.1-8B-SFT-Open} & \href{https://github.com/HIT-SCIR/huozi/tree/main/data/huozi-rlhf}{Huozi-RLHF} \citep{huozi}  & 3.52  & 5.19  & 4.36  & 3.29  & 5.06  & 4.17  & 4.85  & 4.58  \\
          & \href{https://huggingface.co/datasets/wenbopan/Chinese-dpo-pairs}{Chinese-DPO-Pairs} & 3.53  & 5.57  & 4.55  & 3.29  & 5.29  & 4.29  & 4.94  & 4.88  \\
          & \cellcolor[rgb]{ .906,  .902,  .902}TaP (GPT-4) & \cellcolor[rgb]{ .906,  .902,  .902}\textbf{3.96} & \cellcolor[rgb]{ .906,  .902,  .902}\textbf{5.62} & \cellcolor[rgb]{ .906,  .902,  .902}\textbf{4.79} & \cellcolor[rgb]{ .906,  .902,  .902}\textbf{3.50} & \cellcolor[rgb]{ .906,  .902,  .902}\textbf{5.37} & \cellcolor[rgb]{ .906,  .902,  .902}\textbf{4.43} & \cellcolor[rgb]{ .906,  .902,  .902}\textbf{5.41} & \cellcolor[rgb]{ .906,  .902,  .902}\textbf{5.14} \\
    \midrule
    \multirow{3}[2]{*}{Llama-3.1-8B-SFT-TaP} & \href{https://github.com/HIT-SCIR/huozi/tree/main/data/huozi-rlhf}{Huozi-RLHF} \citep{huozi}  & 3.70  & 5.22  & 4.46  & 3.33  & 4.86  & 4.09  & 5.19  & 4.84  \\
          & \href{https://huggingface.co/datasets/wenbopan/Chinese-dpo-pairs}{Chinese-DPO-Pairs} & 4.07  & 5.36  & 4.71  & \textbf{3.59} & \textbf{5.07} & \textbf{4.33} & 5.30  & 4.84  \\
          & \cellcolor[rgb]{ .906,  .902,  .902}TaP (GPT-4) & \cellcolor[rgb]{ .906,  .902,  .902}\textbf{4.10} & \cellcolor[rgb]{ .906,  .902,  .902}\textbf{5.50} & \cellcolor[rgb]{ .906,  .902,  .902}\textbf{4.80} & \cellcolor[rgb]{ .906,  .902,  .902}3.52  & \cellcolor[rgb]{ .906,  .902,  .902}4.99  & \cellcolor[rgb]{ .906,  .902,  .902}4.25  & \cellcolor[rgb]{ .906,  .902,  .902}\textbf{5.59} & \cellcolor[rgb]{ .906,  .902,  .902}\textbf{5.16} \\
    \midrule
    \multirow{3}[2]{*}{Gemma-2-9B-SFT-Open} & \href{https://github.com/HIT-SCIR/huozi/tree/main/data/huozi-rlhf}{Huozi-RLHF} \citep{huozi}  & 3.93  & 5.22  & 4.58  & 3.54  & 5.05  & 4.29  & 4.77  & 4.62  \\
          & \href{https://huggingface.co/datasets/wenbopan/Chinese-dpo-pairs}{Chinese-DPO-Pairs} & 3.86  & 5.32  & 4.59  & 3.61  & 5.21  & 4.41  & 4.91  & 4.84  \\
          & \cellcolor[rgb]{ .906,  .902,  .902}TaP (GPT-4) & \cellcolor[rgb]{ .906,  .902,  .902}\textbf{4.55} & \cellcolor[rgb]{ .906,  .902,  .902}\textbf{5.90} & \cellcolor[rgb]{ .906,  .902,  .902}\textbf{5.22} & \cellcolor[rgb]{ .906,  .902,  .902}\textbf{4.09} & \cellcolor[rgb]{ .906,  .902,  .902}\textbf{5.44} & \cellcolor[rgb]{ .906,  .902,  .902}\textbf{4.76} & \cellcolor[rgb]{ .906,  .902,  .902}\textbf{5.54} & \cellcolor[rgb]{ .906,  .902,  .902}\textbf{5.23} \\
    \midrule
    \multirow{3}[2]{*}{Gemma-2-9B-SFT-TaP} & \href{https://github.com/HIT-SCIR/huozi/tree/main/data/huozi-rlhf}{Huozi-RLHF} \citep{huozi}  & 4.18  & 5.23  & 4.70  & 3.77  & 4.93  & 4.35  & 5.39  & 5.01  \\
          & \href{https://huggingface.co/datasets/wenbopan/Chinese-dpo-pairs}{Chinese-DPO-Pairs} & 4.67  & 5.52  & 5.09  & \textbf{4.09} & \textbf{5.21} & \textbf{4.65} & 5.54  & 5.20  \\
          & \cellcolor[rgb]{ .906,  .902,  .902}TaP (GPT-4) & \cellcolor[rgb]{ .906,  .902,  .902}\textbf{4.72} & \cellcolor[rgb]{ .906,  .902,  .902}\textbf{5.61} & \cellcolor[rgb]{ .906,  .902,  .902}\textbf{5.16} & \cellcolor[rgb]{ .906,  .902,  .902}4.05  & \cellcolor[rgb]{ .906,  .902,  .902}5.14  & \cellcolor[rgb]{ .906,  .902,  .902}4.60  & \cellcolor[rgb]{ .906,  .902,  .902}\textbf{5.78} & \cellcolor[rgb]{ .906,  .902,  .902}\textbf{5.33} \\
    \midrule
    \multirow{3}[2]{*}{Qwen2.5-14B-SFT-Open} & \href{https://github.com/HIT-SCIR/huozi/tree/main/data/huozi-rlhf}{Huozi-RLHF} \citep{huozi}  & 5.75  & 5.90  & 5.82  & 5.64  & 5.81  & 5.73  & 5.41  & 5.48  \\
          & \href{https://huggingface.co/datasets/wenbopan/Chinese-dpo-pairs}{Chinese-DPO-Pairs} & 5.85  & 5.96  & 5.90  & 5.47  & 5.79  & 5.63  & 5.54  & 5.58  \\
          & \cellcolor[rgb]{ .906,  .902,  .902}TaP (GPT-4) & \cellcolor[rgb]{ .906,  .902,  .902}\textbf{6.44} & \cellcolor[rgb]{ .906,  .902,  .902}\textbf{6.39} & \cellcolor[rgb]{ .906,  .902,  .902}\textbf{6.42} & \cellcolor[rgb]{ .906,  .902,  .902}\textbf{6.05} & \cellcolor[rgb]{ .906,  .902,  .902}\textbf{6.17} & \cellcolor[rgb]{ .906,  .902,  .902}\textbf{6.11} & \cellcolor[rgb]{ .906,  .902,  .902}\textbf{5.84} & \cellcolor[rgb]{ .906,  .902,  .902}\textbf{5.89} \\
    \midrule
    \multirow{3}[2]{*}{Qwen2.5-14B-SFT-TaP} & \href{https://github.com/HIT-SCIR/huozi/tree/main/data/huozi-rlhf}{Huozi-RLHF} \citep{huozi}  & 6.94  & 6.64  & 6.79  & 6.41  & 6.28  & 6.35  & 6.25  & 6.11  \\
          & \href{https://huggingface.co/datasets/wenbopan/Chinese-dpo-pairs}{Chinese-DPO-Pairs} & 6.99  & 6.85  & 6.92  & 6.41  & 6.43  & 6.42  & 6.34  & 6.14  \\
          & \cellcolor[rgb]{ .906,  .902,  .902}TaP (GPT-4) & \cellcolor[rgb]{ .906,  .902,  .902}\textbf{7.27} & \cellcolor[rgb]{ .906,  .902,  .902}\textbf{7.08} & \cellcolor[rgb]{ .906,  .902,  .902}\textbf{7.17} & \cellcolor[rgb]{ .906,  .902,  .902}\textbf{6.52} & \cellcolor[rgb]{ .906,  .902,  .902}\textbf{6.52} & \cellcolor[rgb]{ .906,  .902,  .902}\textbf{6.52} & \cellcolor[rgb]{ .906,  .902,  .902}\textbf{6.66} & \cellcolor[rgb]{ .906,  .902,  .902}\textbf{6.35} \\
    \bottomrule
    \end{tabular}%
  \caption{Performance comparison of LLMs trained with \textbf{DPO} using different datasets on AlignBench and MT-Bench-zh. The model names include two possible suffixes: ``Open'' and ``TaP.'' The ``Open'' suffix indicates that the LLMs were initialized from models trained via supervised fine-tuning on open-source datasets, whereas ``TaP'' denotes initialization from models trained on a dataset constructed by TaP. Additionally, ``GPT-4o'' and ``DeepSeek-V3'' specify that responses were evaluated using GPT-4o and DeepSeek-V3, respectively.}
  \label{tab:dpo_all}%
\end{table*}%

\begin{table*}[!t]
  \centering
  \tiny
    \begin{tabular}{c|l|ccc|ccc|cc}
    \toprule
    \multirow{3}[6]{*}{\textbf{Model}} & \multicolumn{1}{c|}{\multirow{3}[6]{*}{\textbf{Dataset}}} & \multicolumn{6}{c|}{\textbf{AlignBench}}      & \multicolumn{2}{c}{\textbf{MT-Bench-zh}} \\
\cmidrule{3-10}          &       & \multicolumn{3}{c|}{\textbf{GPT-4o}} & \multicolumn{3}{c|}{\textbf{DeepSeek-V3}} & \multicolumn{1}{c}{\multirow{2}[4]{*}{\textbf{GPT-4o}}} & \multicolumn{1}{c}{\multirow{2}[4]{*}{\textbf{DeepSeek-V3}}} \\
\cmidrule{3-8}          &       & \multicolumn{1}{c}{\textbf{Reasoning}} & \multicolumn{1}{c}{\textbf{Language}} & \multicolumn{1}{l|}{\textbf{Overall}} & \multicolumn{1}{c}{\textbf{Reasoning}} & \multicolumn{1}{c}{\textbf{Language}} & \multicolumn{1}{l|}{\textbf{Overall}} &       &  \\
    \midrule
    \multirow{3}[2]{*}{Qwen2.5-3B-SFT-Open} & \href{https://github.com/HIT-SCIR/huozi/tree/main/data/huozi-rlhf}{Huozi-RLHF} \citep{huozi}  & 4.13  & 4.88  & 4.51  & 4.44  & 5.36  & 4.90  & 5.25  & 5.20  \\
          & \href{https://huggingface.co/datasets/wenbopan/Chinese-dpo-pairs}{Chinese-DPO-Pairs} & 5.12  & 5.98  & 5.55  & 4.53  & 5.52  & 5.02  & \textbf{5.99} & \textbf{5.71} \\
          & \cellcolor[rgb]{ .906,  .902,  .902}TaP (GPT-4) & \cellcolor[rgb]{ .906,  .902,  .902}\textbf{5.20} & \cellcolor[rgb]{ .906,  .902,  .902}\textbf{6.11} & \cellcolor[rgb]{ .906,  .902,  .902}\textbf{5.66} & \cellcolor[rgb]{ .906,  .902,  .902}\textbf{4.71} & \cellcolor[rgb]{ .906,  .902,  .902}\textbf{5.62} & \cellcolor[rgb]{ .906,  .902,  .902}\textbf{5.17} & \cellcolor[rgb]{ .906,  .902,  .902}5.80  & \cellcolor[rgb]{ .906,  .902,  .902}5.68  \\
    \midrule
    \multirow{3}[2]{*}{Qwen2.5-3B-SFT-TaP} & \href{https://github.com/HIT-SCIR/huozi/tree/main/data/huozi-rlhf}{Huozi-RLHF} \citep{huozi}  & 4.61  & 5.27  & 4.94  & 4.62  & 5.55  & 5.08  & 5.86  & 5.76  \\
          & \href{https://huggingface.co/datasets/wenbopan/Chinese-dpo-pairs}{Chinese-DPO-Pairs} & 4.88  & 5.51  & 5.20  & \textbf{5.07} & \textbf{5.77} & \textbf{5.42} & \textbf{6.08} & 5.79  \\
          & \cellcolor[rgb]{ .906,  .902,  .902}TaP (GPT-4) & \cellcolor[rgb]{ .906,  .902,  .902}\textbf{5.69} & \cellcolor[rgb]{ .906,  .902,  .902}\textbf{6.27} & \cellcolor[rgb]{ .906,  .902,  .902}\textbf{5.98} & \cellcolor[rgb]{ .906,  .902,  .902}4.98  & \cellcolor[rgb]{ .906,  .902,  .902}5.68  & \cellcolor[rgb]{ .906,  .902,  .902}5.33  & \cellcolor[rgb]{ .906,  .902,  .902}6.00  & \cellcolor[rgb]{ .906,  .902,  .902}\textbf{5.81} \\
    \midrule
    
    \multirow{3}[2]{*}{Qwen2.5-7B-SFT-Open} & \href{https://github.com/HIT-SCIR/huozi/tree/main/data/huozi-rlhf}{Huozi-RLHF} \citep{huozi}  & 4.71  & 5.56  & 5.13  & 4.46  & 5.43  & 4.95  & 5.21  & 5.19  \\
          & \href{https://huggingface.co/datasets/wenbopan/Chinese-dpo-pairs}{Chinese-DPO-Pairs} & 4.41  & 5.33  & 4.87  & 4.31  & \textbf{5.52} & 4.91  & 5.23  & 5.14  \\
          & \cellcolor[rgb]{ .906,  .902,  .902}TaP (GPT-4) & \cellcolor[rgb]{ .906,  .902,  .902}\textbf{4.94} & \cellcolor[rgb]{ .906,  .902,  .902}\textbf{5.81} & \cellcolor[rgb]{ .906,  .902,  .902}\textbf{5.38} & \cellcolor[rgb]{ .906,  .902,  .902}\textbf{4.64} & \cellcolor[rgb]{ .906,  .902,  .902}5.41  & \cellcolor[rgb]{ .906,  .902,  .902}\textbf{5.03} & \cellcolor[rgb]{ .906,  .902,  .902}\textbf{5.39} & \cellcolor[rgb]{ .906,  .902,  .902}\textbf{5.49} \\
    \midrule
    \multirow{3}[2]{*}{Qwen2.5-7B-SFT-TaP} & \href{https://github.com/HIT-SCIR/huozi/tree/main/data/huozi-rlhf}{Huozi-RLHF} \citep{huozi}  & 6.24  & 6.44  & 6.34  & 5.67  & 5.98  & 5.82  & 6.25  & 6.19  \\
          & \href{https://huggingface.co/datasets/wenbopan/Chinese-dpo-pairs}{Chinese-DPO-Pairs} & 5.79  & 6.02  & 5.90  & \textbf{5.89} & 6.04  & \textbf{5.97} & 6.35  & 6.31  \\
          & \cellcolor[rgb]{ .906,  .902,  .902}TaP (GPT-4) & \cellcolor[rgb]{ .906,  .902,  .902}\textbf{6.29} & \cellcolor[rgb]{ .906,  .902,  .902}\textbf{6.62} & \cellcolor[rgb]{ .906,  .902,  .902}\textbf{6.46} & \cellcolor[rgb]{ .906,  .902,  .902}5.73  & \cellcolor[rgb]{ .906,  .902,  .902}\textbf{6.15} & \cellcolor[rgb]{ .906,  .902,  .902}5.94  & \cellcolor[rgb]{ .906,  .902,  .902}\textbf{6.36} & \cellcolor[rgb]{ .906,  .902,  .902}\textbf{6.38} \\
    \midrule
    \multirow{3}[2]{*}{Llama-3.1-8B-SFT-Open} & \href{https://github.com/HIT-SCIR/huozi/tree/main/data/huozi-rlhf}{Huozi-RLHF} \citep{huozi}  & 3.72  & 5.27  & 4.50  & \textbf{3.41} & 4.97  & 4.19  & 4.93  & 4.89  \\
          & \href{https://huggingface.co/datasets/wenbopan/Chinese-dpo-pairs}{Chinese-DPO-Pairs} & 2.79  & 4.52  & 3.66  & 3.10  & 5.03  & 4.06  & 4.75  & 4.93  \\
          & \cellcolor[rgb]{ .906,  .902,  .902}TaP (GPT-4) & \cellcolor[rgb]{ .906,  .902,  .902}\textbf{3.78} & \cellcolor[rgb]{ .906,  .902,  .902}\textbf{5.57} & \cellcolor[rgb]{ .906,  .902,  .902}\textbf{4.68} & \cellcolor[rgb]{ .906,  .902,  .902}3.33  & \cellcolor[rgb]{ .906,  .902,  .902}\textbf{5.28} & \cellcolor[rgb]{ .906,  .902,  .902}\textbf{4.30} & \cellcolor[rgb]{ .906,  .902,  .902}\textbf{5.14} & \cellcolor[rgb]{ .906,  .902,  .902}\textbf{4.94} \\
    \midrule
    \multirow{3}[2]{*}{Llama-3.1-8B-SFT-TaP} & \href{https://github.com/HIT-SCIR/huozi/tree/main/data/huozi-rlhf}{Huozi-RLHF} \citep{huozi}  & 3.94  & 5.25  & 4.60  & \textbf{3.61} & 4.86  & 4.24  & 5.25  & 5.04  \\
          & \href{https://huggingface.co/datasets/wenbopan/Chinese-dpo-pairs}{Chinese-DPO-Pairs} & 3.72  & 5.09  & 4.40  & 3.30  & 4.89  & 4.09  & 5.21  & 5.04  \\
          & \cellcolor[rgb]{ .906,  .902,  .902}TaP (GPT-4) & \cellcolor[rgb]{ .906,  .902,  .902}\textbf{4.04} & \cellcolor[rgb]{ .906,  .902,  .902}\textbf{5.59} & \cellcolor[rgb]{ .906,  .902,  .902}\textbf{4.81} & \cellcolor[rgb]{ .906,  .902,  .902}3.43  & \cellcolor[rgb]{ .906,  .902,  .902}\textbf{4.94} & \cellcolor[rgb]{ .906,  .902,  .902}\textbf{4.18} & \cellcolor[rgb]{ .906,  .902,  .902}\textbf{5.46} & \cellcolor[rgb]{ .906,  .902,  .902}\textbf{5.28} \\
    \midrule
    \multirow{3}[2]{*}{Gemma-2-9B-SFT-Open} & \href{https://github.com/HIT-SCIR/huozi/tree/main/data/huozi-rlhf}{Huozi-RLHF} \citep{huozi}  & 3.75  & 4.75  & 4.25  & 3.81  & 5.14  & 4.47  & 4.87  & 4.85  \\
          & \href{https://huggingface.co/datasets/wenbopan/Chinese-dpo-pairs}{Chinese-DPO-Pairs} & 4.05  & 5.48  & 4.77  & 3.73  & 5.25  & 4.49  & 5.03  & 4.99  \\
          & \cellcolor[rgb]{ .906,  .902,  .902}TaP (GPT-4) & \cellcolor[rgb]{ .906,  .902,  .902}\textbf{4.49} & \cellcolor[rgb]{ .906,  .902,  .902}\textbf{5.89} & \cellcolor[rgb]{ .906,  .902,  .902}\textbf{5.19} & \cellcolor[rgb]{ .906,  .902,  .902}\textbf{3.96} & \cellcolor[rgb]{ .906,  .902,  .902}\textbf{5.55} & \cellcolor[rgb]{ .906,  .902,  .902}\textbf{4.75} & \cellcolor[rgb]{ .906,  .902,  .902}\textbf{5.52} & \cellcolor[rgb]{ .906,  .902,  .902}\textbf{5.50} \\
    \midrule
    \multirow{3}[2]{*}{Gemma-2-9B-SFT-TaP} & \href{https://github.com/HIT-SCIR/huozi/tree/main/data/huozi-rlhf}{Huozi-RLHF} \citep{huozi}  & 3.85  & 5.20  & 4.53  & 3.56  & 4.96  & 4.26  & 5.27  & 5.11  \\
          & \href{https://huggingface.co/datasets/wenbopan/Chinese-dpo-pairs}{Chinese-DPO-Pairs} & 4.14  & 5.32  & 4.73  & 3.55  & 4.91  & 4.23  & 5.45  & 5.18  \\
          & \cellcolor[rgb]{ .906,  .902,  .902}TaP (GPT-4) & \cellcolor[rgb]{ .906,  .902,  .902}\textbf{4.42} & \cellcolor[rgb]{ .906,  .902,  .902}\textbf{5.58} & \cellcolor[rgb]{ .906,  .902,  .902}\textbf{5.00} & \cellcolor[rgb]{ .906,  .902,  .902}\textbf{3.91} & \cellcolor[rgb]{ .906,  .902,  .902}\textbf{5.23} & \cellcolor[rgb]{ .906,  .902,  .902}\textbf{4.57} & \cellcolor[rgb]{ .906,  .902,  .902}\textbf{5.63} & \cellcolor[rgb]{ .906,  .902,  .902}\textbf{5.26} \\
    \midrule
    \multirow{3}[2]{*}{Qwen2.5-14B-SFT-Open} & \href{https://github.com/HIT-SCIR/huozi/tree/main/data/huozi-rlhf}{Huozi-RLHF} \citep{huozi}  & 5.61  & 5.90  & 5.75  & 5.69  & 5.93  & 5.81  & 5.78  & 5.78  \\
          & \href{https://huggingface.co/datasets/wenbopan/Chinese-dpo-pairs}{Chinese-DPO-Pairs} & 6.05  & 6.30  & 6.18  & 5.85  & \textbf{6.17} & 6.01  & 5.86  & 5.78  \\
          & \cellcolor[rgb]{ .906,  .902,  .902}TaP (GPT-4) & \cellcolor[rgb]{ .906,  .902,  .902}\textbf{6.53} & \cellcolor[rgb]{ .906,  .902,  .902}\textbf{6.49} & \cellcolor[rgb]{ .906,  .902,  .902}\textbf{6.51} & \cellcolor[rgb]{ .906,  .902,  .902}\textbf{6.02} & \cellcolor[rgb]{ .906,  .902,  .902}6.16  & \cellcolor[rgb]{ .906,  .902,  .902}\textbf{6.09} & \cellcolor[rgb]{ .906,  .902,  .902}\textbf{6.32} & \cellcolor[rgb]{ .906,  .902,  .902}\textbf{6.21} \\
    \midrule
    \multirow{3}[2]{*}{Qwen2.5-14B-SFT-TaP} & \href{https://github.com/HIT-SCIR/huozi/tree/main/data/huozi-rlhf}{Huozi-RLHF} \citep{huozi}  & 6.61  & 6.50  & 6.55  & 6.42  & 6.34  & 6.38  & 6.41  & \textbf{6.51} \\
          & \href{https://huggingface.co/datasets/wenbopan/Chinese-dpo-pairs}{Chinese-DPO-Pairs} & 6.37  & 6.40  & 6.38  & 6.48  & \textbf{6.59} & 6.53  & \textbf{6.83} & 6.46  \\
          & \cellcolor[rgb]{ .906,  .902,  .902}TaP (GPT-4) & \cellcolor[rgb]{ .906,  .902,  .902}\textbf{7.42} & \cellcolor[rgb]{ .906,  .902,  .902}\textbf{7.12} & \cellcolor[rgb]{ .906,  .902,  .902}\textbf{7.27} & \cellcolor[rgb]{ .906,  .902,  .902}\textbf{6.84} & \cellcolor[rgb]{ .906,  .902,  .902}6.50  & \cellcolor[rgb]{ .906,  .902,  .902}\textbf{6.67} & \cellcolor[rgb]{ .906,  .902,  .902}6.58  & \cellcolor[rgb]{ .906,  .902,  .902}6.43  \\
    \bottomrule
    \end{tabular}%
  \caption{Performance comparison of LLMs trained with \textbf{PPO} using different datasets on AlignBench and MT-Bench-zh. The model names include two possible suffixes: ``Open'' and ``TaP.'' The ``Open'' suffix indicates that the LLMs were initialized from models trained via supervised fine-tuning on open-source datasets, whereas ``TaP'' denotes initialization from models trained on a dataset constructed by TaP. Additionally, ``GPT-4o'' and ``DeepSeek-V3'' specify that responses were evaluated using GPT-4o and DeepSeek-V3, respectively.}
  \label{tab:ppo_all}%
\end{table*}%

\subsection{Evaluation}

To evaluate the performance of trained LLMs, we employ MT-Bench-zh\footnote{\url{https://github.com/HIT-SCIR/huozi/tree/main/data/mt-bench-zh}} as the validation set, a Chinese translation of MT-Bench \citep{DBLP:conf/nips/ZhengC00WZL0LXZ23}. For testing, we adopt AlignBench \citep{DBLP:conf/acl/LiuLWHFWCKXTZ0G24}. Both benchmarks rely on strong LLMs (e.g., GPT-4o) to evaluate model responses. Accordingly, we report scores from both GPT-4o and DeepSeek-V3 for a comprehensive evaluation. Further details on the evaluation setup are provided in Appendix~\ref{subsec:appendix_evaluation}.

\section{Experimental Results}

We present the aggregated experimental results for supervised fine-tuning, DPO, and PPO in Table~\ref{tab:sft_all}, \ref{tab:dpo_all}, and \ref{tab:ppo_all}, respectively. The corresponding detailed experimental results are provided in Tables~\ref{tab:sft_mt_bench} through \ref{tab:ppo_alignbench_deepseek} in Appendix~\ref{sec:appendix_experimental_results}. When trained on the same dataset, we observe that larger LLMs consistently outperform their smaller counterparts within the Qwen2.5 series. Moreover, LLMs trained on data constructed using TaP consistently demonstrate strong performance across supervised fine-tuning, DPO, and PPO. This finding underscores the effectiveness of TaP in generating high-quality training data for both supervised and preference fine-tuning of LLMs. Furthermore, the experimental results yield additional findings:

\paragraph{To ensure optimal model performance when scaling up data for supervised fine-tuning, it is crucial to consider factors such as data quality, diversity, and coverage.} The experimental results in Table~\ref{tab:sft_all} suggest that despite the relatively small size of \texttt{TaP-SFT (GPT-4)}, LLMs trained on this dataset outperform those trained on other open-source datasets. \textbf{Notably, LLMs trained on \texttt{TaP-SFT (GPT-4)} consistently surpass those trained on \texttt{BELLE-SFT}, even though \texttt{BELLE-SFT} has over 180$\times$ more samples.} This finding indicates that, beyond dataset size, factors such as data quality, diversity, and coverage play critical roles in model performance. These results align with the findings of LIMA \citep{DBLP:conf/nips/ZhouLX0SMMEYYZG23}, which show that LLMs trained on only 1,000 carefully curated samples can still achieve strong performance. However, LIMA's heavy reliance on human curation poses scalability challenges. In contrast, TaP requires human involvement only to design the taxonomy,\footnote{Instead of constructing the taxonomy manually, the taxonomy can also be constructed by LLMs alone or through human--LLM collaboration.} making it significantly more scalable.

\paragraph{The capabilities of pre-trained LLMs provide a critical foundation for the performance of models subsequently trained through supervised fine-tuning or preference fine-tuning.} The experimental results in Tables~\ref{tab:sft_all}, \ref{tab:dpo_all}, and \ref{tab:ppo_all} indicate that LLMs initialized from Qwen2.5-3B generally outperform those initialized from Llama-3.1-8B and Gemma-2-9B, even though the latter two have nearly three times as many parameters as Qwen2.5-3B when trained on the same dataset. We hypothesize that this gap arises from the limited volume of Chinese corpora used to pre-train Llama-3.1-8B and Gemma-2-9B, which constrains their ability to generate high-quality Chinese responses after supervised fine-tuning or preference fine-tuning. In contrast, the Qwen2.5 series of LLMs have been extensively trained on Chinese corpora, enhancing their capacity to follow instructions and better aligning with human preferences and values.

\paragraph{The effectiveness of applying supervised fine-tuning and preference fine-tuning using the same training dataset varies across LLMs.} The experimental results presented in Tables~\ref{tab:sft_all}, \ref{tab:dpo_all}, and \ref{tab:ppo_all} indicate that the optimal open-source datasets for fine-tuning vary across LLMs. For instance, as shown in Table~\ref{tab:sft_all}, when GPT-4o serves as the judge, the open-source fine-tuning dataset that yields the best performance for Qwen2.5-7B on MT-Bench-zh is MOSS-SFT. However, for Llama-3.1-8B, the open-source dataset that yields the best performance under the same evaluation conditions is Infinity-Instruct. These findings align with the claim proposed by \citet{DBLP:journals/corr/abs-2502-04194}, which posits that supervised fine-tuning datasets that are more closely aligned with the distribution of pre-trained LLMs tend to be more effective. Nevertheless, LLMs trained on data constructed by TaP consistently achieve strong performance and generally outperform those trained on other open-source datasets. These results demonstrates that data constructed by TaP can be effectively employed for both supervised fine-tuning and preference fine-tuning across a range of LLMs.

\section{Discussion}
Although LLMs trained on the synthetic dataset constructed by TaP consistently outperform those trained on other open-source datasets, the mechanisms behind this strong performance remain insufficiently understood. More broadly, while LLMs trained on different datasets can exhibit substantial performance differences, a comprehensive theory explaining why training on one dataset yields better performance than another remains an open problem, and a complete theoretical account of this phenomenon is still lacking. Nevertheless, we offer our current understanding as follows.

Recent studies have shown that LLMs benefit substantially from training data that are both diverse and knowledge-rich \citep{DBLP:journals/corr/abs-2309-05463,DBLP:conf/iclr/ZhangZZCWZBQ0FY25}. Similarly, \citet{lozhkov2024fineweb-edu,DBLP:journals/corr/abs-2502-02737} construct large-scale corpora by selectively retaining samples with high educational value. We hypothesize that the superior performance of LLMs trained on the TaP-constructed synthetic dataset, relative to those trained on open-source datasets, can be attributed to several desirable properties of the samples it contains: (1) they are knowledge-rich, (2) they exhibit high information density, and (3) they span a broad range of domains and tasks. Figure~\ref{fig:illustrative_example} in Appendix~\ref{sec:appendix_an_illustrative_example} provides an illustrative example highlighting these properties. These desirable characteristics arise primarily from the taxonomy employed in the TaP framework to guide synthetic data generation. Specifically, the taxonomy is derived from the \textit{Undergraduate Program Catalog of Regular Higher Education Institutions of China}, which constitutes a comprehensive and well-structured repository of foundational knowledge across disciplines. As a result, the synthetic samples generated by TaP capture a wide spectrum of essential knowledge, thereby making them particularly effective for training LLMs.

\section{Conclusion}
In this paper, we propose the TaP framework to enable the automated, scalable construction of preference datasets across languages. Grounded in a structured taxonomy, TaP offers fine-grained control over dataset composition, ensuring both diversity and broad coverage. To evaluate TaP and address the scarcity of Chinese preference data, we apply TaP to construct Chinese preference datasets using a taxonomy derived from the \textit{Undergraduate Program Catalog of Regular Higher Education Institutions of China}. We perform supervised fine-tuning and preference fine-tuning on LLMs ranging from 3B to 14B parameters across three model families using the TaP-generated datasets. Experimental results show that LLMs trained on TaP-generated datasets outperform those trained on open-source datasets.

\section*{Limitations}
Previous studies have highlighted the risk of model collapse when training LLMs on synthetic data \citep{DBLP:journals/corr/abs-2404-01413,DBLP:conf/icml/KazdanSDGRDK25,DBLP:journals/nature/ShumailovSZPAG24}. As the TaP framework can be employed to generate synthetic data, LLMs trained on TaP-constructed datasets may also be susceptible to model collapse; this is a limitation of TaP. Nevertheless, our work focuses on automated and scalable preference dataset construction rather than mitigating the risks associated with synthetic data. Substantial research has been devoted to preventing model collapse caused by synthetic data, such as mixing real and synthetic data \citep{DBLP:journals/corr/abs-2404-05090}, selecting samples based on entropy metrics \citep{DBLP:journals/corr/abs-2509-16499}, and proposing new training objectives to counteract overconfidence \citep{shabgahi2025fortifaifendingrecursivetraining}. Addressing these risks is beyond the scope of this study, and we leave it for future work.

\clearpage
\bibliography{anthology}

\clearpage

\appendix

\section{Alternative Approaches for Constructing the Taxonomy}
\label{sec:appendix_alternative_approaches_for_constructing_the_taxonomy}

In cases where the undergraduate program catalog is unavailable, the undergraduate program catalogs from high-resource languages can serve as representative taxonomies to guide the construction of analogous taxonomies for other languages. Furthermore, we emphasize that using the undergraduate program catalog as the taxonomy represents a specific and illustrative case of applying the TaP framework for synthetic data generation. The taxonomy employed by the TaP framework is not limited to the undergraduate program catalog. An alternative approach to constructing the taxonomy involves analyzing and categorizing the potential use cases of LLMs, a method that can be extended to other languages.

While comprehensively enumerating and categorizing all possible use cases of LLMs is inherently challenging, a practical strategy is to collect user–model interactions and analyze these conversations to systematically identify and categorize use cases \citep{DBLP:conf/iclr/Zhao0HC0D24,DBLP:conf/iclr/ZhengC0LZW00LXG24}. For low-resource languages with limited resources, a feasible approach would be to initially deploy a LLM trained on all available data for these languages, potentially supplemented with data translated from high-resource languages. Following deployment, user–model conversations can be continuously collected and analyzed to construct the taxonomy grounded in real-world use cases. Additionally, both the taxonomy and the LLMs themselves can be incrementally updated as sufficient new user-model conversations become available, thereby improving model performance over time.

\section{An Illustrative Example}
\label{sec:appendix_an_illustrative_example}
\begin{figure*}[!t]
    \centering
    \includegraphics[width=.75\textwidth]{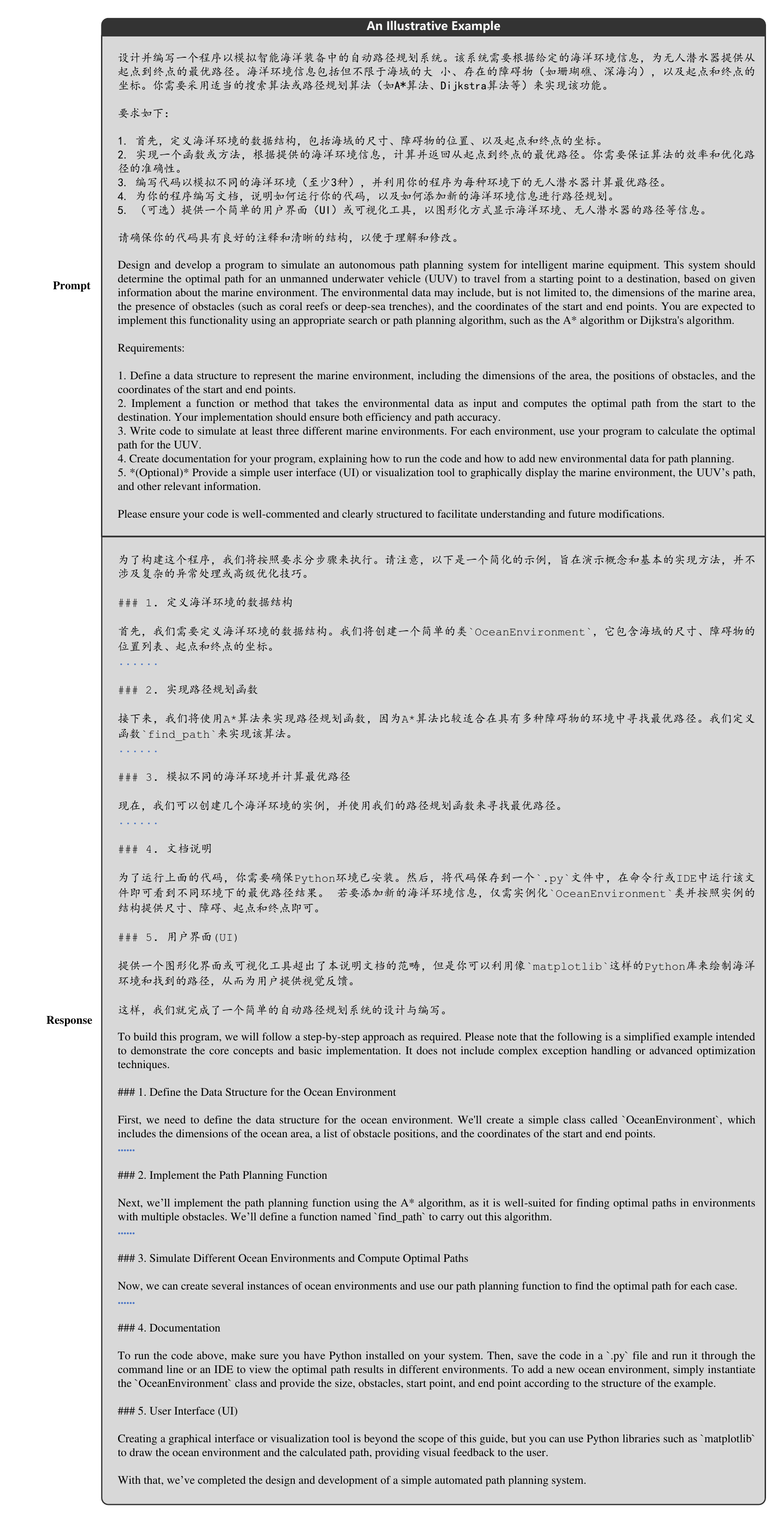}
    \caption{An example prompt and its corresponding response from the preference dataset constructed using TaP. Both the prompt and the response were generated by GPT-4. Portions of the response that were excessively long have been omitted for brevity and are indicated by blue ellipses.}
    \label{fig:illustrative_example}
\end{figure*}

Figure~\ref{fig:illustrative_example} presents an example prompt and its corresponding response from the preference dataset constructed using TaP. Both the prompt and the response were generated by GPT-4.

\section{Preference Dataset Curation}
\label{sec:appendix_dataset_construction}

\subsection{Prompts}
\label{subsec:appendix_prompts}
\begin{figure*}[!t]
    \centering
    \includegraphics[width=\textwidth]{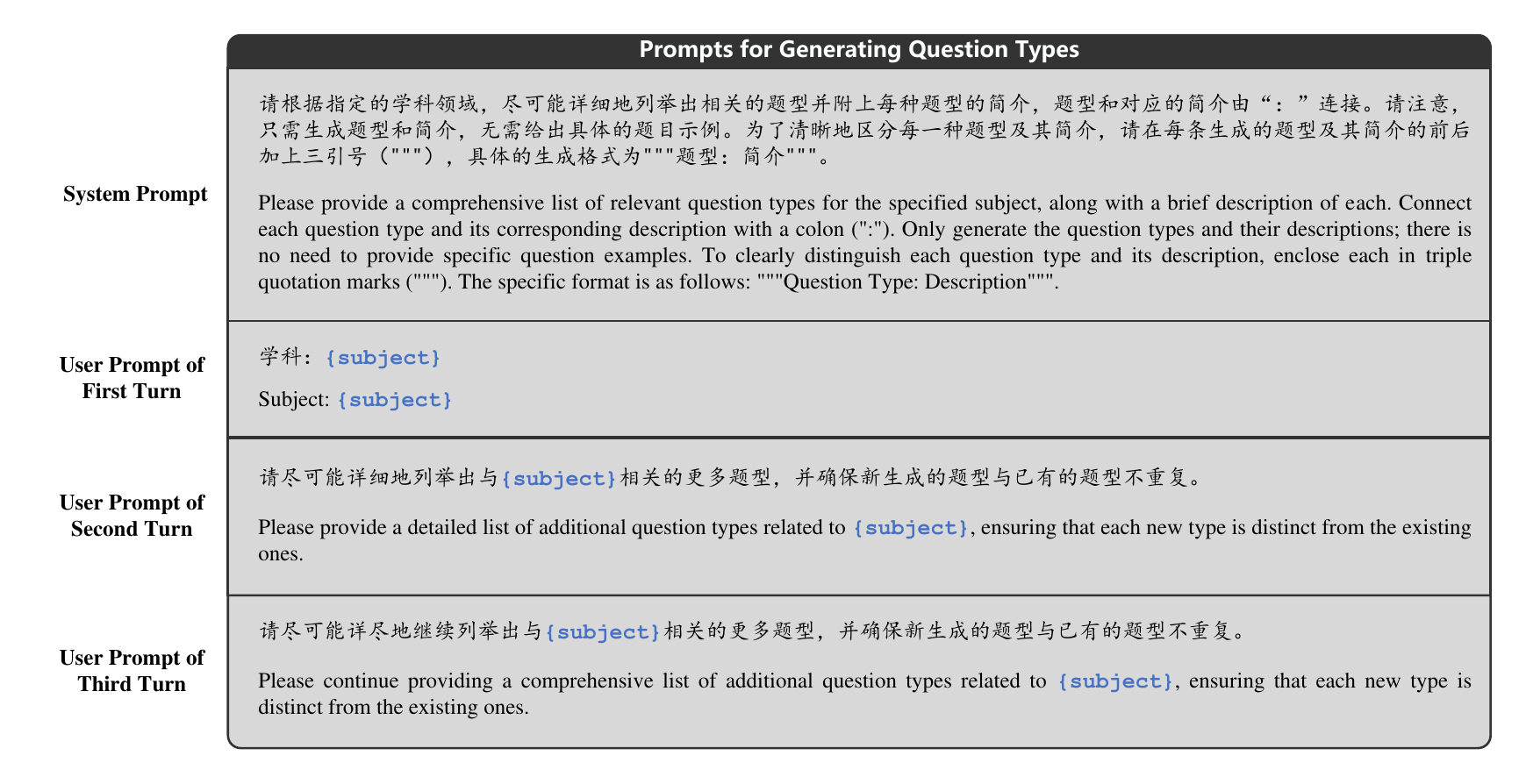}
    \caption{Prompts for generating different question types with brief descriptions of each type.}
    \label{fig:question_type_prompt}
\end{figure*}

\begin{figure*}[!t]
    \centering
    \includegraphics[width=\textwidth]{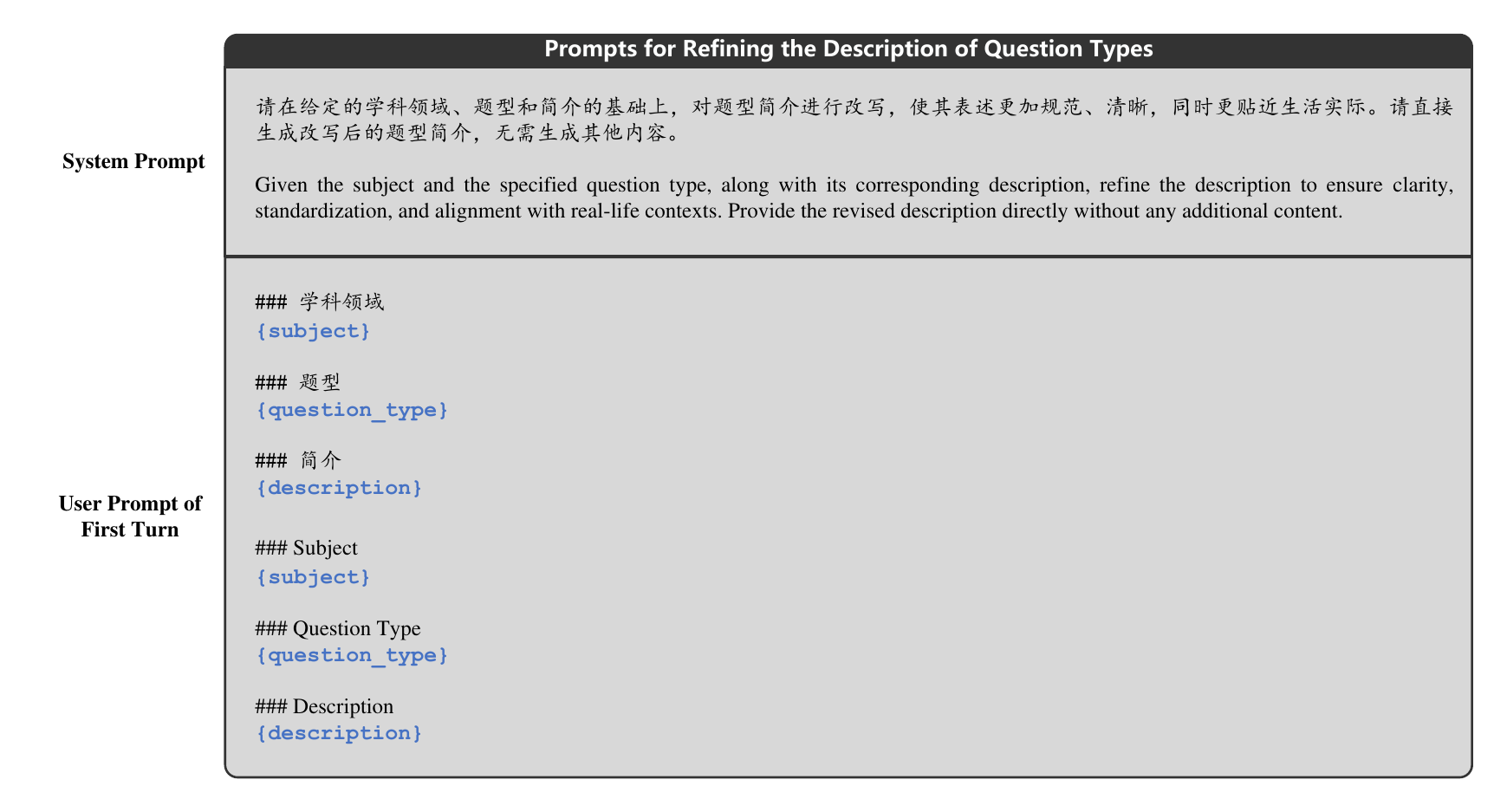}
    \caption{Prompts for refining the description of various question types.}
    \label{fig:description_revise_prompt}
\end{figure*}

\begin{figure*}[!t]
    \centering
    \includegraphics[width=\textwidth]{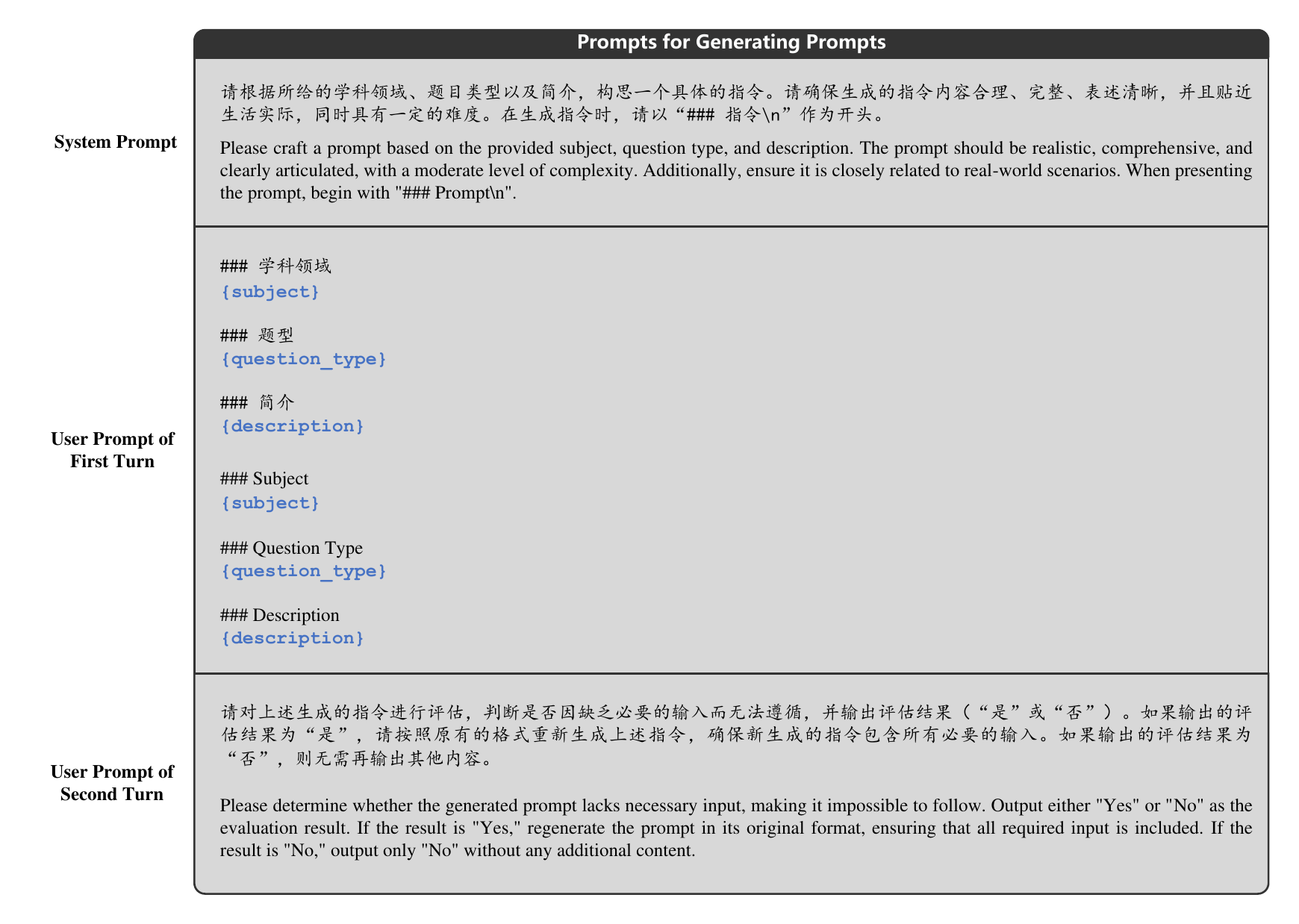}
    \caption{Prompts for generating prompts based on specified question types and their corresponding refined descriptions.}
    \label{fig:prompt_generation_prompt}
\end{figure*}

\begin{figure*}[!t]
    \centering
    \includegraphics[width=\textwidth]{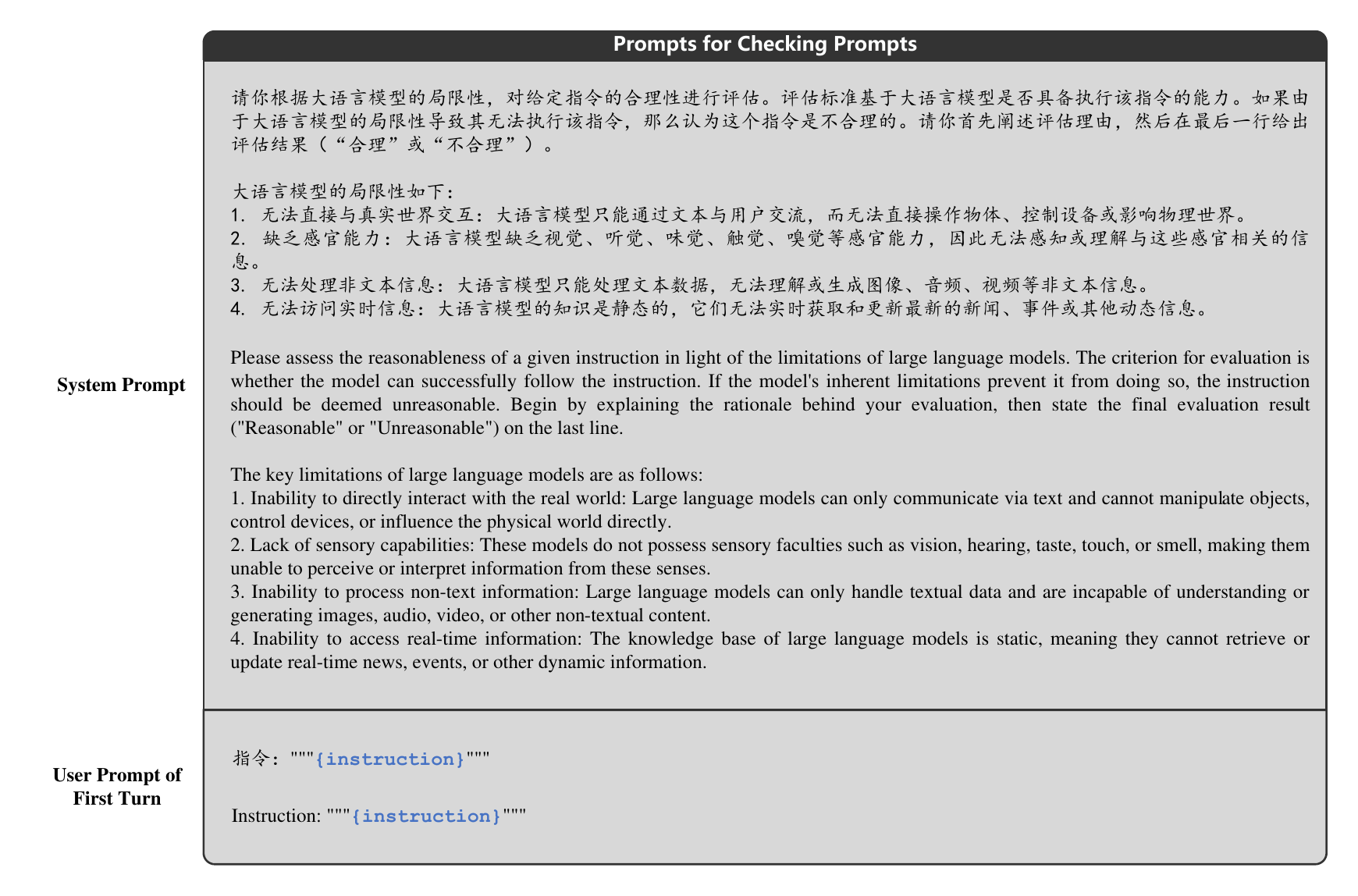}
    \caption{Prompts for evaluating the reasonableness of a given instruction, considering the inherent limitations of LLMs.}
    \label{fig:prompt_check_prompt}
\end{figure*}

\begin{figure*}[!t]
    \centering
    \includegraphics[width=\textwidth]{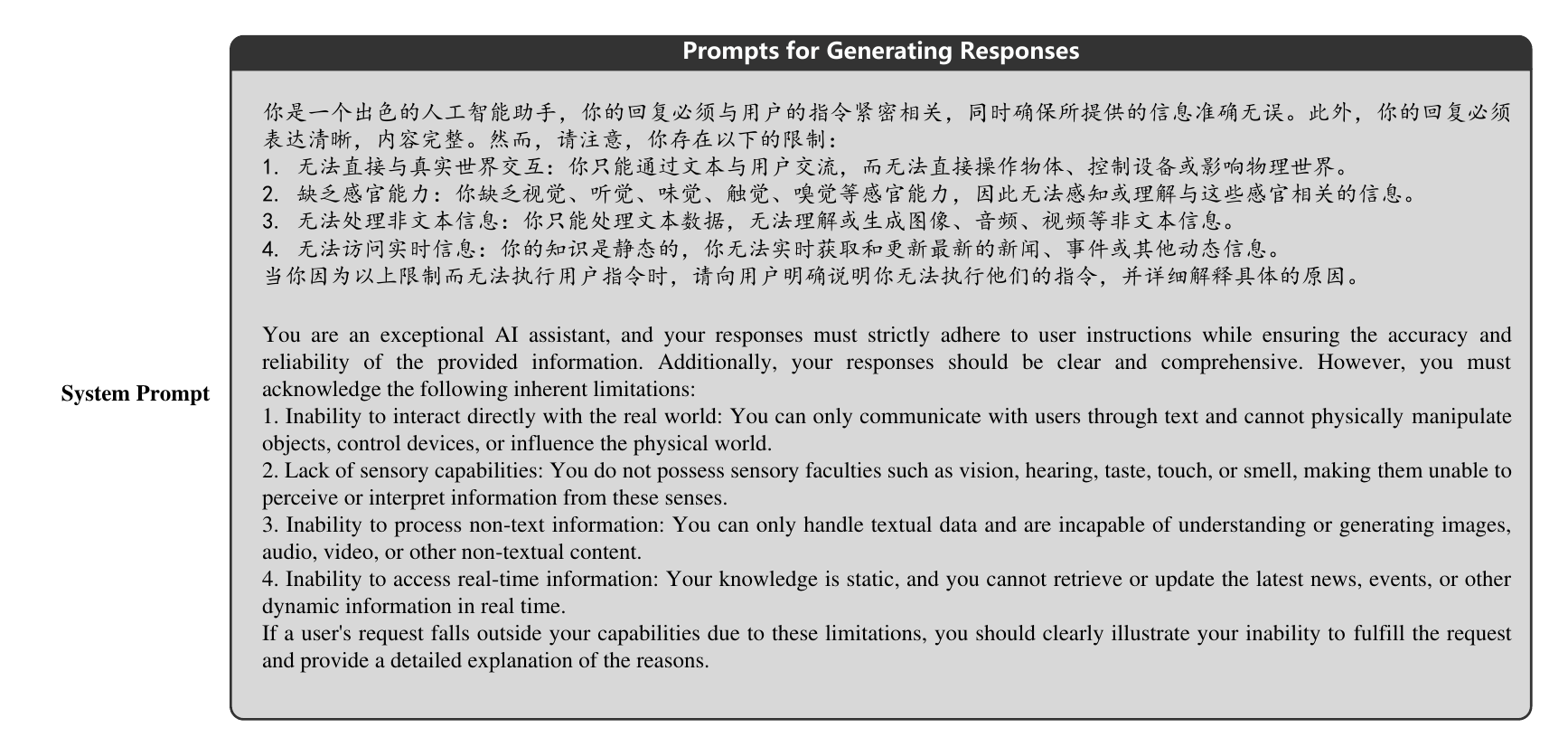}
    \caption{Prompts used for response generation.}
    \label{fig:response_generation_prompt}
\end{figure*}

\begin{figure*}[!t]
    \centering
    \includegraphics[width=\textwidth]{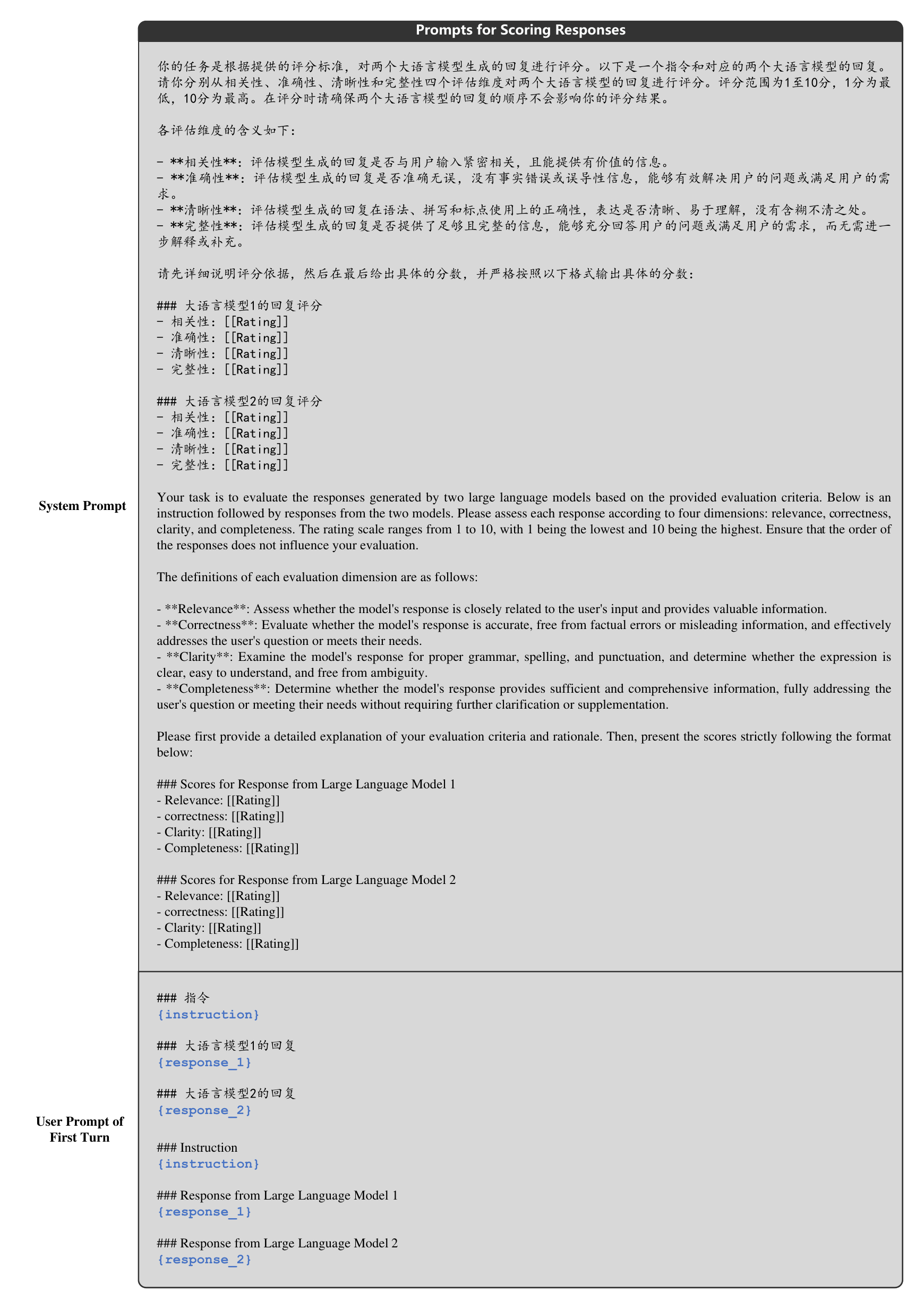}
    \caption{Prompts for conducting pairwise comparisons, where two responses are evaluated on a scale from 1 to 10 across four evaluation dimensions: (1) relevance, (2) correctness, (3) clarity, and (4) completeness.}
    \label{fig:response_eval_prompt}
\end{figure*}

The prompts utilized in the data construction process are depicted in Figures~\ref{fig:question_type_prompt} through~\ref{fig:response_eval_prompt}.

\subsection{Human Evaluation on Feasibility Check}
\label{subsec:appendix_human_evaluation_on_feasibility_check}
To evaluate the reliability of LLMs in identifying instructions that exceed their capabilities, we randomly sampled 100 instructions along with the corresponding assessments generated by GPT-4 indicating whether each instruction was beyond the model's capabilities. Sampling was performed using NumPy with a fixed random seed of 42 to ensure reproducibility.

According to GPT-4's assessments, 71 of the 100 sampled instructions were classified as within the model's capabilities, while the remaining 29 were identified as exceeding them. We then manually verified the correctness of these assessments. Notably, we found that 99 out of the 100 assessments were correct. This result demonstrates GPT-4's high reliability in recognizing its own limitations when explicitly prompted with instructions that incorporate descriptions of its key limitations.

\begin{figure*}[!t]
    \centering
    \includegraphics[width=\textwidth]{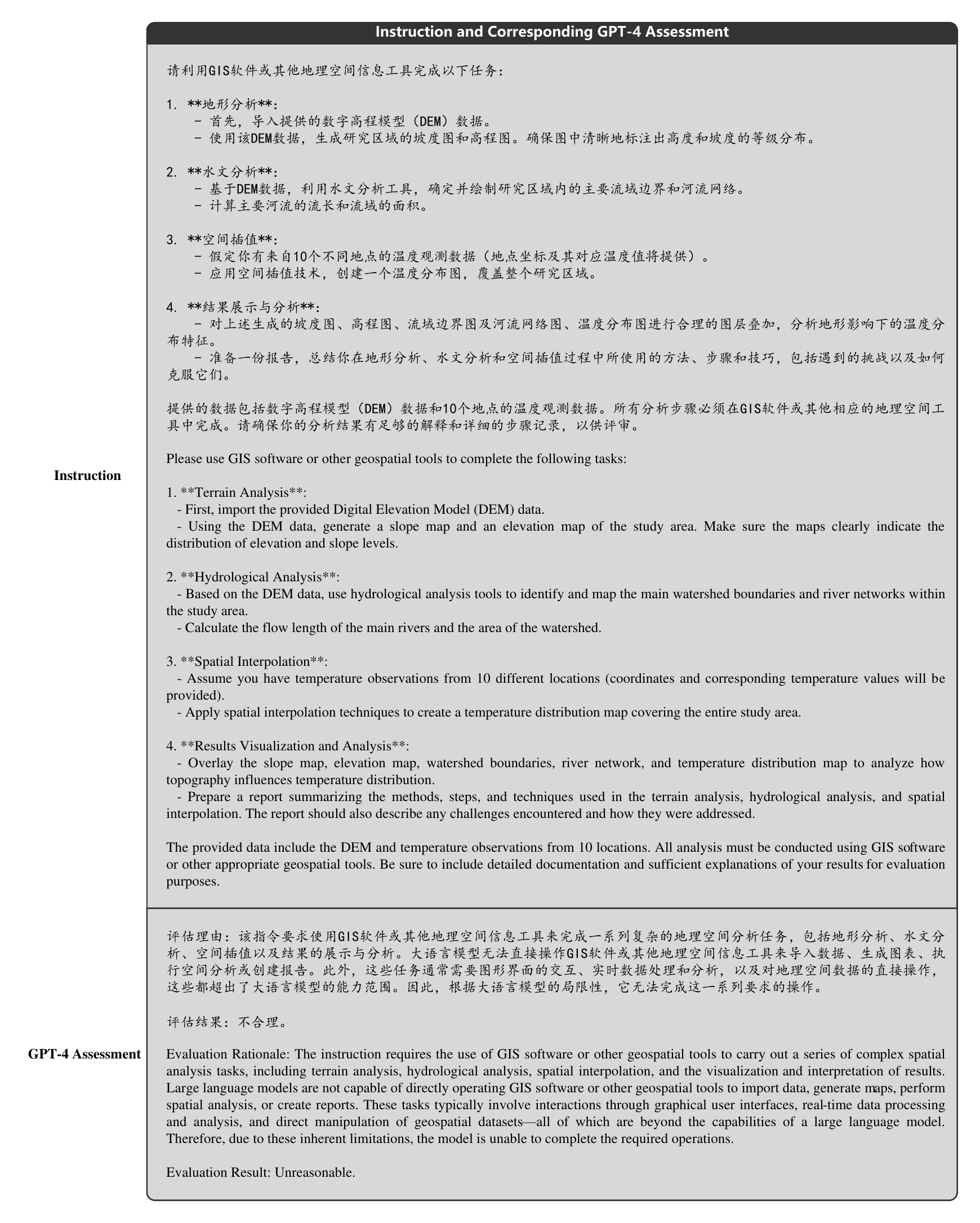}
    \caption{A representative example from the sampled set in which GPT-4 correctly recognizes the instruction as exceeding the capabilities of LLMs. The figure includes both the instruction and GPT-4's corresponding assessment.}
    \label{fig:instruction_and_assessment}
\end{figure*}

For illustrative purposes, Figure~\ref{fig:instruction_and_assessment} presents a representative example from the sampled set, comprising both the instruction and GPT-4's corresponding assessment. In this case, GPT-4 correctly identified the instruction as exceeding the capabilities of LLMs.

\subsection{LLMs for Response Generation}
\label{subsec:appendix_llms_for_response_generation}

To enhance the diversity and coverage of responses generated by the selected models, we aimed to include a broad spectrum of LLM families and architectures. Given the substantial computational and time costs associated with inference across such a large number of models, inference efficiency was prioritized in selecting open-source models. Specifically, aside from proprietary models such as GPT-4, inference efficiency served as the primary criterion for open-source model selection.

To facilitate efficient and scalable inference, we utilized the vLLM library,\footnote{\url{https://github.com/vllm-project/vllm}} which is known for its high throughput and memory efficiency in LLM inference. While vLLM supports a wide range of models, its compatibility is primarily limited to widely adopted models. Consequently, our selection was constrained to models supported by version 0.6.1.post2 of vLLM, which was the version used to generate responses for all prompts.

From the set of models compatible with this version of vLLM, we selected those that met at least one of the following criteria: (1) The model demonstrated exceptional performance across various benchmarks and attained significant popularity in academic and industrial communities (e.g., models from the LLaMA, Qwen, and Gemma families); (2) The model was trained on extensive Chinese corpora, thereby ensuring the generation of high-quality synthetic Chinese data.

Based on these criteria, we selected 63 open-source LLMs. For the remaining two models, GPT-4 (proprietary) and DeepSeek-V2 (unsupported by the vLLM version we used), we employed their official APIs to generate responses. Table~\ref{tab:response_generation_model_list} presents a detailed summary of the architectures and parameter sizes of all LLMs used for response generation.

\section{Experiments}

\subsection{Preference Fine-tuning}
\label{subsec:appendix_preference_finetuning}
For all preference datasets used in our experiments, we randomly split each dataset into two equal subsets, using one for training reward models and the other for PPO and DPO training.

For PPO and DPO training, we initialize the policies using checkpoints from supervised fine-tuning of five open-source pre-trained LLMs. Specifically, we select the best-performing checkpoint trained on two distinct types of training data for each open-source pre-trained LLM, based on their performance on the validation dataset: (1) the highest-performing checkpoint among LLMs trained on seven open-source supervised fine-tuning datasets and (2) the highest-performing checkpoint obtained from an LLM trained on supervised fine-tuning datasets derived from our constructed preference dataset. Consequently, we conduct PPO and DPO training on a total of twelve checkpoints.

For PPO training, we first train the reward model to score the responses generated by the policy. Following common practice, we initialize the reward models from pre-trained LLMs \citep{DBLP:journals/corr/abs-2311-04919,DBLP:conf/icml/CuiY0YH0NXXL0024,DBLP:conf/iclr/KirkMNLHGR24}. Specifically, the reward models are initialized from the same pre-trained LLMs used to initialize the LLMs for supervised fine-tuning, with the exception that the reward model for the policy initialized from the checkpoints derived from supervised fine-tuning of Qwen2.5-3B is initialized from Qwen2.5-14B. Additionally, we constrain the rewards produced by the reward models to a predefined range $\left[ -R, R \right]$ using an auxiliary loss function to stabilize PPO training, where $R$ is a positive constant. Given a prompt $x$, a preferred response $y_{w}$ and less preferred response $y_{l}$, the reward model is parameterized by $\phi$, with $\mathcal{D}$ denoting the preference dataset used for training. The auxiliary loss function is formulated as follows:

\begin{equation}
\begin{aligned}
    \mathcal{L}_{\text{aux}} \left( \phi \right) = & -\mathbb{E}_{\left( x, y_{w}, y_{l} \right) \sim \mathcal{D}} \\
    & \quad \bigl[  \max\left( 0, r_{\phi}\left( x, y_{w} \right) - R \right)^2 \\
    & \: \: \: \quad + \max\left( 0, -r_{\phi}\left( x, y_{w} \right) - R \right)^2 \\
    & \: \: \: \quad + \max\left( 0, r_{\phi}\left( x, y_{l} \right) - R \right)^2 \\
    & \: \: \: \quad + \max\left( 0, -r_{\phi}\left( x, y_{l} \right) - R \right)^2 \bigr]
\end{aligned}
\end{equation}

The auxiliary loss function is incorporated into the overall loss function for training the reward models using a hyperparameter $\lambda$ that adjusts its weight. The resulting loss function for training reward models is defined as follows:

\begin{equation}
\begin{aligned}
    \mathcal{L}_{\text{RM}}\left( \phi \right) = & -\mathbb{E}_{\left( x, y_{w}, y_{l} \right) \sim \mathcal{D}} \bigl[ \lambda \mathcal{L}_{\text{aux}} \left( \phi \right) + \\
    & \: \: \quad \log \sigma \left( r_{\phi}\left( x, y_{w} \right) - r_{\phi}\left( x, y_{l} \right) \right) \bigr]
\end{aligned}
\end{equation}

The hyperparameter $\lambda$ is set to 0.05 when training reward models with our constructed preference dataset. However, due to the relatively smaller size of open-source preference datasets, we reduce $\lambda$ to 0.01 when using them for training. Additionally, the hyperparameter $R$ is fixed at 5, and all reward models are trained for one epoch, which we found sufficient for convergence.

The objective of PPO is to maximize the expected reward of responses generated by the LLM. Let the LLM policy trained using PPO be parameterized by $\theta$ and denoted as $\pi_{\theta}$, and let $\pi_{\text{ref}}$ represent the reference LLM policy. Given a KL reward coefficient $\beta$, the PPO optimization objective is defined as:
\begin{equation}
    \begin{aligned}
        \mathcal{J}\left(\theta\right) = \: & \mathbb{E}_{x \sim \mathcal{D}, y \sim \pi_{\theta}\left(\cdot \vert x \right)} \\
        & \left[ r_{\phi}\left(x, y\right) - \beta \log \left( \frac{\pi_{\theta}\left(y \vert x \right)}{\pi_{\text{ref}}\left(y \vert x \right)} \right)  \right]
    \end{aligned}
\end{equation}

For PPO training, we train the policy for two epochs when using our constructed dataset and for three epochs when using open-source datasets, given their relatively smaller size. We adjust the learning rate according to the \texttt{cosine} schedule with a warm-up phase of 50 steps, using a peak learning rate of 1e-6 for policy models and 5e-6 for critic models. Furthermore, we vary the KL reward coefficient within $\left\{0.01, 0.05\right\}$ based on the KL divergence between the policy and reference policy during PPO training.

For DPO training, we follow a similar strategy to PPO. Specifically, we train the policy for two epochs when using our constructed dataset and for three epochs when using open-source datasets. The learning rate follows the \texttt{cosine} schedule with a warm-up phase of 50 step and a peak learning rate of 2e-7. Following \citet{DBLP:conf/nips/IvisonW0WP0S0H24}, we set $\beta$ to 0.01 across all experiments.

\subsection{Evaluation}
\label{subsec:appendix_evaluation}
For LLMs trained through supervised fine-tuning and DPO, we exclusively use scores from GPT-4o to select the best-performing checkpoint. Conversely, for models trained via supervised learning on \texttt{TaP-SFT (DeepSeek-V2)}, we rely on DeepSeek-V3 scores due to the significant API costs associated with GPT-4o. Similarly, for models trained using PPO, we also utilize DeepSeek-V3 scores for checkpoint selection.

\section{Experimental Results}
\label{sec:appendix_experimental_results}

The detailed experimental results for supervised fine-tuning, PPO, and DPO are reported in Tables~\ref{tab:sft_mt_bench} through \ref{tab:ppo_alignbench_deepseek}.

\begin{table*}[!t]
  \centering
  \small
    \begin{tabular}{lll}
    \toprule
    \textbf{Model Name} & \textbf{\#Parameters} & \textbf{Model Type} \\
    \midrule
    gpt-4-0125-preview & N/A   & N/A \\
    DeepSeek-V2 & 236B (21B) & MoE \\
    Baichuan-13B-Chat & 13B   & Dense \\
    Baichuan2-7B-Chat & 7B    & Dense \\
    Baichuan2-13B-Chat & 13B   & Dense \\
    Qwen1.5-0.5B-Chat & 0.5B  & Dense \\
    Qwen1.5-1.8B-Chat & 1.8B  & Dense \\
    Qwen1.5-4B-Chat & 4B    & Dense \\
    Qwen1.5-7B-Chat & 7B    & Dense \\
    Qwen1.5-14B-Chat & 14B   & Dense \\
    Qwen1.5-32B-Chat & 32B   & Dense \\
    Qwen1.5-72B-Chat & 72B   & Dense \\
    Qwen1.5-110B-Chat & 110B  & Dense \\
    Qwen1.5-MoE-A2.7B-Chat & 14B (3B) & MoE \\
    Qwen2-0.5B-Instruct & 0.5B  & Dense \\
    Qwen2-1.5B-Instruct & 1.5B  & Dense \\
    Qwen2-7B-Instruct & 7B    & Dense \\
    Qwen2-72B-Instruct & 72B   & Dense \\
    Qwen2-57B-A14B-Instruct & 57B (14B) & MoE \\
    Yi-6B-Chat & 6B    & Dense \\
    Yi-34B-Chat & 34B   & Dense \\
    Yi-1.5-6B-Chat & 6B    & Dense \\
    Yi-1.5-9B-Chat & 9B    & Dense \\
    Yi-1.5-34B-Chat & 34B   & Dense \\
    Llama-3-8B-Instruct & 8B    & Dense \\
    Llama-3-70B-Instruct & 70B   & Dense \\
    Llama-3.1-8B-Instruct & 8B    & Dense \\
    Llama-3.1-70B-Instruct & 70B   & Dense \\
    gemma-2b-it & 2B    & Dense \\
    gemma-7b-it & 7B    & Dense \\
    gemma-1.1-2b-it & 2B    & Dense \\
    gemma-1.1-7b-it & 7B    & Dense \\
    gemma-2-2b-it & 2B    & Dense \\
    gemma-2-9b-it & 9B    & Dense \\
    gemma-2-27b-it & 27B   & Dense \\
    Mistral-7B-Instruct-v0.3 & 7B    & Dense \\
    Mistral-Large-Instruct-2407 & 123B  & Dense \\
    Mistral-Nemo-Instruct-2407 & 12B   & Dense \\
    Mixtral-8x22B-Instruct-v0.1 & 141B (39B) & MoE \\
    Mixtral-8x7B-Instruct-v0.1 & 47B (13B) & MoE \\
    c4ai-command-r-v01 & 35B   & Dense \\
    c4ai-command-r-plus & 104B  & Dense \\
    aya-23-8B & 8B    & Dense \\
    aya-23-35B & 35B   & Dense \\
    chatglm3-6b & 6B    & Dense \\
    glm-4-9b-chat & 9B    & Dense \\
    AquilaChat-7B & 7B    & Dense \\
    AquilaChat2-7B & 7B    & Dense \\
    AquilaChat2-34B & 34B   & Dense \\
    dbrx-instruct & 132B (36B) & MoE \\
    internlm-chat-7b & 7B    & Dense \\
    internlm-chat-20b & 20B   & Dense \\
    internlm2-chat-1\_8b & 1.8B  & Dense \\
    internlm2-chat-7b & 7B    & Dense \\
    internlm2-chat-20b & 20B   & Dense \\
    internlm2\_5-1\_8b-chat & 1.8B  & Dense \\
    internlm2\_5-7b-chat & 7B    & Dense \\
    internlm2\_5-20b-chat & 20B   & Dense \\
    MiniCPM-2B-dpo-bf16 & 2B    & Dense \\
    MiniCPM-2B-sft-bf16 & 2B    & Dense \\
    MiniCPM-1B-sft-bf16 & 1B    & Dense \\
    Orion-14B-Chat & 14B   & Dense \\
    XVERSE-7B-Chat & 7B    & Dense \\
    XVERSE-13B-Chat & 13B   & Dense \\
    XVERSE-65B-Chat & 65B   & Dense \\
    \bottomrule
    \end{tabular}%
  \caption{Model architectures and parameter counts of LLMs employed in response generation. Parameter counts enclosed in parentheses represent the number of activated parameters for MoE LLMs.}
  \label{tab:response_generation_model_list}%
\end{table*}%

\begin{table*}[!t]
  \centering
  \tiny
    \begin{tabular}{c|l|ccc|ccc}
    \toprule
    \multirow{2}[4]{*}{\textbf{Model}} & \multicolumn{1}{c|}{\multirow{2}[4]{*}{\textbf{Dataset}}} & \multicolumn{3}{c|}{\textbf{GPT-4o}} & \multicolumn{3}{c}{\textbf{DeepSeek-V3}} \\
\cmidrule{3-8}          &       & \multicolumn{1}{l}{\textbf{First turn}} & \multicolumn{1}{l}{\textbf{Second turn}} & \multicolumn{1}{l|}{\textbf{Average}} & \multicolumn{1}{l}{\textbf{First turn}} & \multicolumn{1}{l}{\textbf{Second turn}} & \multicolumn{1}{l}{\textbf{Average}} \\
    \midrule
    \multirow{9}[2]{*}{Qwen2.5-3B} & \href{https://huggingface.co/datasets/YeungNLP/firefly-train-1.1M}{Firefly} & 5.73  & 3.10  & 4.41  & 5.88  & 3.03  & 4.45  \\
          & \href{https://github.com/Instruction-Tuning-with-GPT-4/GPT-4-LLM/blob/main/data/alpaca_gpt4_data_zh.json}{Alpaca-GPT-4-ZH} \citep{DBLP:journals/corr/abs-2304-03277} & 6.29  & 3.36  & 4.83  & 6.39  & 2.98  & 4.68  \\
          & \href{https://huggingface.co/datasets/BAAI/COIG}{COIG} \citep{DBLP:journals/corr/abs-2304-07987}  & 4.18  & 2.53  & 3.35  & 4.56  & 2.31  & 3.44  \\
          & \href{https://huggingface.co/datasets/fnlp/moss-003-sft-data}{MOSS-SFT} \citep{MIR-2023-12-294} & 6.31  & 3.00  & 4.66  & 6.01  & 2.74  & 4.38  \\
          & \href{https://huggingface.co/datasets/m-a-p/COIG-CQIA}{COIG-CQIA} \citep{DBLP:journals/corr/abs-2403-18058} & 5.99  & 2.95  & 4.47  & 6.01  & 2.69  & 4.35  \\
          & \href{https://huggingface.co/datasets/BAAI/Infinity-Instruct}{Infinity-Instruct} & 6.61  & 3.18  & 4.89  & 6.46  & 2.90  & 4.68  \\
          & \href{https://huggingface.co/datasets/BelleGroup/train_3.5M_CN}{BELLE-SFT} \citep{BELLE} & 6.11  & 3.18  & 4.65  & 6.04  & 2.94  & 4.49  \\
          & \cellcolor[rgb]{ .906,  .902,  .902}TaP-SFT (GPT-4) & \cellcolor[rgb]{ .906,  .902,  .902}\textbf{7.63} & \cellcolor[rgb]{ .906,  .902,  .902}\textbf{3.96} & \cellcolor[rgb]{ .906,  .902,  .902}\textbf{5.79} & \cellcolor[rgb]{ .906,  .902,  .902}7.36  & \cellcolor[rgb]{ .906,  .902,  .902}\textbf{3.69} & \cellcolor[rgb]{ .906,  .902,  .902}5.53  \\
          & \cellcolor[rgb]{ .906,  .902,  .902}TaP-SFT (DeepSeek-V2) & \cellcolor[rgb]{ .906,  .902,  .902}7.56  & \cellcolor[rgb]{ .906,  .902,  .902}3.78  & \cellcolor[rgb]{ .906,  .902,  .902}5.67  & \cellcolor[rgb]{ .906,  .902,  .902}\textbf{7.59} & \cellcolor[rgb]{ .906,  .902,  .902}3.61  & \cellcolor[rgb]{ .906,  .902,  .902}\textbf{5.60} \\
    \midrule
    \multirow{9}[2]{*}{Qwen2.5-7B} & \href{https://huggingface.co/datasets/YeungNLP/firefly-train-1.1M}{Firefly} & 5.75  & 3.10  & 4.43  & 6.09  & 2.85  & 4.47  \\
          & \href{https://github.com/Instruction-Tuning-with-GPT-4/GPT-4-LLM/blob/main/data/alpaca_gpt4_data_zh.json}{Alpaca-GPT-4-ZH} \citep{DBLP:journals/corr/abs-2304-03277} & 6.64  & 3.40  & 5.02  & 6.54  & 3.24  & 4.89  \\
          & \href{https://huggingface.co/datasets/BAAI/COIG}{COIG} \citep{DBLP:journals/corr/abs-2304-07987}  & 4.40  & 2.55  & 3.48  & 4.86  & 2.46  & 3.66  \\
          & \href{https://huggingface.co/datasets/fnlp/moss-003-sft-data}{MOSS-SFT} \citep{MIR-2023-12-294} & 6.94  & 3.18  & 5.06  & 6.95  & 2.85  & 4.90  \\
          & \href{https://huggingface.co/datasets/m-a-p/COIG-CQIA}{COIG-CQIA} \citep{DBLP:journals/corr/abs-2403-18058} & 6.50  & 2.96  & 4.73  & 6.49  & 2.54  & 4.51  \\
          & \href{https://huggingface.co/datasets/BAAI/Infinity-Instruct}{Infinity-Instruct} & 6.61  & 3.31  & 4.96  & 6.96  & 2.94  & 4.95  \\
          & \href{https://huggingface.co/datasets/BelleGroup/train_3.5M_CN}{BELLE-SFT} \citep{BELLE} & 6.36  & 3.50  & 4.93  & 6.44  & 3.23  & 4.83  \\
          & \cellcolor[rgb]{ .906,  .902,  .902}TaP-SFT (GPT-4) & \cellcolor[rgb]{ .906,  .902,  .902}8.01  & \cellcolor[rgb]{ .906,  .902,  .902}\textbf{4.30} & \cellcolor[rgb]{ .906,  .902,  .902}\textbf{6.16} & \cellcolor[rgb]{ .906,  .902,  .902}8.16  & \cellcolor[rgb]{ .906,  .902,  .902}\textbf{3.81} & \cellcolor[rgb]{ .906,  .902,  .902}\textbf{5.99} \\
          & \cellcolor[rgb]{ .906,  .902,  .902}TaP-SFT (DeepSeek-V2) & \cellcolor[rgb]{ .906,  .902,  .902}\textbf{8.39} & \cellcolor[rgb]{ .906,  .902,  .902}3.81  & \cellcolor[rgb]{ .906,  .902,  .902}6.10  & \cellcolor[rgb]{ .906,  .902,  .902}\textbf{8.49} & \cellcolor[rgb]{ .906,  .902,  .902}3.45  & \cellcolor[rgb]{ .906,  .902,  .902}5.97  \\
    \midrule
    \multirow{9}[2]{*}{Llama-3.1-8B} & \href{https://huggingface.co/datasets/YeungNLP/firefly-train-1.1M}{Firefly} & 4.79  & 2.65  & 3.72  & 4.94  & 2.44  & 3.69  \\
          & \href{https://github.com/Instruction-Tuning-with-GPT-4/GPT-4-LLM/blob/main/data/alpaca_gpt4_data_zh.json}{Alpaca-GPT-4-ZH} \citep{DBLP:journals/corr/abs-2304-03277} & 5.46  & 2.95  & 4.21  & 5.66  & 2.90  & 4.28  \\
          & \href{https://huggingface.co/datasets/BAAI/COIG}{COIG} \citep{DBLP:journals/corr/abs-2304-07987}  & 3.39  & 1.98  & 2.68  & 3.63  & 1.86  & 2.74  \\
          & \href{https://huggingface.co/datasets/fnlp/moss-003-sft-data}{MOSS-SFT} \citep{MIR-2023-12-294} & 5.28  & 3.15  & 4.21  & 5.35  & 2.76  & 4.06  \\
          & \href{https://huggingface.co/datasets/m-a-p/COIG-CQIA}{COIG-CQIA} \citep{DBLP:journals/corr/abs-2403-18058} & 4.36  & 2.26  & 3.32  & 4.66  & 2.03  & 3.34  \\
          & \href{https://huggingface.co/datasets/BAAI/Infinity-Instruct}{Infinity-Instruct} & 6.06  & 3.35  & 4.71  & 6.05  & 3.15  & 4.60  \\
          & \href{https://huggingface.co/datasets/BelleGroup/train_3.5M_CN}{BELLE-SFT} \citep{BELLE} & 5.98  & 3.27  & 4.63  & 6.30  & 3.18  & 4.74  \\
          & \cellcolor[rgb]{ .906,  .902,  .902}TaP-SFT (GPT-4) & \cellcolor[rgb]{ .906,  .902,  .902}\textbf{6.80} & \cellcolor[rgb]{ .906,  .902,  .902}3.69  & \cellcolor[rgb]{ .906,  .902,  .902}\textbf{5.24} & \cellcolor[rgb]{ .906,  .902,  .902}\textbf{6.83} & \cellcolor[rgb]{ .906,  .902,  .902}3.08  & \cellcolor[rgb]{ .906,  .902,  .902}\textbf{4.95} \\
          & \cellcolor[rgb]{ .906,  .902,  .902}TaP-SFT (DeepSeek-V2) & \cellcolor[rgb]{ .906,  .902,  .902}6.53  & \cellcolor[rgb]{ .906,  .902,  .902}\textbf{3.70} & \cellcolor[rgb]{ .906,  .902,  .902}5.11  & \cellcolor[rgb]{ .906,  .902,  .902}6.51  & \cellcolor[rgb]{ .906,  .902,  .902}\textbf{3.26} & \cellcolor[rgb]{ .906,  .902,  .902}4.89  \\
    \midrule
    \multirow{9}[2]{*}{Gemma-2-9B} & \href{https://huggingface.co/datasets/YeungNLP/firefly-train-1.1M}{Firefly} & 5.00  & 2.84  & 3.92  & 4.98  & 2.56  & 3.77  \\
          & \href{https://github.com/Instruction-Tuning-with-GPT-4/GPT-4-LLM/blob/main/data/alpaca_gpt4_data_zh.json}{Alpaca-GPT-4-ZH} \citep{DBLP:journals/corr/abs-2304-03277} & 5.56  & 3.21  & 4.39  & 5.51  & 2.66  & 4.09  \\
          & \href{https://huggingface.co/datasets/BAAI/COIG}{COIG} \citep{DBLP:journals/corr/abs-2304-07987}  & 3.35  & 2.03  & 2.69  & 3.49  & 1.81  & 2.65  \\
          & \href{https://huggingface.co/datasets/fnlp/moss-003-sft-data}{MOSS-SFT} \citep{MIR-2023-12-294} & 5.31  & 2.98  & 4.14  & 5.44  & 3.03  & 4.23  \\
          & \href{https://huggingface.co/datasets/m-a-p/COIG-CQIA}{COIG-CQIA} \citep{DBLP:journals/corr/abs-2403-18058} & 4.26  & 2.46  & 3.36  & 4.19  & 2.08  & 3.13  \\
          & \href{https://huggingface.co/datasets/BAAI/Infinity-Instruct}{Infinity-Instruct} & 6.34  & 3.23  & 4.78  & 6.08  & 2.83  & 4.45  \\
          & \href{https://huggingface.co/datasets/BelleGroup/train_3.5M_CN}{BELLE-SFT} \citep{BELLE} & 6.13  & 3.25  & 4.69  & 6.46  & \textbf{3.28} & 4.87  \\
          & \cellcolor[rgb]{ .906,  .902,  .902}TaP-SFT (GPT-4) & \cellcolor[rgb]{ .906,  .902,  .902}\textbf{7.03} & \cellcolor[rgb]{ .906,  .902,  .902}\textbf{3.71} & \cellcolor[rgb]{ .906,  .902,  .902}\textbf{5.37} & \cellcolor[rgb]{ .906,  .902,  .902}\textbf{6.89} & \cellcolor[rgb]{ .906,  .902,  .902}3.16  & \cellcolor[rgb]{ .906,  .902,  .902}\textbf{5.03} \\
          & \cellcolor[rgb]{ .906,  .902,  .902}TaP-SFT (DeepSeek-V2) & \cellcolor[rgb]{ .906,  .902,  .902}6.50  & \cellcolor[rgb]{ .906,  .902,  .902}3.30  & \cellcolor[rgb]{ .906,  .902,  .902}4.90  & \cellcolor[rgb]{ .906,  .902,  .902}6.46  & \cellcolor[rgb]{ .906,  .902,  .902}2.93  & \cellcolor[rgb]{ .906,  .902,  .902}4.69  \\
    \midrule
    \multirow{9}[2]{*}{Qwen2.5-14B} & \href{https://huggingface.co/datasets/YeungNLP/firefly-train-1.1M}{Firefly} & 6.11  & 3.43  & 4.77  & 6.36  & 3.23  & 4.79  \\
          & \href{https://github.com/Instruction-Tuning-with-GPT-4/GPT-4-LLM/blob/main/data/alpaca_gpt4_data_zh.json}{Alpaca-GPT-4-ZH} \citep{DBLP:journals/corr/abs-2304-03277} & 7.14  & 3.75  & 5.44  & 7.01  & 3.71  & 5.36  \\
          & \href{https://huggingface.co/datasets/BAAI/COIG}{COIG} \citep{DBLP:journals/corr/abs-2304-07987}  & 4.64  & 2.63  & 3.63  & 4.69  & 2.43  & 3.56  \\
          & \href{https://huggingface.co/datasets/fnlp/moss-003-sft-data}{MOSS-SFT} \citep{MIR-2023-12-294} & 6.99  & 3.53  & 5.26  & 7.00  & 3.08  & 5.04  \\
          & \href{https://huggingface.co/datasets/m-a-p/COIG-CQIA}{COIG-CQIA} \citep{DBLP:journals/corr/abs-2403-18058} & 6.75  & 3.09  & 4.92  & 6.74  & 2.83  & 4.78  \\
          & \href{https://huggingface.co/datasets/BAAI/Infinity-Instruct}{Infinity-Instruct} & 7.00  & 3.66  & 5.33  & 6.93  & 3.19  & 5.06  \\
          & \href{https://huggingface.co/datasets/BelleGroup/train_3.5M_CN}{BELLE-SFT} \citep{BELLE} & 6.60  & 3.39  & 4.99  & 6.65  & 3.00  & 4.83  \\
          & \cellcolor[rgb]{ .906,  .902,  .902}TaP-SFT (GPT-4) & \cellcolor[rgb]{ .906,  .902,  .902}\textbf{8.29} & \cellcolor[rgb]{ .906,  .902,  .902}\textbf{4.40} & \cellcolor[rgb]{ .906,  .902,  .902}\textbf{6.34} & \cellcolor[rgb]{ .906,  .902,  .902}\textbf{8.31} & \cellcolor[rgb]{ .906,  .902,  .902}4.04  & \cellcolor[rgb]{ .906,  .902,  .902}6.18  \\
          & \cellcolor[rgb]{ .906,  .902,  .902}TaP-SFT (DeepSeek-V2) & \cellcolor[rgb]{ .906,  .902,  .902}8.20  & \cellcolor[rgb]{ .906,  .902,  .902}4.04  & \cellcolor[rgb]{ .906,  .902,  .902}6.12  & \cellcolor[rgb]{ .906,  .902,  .902}8.29  & \cellcolor[rgb]{ .906,  .902,  .902}\textbf{4.09} & \cellcolor[rgb]{ .906,  .902,  .902}\textbf{6.19} \\
    \bottomrule
    \end{tabular}%
  \caption{Performance comparison of five LLMs on MT-Bench-zh after \textbf{supervised fine-tuning} with various datasets. ``GPT-4o'' and ``DeepSeek-V3'' indicate that the responses were evaluated using GPT-4o and DeepSeek-V3, respectively.}
  \label{tab:sft_mt_bench}%
\end{table*}%

\begin{table*}[!t]
  \centering
  \tiny
    \begin{tabular}{c|l|ccc|ccc}
    \toprule
    \multirow{2}[4]{*}{\textbf{Model}} & \multicolumn{1}{c|}{\multirow{2}[4]{*}{\textbf{Dataset}}} & \multicolumn{3}{c|}{\textbf{GPT-4o}} & \multicolumn{3}{c}{\textbf{DeepSeek-V3}} \\
\cmidrule{3-8}          &       & \multicolumn{1}{l}{\textbf{First turn}} & \multicolumn{1}{l}{\textbf{Second turn}} & \multicolumn{1}{l|}{\textbf{Average}} & \multicolumn{1}{l}{\textbf{First turn}} & \multicolumn{1}{l}{\textbf{Second turn}} & \multicolumn{1}{l}{\textbf{Average}} \\
    \midrule
    \multirow{3}[2]{*}{Qwen2.5-3B-SFT-Open} & \href{https://github.com/HIT-SCIR/huozi/tree/main/data/huozi-rlhf}{Huozi-RLHF} \citep{huozi}  & 6.71  & 3.35  & 5.03  & 6.79  & 3.09  & 4.94  \\
          & \href{https://huggingface.co/datasets/wenbopan/Chinese-dpo-pairs}{Chinese-DPO-Pairs} & 6.53  & 3.56  & 5.04  & 6.66  & 3.24  & 4.95  \\
          & \cellcolor[rgb]{ .906,  .902,  .902}TaP (GPT-4) & \cellcolor[rgb]{ .906,  .902,  .902}\textbf{7.69} & \cellcolor[rgb]{ .906,  .902,  .902}\textbf{4.08} & \cellcolor[rgb]{ .906,  .902,  .902}\textbf{5.88} & \cellcolor[rgb]{ .906,  .902,  .902}\textbf{7.76} & \cellcolor[rgb]{ .906,  .902,  .902}\textbf{3.49} & \cellcolor[rgb]{ .906,  .902,  .902}\textbf{5.63} \\
    \midrule
    \multirow{3}[2]{*}{Qwen2.5-3B-SFT-TaP} & \href{https://github.com/HIT-SCIR/huozi/tree/main/data/huozi-rlhf}{Huozi-RLHF} \citep{huozi}  & 7.46  & 3.98  & 5.72  & 7.18  & 3.31  & 5.24  \\
          & \href{https://huggingface.co/datasets/wenbopan/Chinese-dpo-pairs}{Chinese-DPO-Pairs} & 7.63  & \textbf{4.19} & 5.91  & 7.68  & \textbf{3.69} & 5.68  \\
          & \cellcolor[rgb]{ .906,  .902,  .902}TaP (GPT-4) & \cellcolor[rgb]{ .906,  .902,  .902}\textbf{7.94} & \cellcolor[rgb]{ .906,  .902,  .902}4.13  & \cellcolor[rgb]{ .906,  .902,  .902}\textbf{6.03} & \cellcolor[rgb]{ .906,  .902,  .902}\textbf{7.98} & \cellcolor[rgb]{ .906,  .902,  .902}3.60  & \cellcolor[rgb]{ .906,  .902,  .902}\textbf{5.79} \\
    \midrule
    
    \multirow{3}[2]{*}{Qwen2.5-7B-SFT-Open} & \href{https://github.com/HIT-SCIR/huozi/tree/main/data/huozi-rlhf}{Huozi-RLHF} \citep{huozi}  & 7.09  & 3.41  & 5.25  & 6.96  & 3.00  & 4.98  \\
          & \href{https://huggingface.co/datasets/wenbopan/Chinese-dpo-pairs}{Chinese-DPO-Pairs} & 7.09  & 3.36  & 5.23  & 7.26  & \textbf{3.20} & 5.23  \\
          & \cellcolor[rgb]{ .906,  .902,  .902}TaP (GPT-4) & \cellcolor[rgb]{ .906,  .902,  .902}\textbf{7.54} & \cellcolor[rgb]{ .906,  .902,  .902}\textbf{3.55} & \cellcolor[rgb]{ .906,  .902,  .902}\textbf{5.54} & \cellcolor[rgb]{ .906,  .902,  .902}\textbf{7.59} & \cellcolor[rgb]{ .906,  .902,  .902}3.18  & \cellcolor[rgb]{ .906,  .902,  .902}\textbf{5.38} \\
    \midrule
    \multirow{3}[2]{*}{Qwen2.5-7B-SFT-TaP} & \href{https://github.com/HIT-SCIR/huozi/tree/main/data/huozi-rlhf}{Huozi-RLHF} \citep{huozi}  & 8.10  & 4.10  & 6.10  & 7.99  & 3.71  & 5.85  \\
          & \href{https://huggingface.co/datasets/wenbopan/Chinese-dpo-pairs}{Chinese-DPO-Pairs} & 8.04  & 4.29  & 6.16  & \textbf{8.24} & 3.73  & 5.98  \\
          & \cellcolor[rgb]{ .906,  .902,  .902}TaP (GPT-4) & \cellcolor[rgb]{ .906,  .902,  .902}\textbf{8.15} & \cellcolor[rgb]{ .906,  .902,  .902}\textbf{4.58} & \cellcolor[rgb]{ .906,  .902,  .902}\textbf{6.36} & \cellcolor[rgb]{ .906,  .902,  .902}8.23  & \cellcolor[rgb]{ .906,  .902,  .902}\textbf{4.00} & \cellcolor[rgb]{ .906,  .902,  .902}\textbf{6.11} \\
    \midrule
    \multirow{3}[2]{*}{Llama-3.1-8B-SFT-Open} & \href{https://github.com/HIT-SCIR/huozi/tree/main/data/huozi-rlhf}{Huozi-RLHF} \citep{huozi}  & 6.34  & 3.36  & 4.85  & 6.38  & 2.78  & 4.58  \\
          & \href{https://huggingface.co/datasets/wenbopan/Chinese-dpo-pairs}{Chinese-DPO-Pairs} & 6.51  & 3.38  & 4.94  & 6.50  & \textbf{3.25} & 4.88  \\
          & \cellcolor[rgb]{ .906,  .902,  .902}TaP (GPT-4) & \cellcolor[rgb]{ .906,  .902,  .902}\textbf{7.11} & \cellcolor[rgb]{ .906,  .902,  .902}\textbf{3.70} & \cellcolor[rgb]{ .906,  .902,  .902}\textbf{5.41} & \cellcolor[rgb]{ .906,  .902,  .902}\textbf{7.11} & \cellcolor[rgb]{ .906,  .902,  .902}3.16  & \cellcolor[rgb]{ .906,  .902,  .902}\textbf{5.14} \\
    \midrule
    \multirow{3}[2]{*}{Llama-3.1-8B-SFT-TaP} & \href{https://github.com/HIT-SCIR/huozi/tree/main/data/huozi-rlhf}{Huozi-RLHF} \citep{huozi}  & 6.81  & 3.56  & 5.19  & 6.59  & 3.09  & 4.84  \\
          & \href{https://huggingface.co/datasets/wenbopan/Chinese-dpo-pairs}{Chinese-DPO-Pairs} & 6.85  & 3.75  & 5.30  & 6.69  & 2.99  & 4.84  \\
          & \cellcolor[rgb]{ .906,  .902,  .902}TaP (GPT-4) & \cellcolor[rgb]{ .906,  .902,  .902}\textbf{6.99} & \cellcolor[rgb]{ .906,  .902,  .902}\textbf{4.20} & \cellcolor[rgb]{ .906,  .902,  .902}\textbf{5.59} & \cellcolor[rgb]{ .906,  .902,  .902}\textbf{7.00} & \cellcolor[rgb]{ .906,  .902,  .902}\textbf{3.31} & \cellcolor[rgb]{ .906,  .902,  .902}\textbf{5.16} \\
    \midrule
    \multirow{3}[2]{*}{Gemma-2-9B-SFT-Open} & \href{https://github.com/HIT-SCIR/huozi/tree/main/data/huozi-rlhf}{Huozi-RLHF} \citep{huozi}  & 6.21  & 3.33  & 4.77  & 6.14  & 3.10  & 4.62  \\
          & \href{https://huggingface.co/datasets/wenbopan/Chinese-dpo-pairs}{Chinese-DPO-Pairs} & 6.43  & 3.40  & 4.91  & 6.65  & 3.04  & 4.84  \\
          & \cellcolor[rgb]{ .906,  .902,  .902}TaP (GPT-4) & \cellcolor[rgb]{ .906,  .902,  .902}\textbf{7.29} & \cellcolor[rgb]{ .906,  .902,  .902}\textbf{3.79} & \cellcolor[rgb]{ .906,  .902,  .902}\textbf{5.54} & \cellcolor[rgb]{ .906,  .902,  .902}\textbf{7.29} & \cellcolor[rgb]{ .906,  .902,  .902}\textbf{3.18} & \cellcolor[rgb]{ .906,  .902,  .902}\textbf{5.23} \\
    \midrule
    \multirow{3}[2]{*}{Gemma-2-9B-SFT-TaP} & \href{https://github.com/HIT-SCIR/huozi/tree/main/data/huozi-rlhf}{Huozi-RLHF} \citep{huozi}  & 6.95  & 3.84  & 5.39  & 6.81  & 3.21  & 5.01  \\
          & \href{https://huggingface.co/datasets/wenbopan/Chinese-dpo-pairs}{Chinese-DPO-Pairs} & 7.00  & 4.09  & 5.54  & 7.00  & \textbf{3.40} & 5.20  \\
          & \cellcolor[rgb]{ .906,  .902,  .902}TaP (GPT-4) & \cellcolor[rgb]{ .906,  .902,  .902}\textbf{7.44} & \cellcolor[rgb]{ .906,  .902,  .902}\textbf{4.13} & \cellcolor[rgb]{ .906,  .902,  .902}\textbf{5.78} & \cellcolor[rgb]{ .906,  .902,  .902}\textbf{7.49} & \cellcolor[rgb]{ .906,  .902,  .902}3.16  & \cellcolor[rgb]{ .906,  .902,  .902}\textbf{5.33} \\
    \midrule
    \multirow{3}[2]{*}{Qwen2.5-14B-SFT-Open} & \href{https://github.com/HIT-SCIR/huozi/tree/main/data/huozi-rlhf}{Huozi-RLHF} \citep{huozi}  & 7.16  & 3.66  & 5.41  & 7.50  & 3.46  & 5.48  \\
          & \href{https://huggingface.co/datasets/wenbopan/Chinese-dpo-pairs}{Chinese-DPO-Pairs} & 7.36  & 3.71  & 5.54  & 7.64  & 3.53  & 5.58  \\
          & \cellcolor[rgb]{ .906,  .902,  .902}TaP (GPT-4) & \cellcolor[rgb]{ .906,  .902,  .902}\textbf{7.64} & \cellcolor[rgb]{ .906,  .902,  .902}\textbf{4.05} & \cellcolor[rgb]{ .906,  .902,  .902}\textbf{5.84} & \cellcolor[rgb]{ .906,  .902,  .902}\textbf{7.81} & \cellcolor[rgb]{ .906,  .902,  .902}\textbf{3.96} & \cellcolor[rgb]{ .906,  .902,  .902}\textbf{5.89} \\
    \midrule
    \multirow{3}[2]{*}{Qwen2.5-14B-SFT-TaP} & \href{https://github.com/HIT-SCIR/huozi/tree/main/data/huozi-rlhf}{Huozi-RLHF} \citep{huozi}  & 8.21  & 4.29  & 6.25  & 8.30  & 3.93  & 6.11  \\
          & \href{https://huggingface.co/datasets/wenbopan/Chinese-dpo-pairs}{Chinese-DPO-Pairs} & 8.31  & 4.38  & 6.34  & 8.33  & 3.95  & 6.14  \\
          & \cellcolor[rgb]{ .906,  .902,  .902}TaP (GPT-4) & \cellcolor[rgb]{ .906,  .902,  .902}\textbf{8.50} & \cellcolor[rgb]{ .906,  .902,  .902}\textbf{4.83} & \cellcolor[rgb]{ .906,  .902,  .902}\textbf{6.66} & \cellcolor[rgb]{ .906,  .902,  .902}\textbf{8.54} & \cellcolor[rgb]{ .906,  .902,  .902}\textbf{4.16} & \cellcolor[rgb]{ .906,  .902,  .902}\textbf{6.35} \\
    \bottomrule
    \end{tabular}%
  \caption{Performance comparison of LLMs trained with \textbf{DPO} using different datasets on MT-Bench-zh. The model names include two possible suffixes: ``Open'' and ``TaP.'' The ``Open'' suffix indicates that the LLMs were initialized from models trained via supervised fine-tuning on open-source datasets, whereas ``TaP'' denotes initialization from models trained on a dataset constructed by TaP. Additionally, ``GPT-4o'' and ``DeepSeek-V3'' specify that responses were evaluated using GPT-4o and DeepSeek-V3, respectively.}
  \label{tab:dpo_mt_bench}%
\end{table*}%

\begin{table*}[!t]
  \centering
  \tiny
    \begin{tabular}{c|l|ccc|ccc}
    \toprule
    \multirow{2}[4]{*}{\textbf{Model}} & \multicolumn{1}{c|}{\multirow{2}[4]{*}{\textbf{Dataset}}} & \multicolumn{3}{c|}{\textbf{GPT-4o}} & \multicolumn{3}{c}{\textbf{DeepSeek-V3}} \\
\cmidrule{3-8}          &       & \multicolumn{1}{l}{\textbf{First turn}} & \multicolumn{1}{l}{\textbf{Second turn}} & \multicolumn{1}{l|}{\textbf{Average}} & \multicolumn{1}{l}{\textbf{First turn}} & \multicolumn{1}{l}{\textbf{Second turn}} & \multicolumn{1}{l}{\textbf{Average}} \\
    \midrule
    \multirow{3}[2]{*}{Qwen2.5-3B-SFT-Open} & \href{https://github.com/HIT-SCIR/huozi/tree/main/data/huozi-rlhf}{Huozi-RLHF} \citep{huozi}  & 6.86  & 3.64  & 5.25  & 6.91  & 3.49  & 5.20  \\
          & \href{https://huggingface.co/datasets/wenbopan/Chinese-dpo-pairs}{Chinese-DPO-Pairs} & \textbf{7.86} & \textbf{4.13} & \textbf{5.99} & \textbf{7.78} & 3.65  & \textbf{5.71} \\
          & \cellcolor[rgb]{ .906,  .902,  .902}TaP (GPT-4) & \cellcolor[rgb]{ .906,  .902,  .902}7.84  & \cellcolor[rgb]{ .906,  .902,  .902}3.76  & \cellcolor[rgb]{ .906,  .902,  .902}5.80  & \cellcolor[rgb]{ .906,  .902,  .902}7.70  & \cellcolor[rgb]{ .906,  .902,  .902}\textbf{3.66} & \cellcolor[rgb]{ .906,  .902,  .902}5.68  \\
    \midrule
    \multirow{3}[2]{*}{Qwen2.5-3B-SFT-TaP} & \href{https://github.com/HIT-SCIR/huozi/tree/main/data/huozi-rlhf}{Huozi-RLHF} \citep{huozi}  & 7.51  & 4.21  & 5.86  & 7.65  & \textbf{3.86} & 5.76  \\
          & \href{https://huggingface.co/datasets/wenbopan/Chinese-dpo-pairs}{Chinese-DPO-Pairs} & 7.55  & \textbf{4.60} & \textbf{6.08} & 7.79  & 3.80  & 5.79  \\
          & \cellcolor[rgb]{ .906,  .902,  .902}TaP (GPT-4) & \cellcolor[rgb]{ .906,  .902,  .902}\textbf{7.80} & \cellcolor[rgb]{ .906,  .902,  .902}4.20  & \cellcolor[rgb]{ .906,  .902,  .902}6.00  & \cellcolor[rgb]{ .906,  .902,  .902}\textbf{7.99} & \cellcolor[rgb]{ .906,  .902,  .902}3.64  & \cellcolor[rgb]{ .906,  .902,  .902}\textbf{5.81} \\
    \midrule
    
    \multirow{3}[2]{*}{Qwen2.5-7B-SFT-Open} & \href{https://github.com/HIT-SCIR/huozi/tree/main/data/huozi-rlhf}{Huozi-RLHF} \citep{huozi}  & 7.05  & 3.36  & 5.21  & 7.21  & 3.18  & 5.19  \\
          & \href{https://huggingface.co/datasets/wenbopan/Chinese-dpo-pairs}{Chinese-DPO-Pairs} & 7.10  & 3.36  & 5.23  & 7.03  & 3.25  & 5.14  \\
          & \cellcolor[rgb]{ .906,  .902,  .902}TaP (GPT-4) & \cellcolor[rgb]{ .906,  .902,  .902}\textbf{7.31} & \cellcolor[rgb]{ .906,  .902,  .902}\textbf{3.46} & \cellcolor[rgb]{ .906,  .902,  .902}\textbf{5.39} & \cellcolor[rgb]{ .906,  .902,  .902}\textbf{7.56} & \cellcolor[rgb]{ .906,  .902,  .902}\textbf{3.41} & \cellcolor[rgb]{ .906,  .902,  .902}\textbf{5.49} \\
    \midrule
    \multirow{3}[2]{*}{Qwen2.5-7B-SFT-TaP} & \href{https://github.com/HIT-SCIR/huozi/tree/main/data/huozi-rlhf}{Huozi-RLHF} \citep{huozi}  & 8.15  & \textbf{4.35} & 6.25  & 8.35  & 4.03  & 6.19  \\
          & \href{https://huggingface.co/datasets/wenbopan/Chinese-dpo-pairs}{Chinese-DPO-Pairs} & 8.36  & 4.34  & 6.35  & 8.45  & 4.18  & 6.31  \\
          & \cellcolor[rgb]{ .906,  .902,  .902}TaP (GPT-4) & \cellcolor[rgb]{ .906,  .902,  .902}\textbf{8.54} & \cellcolor[rgb]{ .906,  .902,  .902}4.19  & \cellcolor[rgb]{ .906,  .902,  .902}\textbf{6.36} & \cellcolor[rgb]{ .906,  .902,  .902}\textbf{8.55} & \cellcolor[rgb]{ .906,  .902,  .902}\textbf{4.21} & \cellcolor[rgb]{ .906,  .902,  .902}\textbf{6.38} \\
    \midrule
    \multirow{3}[2]{*}{Llama-3.1-8B-SFT-Open} & \href{https://github.com/HIT-SCIR/huozi/tree/main/data/huozi-rlhf}{Huozi-RLHF} \citep{huozi}  & 6.34  & \textbf{3.53} & 4.93  & 6.46  & \textbf{3.33} & 4.89  \\
          & \href{https://huggingface.co/datasets/wenbopan/Chinese-dpo-pairs}{Chinese-DPO-Pairs} & 6.25  & 3.24  & 4.75  & 6.64  & 3.21  & 4.93  \\
          & \cellcolor[rgb]{ .906,  .902,  .902}TaP (GPT-4) & \cellcolor[rgb]{ .906,  .902,  .902}\textbf{6.88} & \cellcolor[rgb]{ .906,  .902,  .902}3.41  & \cellcolor[rgb]{ .906,  .902,  .902}\textbf{5.14} & \cellcolor[rgb]{ .906,  .902,  .902}\textbf{6.79} & \cellcolor[rgb]{ .906,  .902,  .902}3.10  & \cellcolor[rgb]{ .906,  .902,  .902}\textbf{4.94} \\
    \midrule
    \multirow{3}[2]{*}{Llama-3.1-8B-SFT-TaP} & \href{https://github.com/HIT-SCIR/huozi/tree/main/data/huozi-rlhf}{Huozi-RLHF} \citep{huozi}  & 6.61  & 3.89  & 5.25  & 6.85  & 3.24  & 5.04  \\
          & \href{https://huggingface.co/datasets/wenbopan/Chinese-dpo-pairs}{Chinese-DPO-Pairs} & 6.79  & 3.63  & 5.21  & \textbf{7.01} & 3.08  & 5.04  \\
          & \cellcolor[rgb]{ .906,  .902,  .902}TaP (GPT-4) & \cellcolor[rgb]{ .906,  .902,  .902}\textbf{6.94} & \cellcolor[rgb]{ .906,  .902,  .902}\textbf{3.99} & \cellcolor[rgb]{ .906,  .902,  .902}\textbf{5.46} & \cellcolor[rgb]{ .906,  .902,  .902}6.99  & \cellcolor[rgb]{ .906,  .902,  .902}\textbf{3.58} & \cellcolor[rgb]{ .906,  .902,  .902}\textbf{5.28} \\
    \midrule
    \multirow{3}[2]{*}{Gemma-2-9B-SFT-Open} & \href{https://github.com/HIT-SCIR/huozi/tree/main/data/huozi-rlhf}{Huozi-RLHF} \citep{huozi}  & 6.41  & 3.33  & 4.87  & 6.49  & 3.21  & 4.85  \\
          & \href{https://huggingface.co/datasets/wenbopan/Chinese-dpo-pairs}{Chinese-DPO-Pairs} & 6.58  & 3.48  & 5.03  & 6.64  & 3.34  & 4.99  \\
          & \cellcolor[rgb]{ .906,  .902,  .902}TaP (GPT-4) & \cellcolor[rgb]{ .906,  .902,  .902}\textbf{7.49} & \cellcolor[rgb]{ .906,  .902,  .902}\textbf{3.55} & \cellcolor[rgb]{ .906,  .902,  .902}\textbf{5.52} & \cellcolor[rgb]{ .906,  .902,  .902}\textbf{7.53} & \cellcolor[rgb]{ .906,  .902,  .902}\textbf{3.48} & \cellcolor[rgb]{ .906,  .902,  .902}\textbf{5.50} \\
    \midrule
    \multirow{3}[2]{*}{Gemma-2-9B-SFT-TaP} & \href{https://github.com/HIT-SCIR/huozi/tree/main/data/huozi-rlhf}{Huozi-RLHF} \citep{huozi}  & 6.85  & 3.69  & 5.27  & 7.05  & 3.16  & 5.11  \\
          & \href{https://huggingface.co/datasets/wenbopan/Chinese-dpo-pairs}{Chinese-DPO-Pairs} & 6.93  & 3.98  & 5.45  & 7.04  & \textbf{3.33} & 5.18  \\
          & \cellcolor[rgb]{ .906,  .902,  .902}TaP (GPT-4) & \cellcolor[rgb]{ .906,  .902,  .902}\textbf{7.25} & \cellcolor[rgb]{ .906,  .902,  .902}\textbf{4.01} & \cellcolor[rgb]{ .906,  .902,  .902}\textbf{5.63} & \cellcolor[rgb]{ .906,  .902,  .902}\textbf{7.29} & \cellcolor[rgb]{ .906,  .902,  .902}3.23  & \cellcolor[rgb]{ .906,  .902,  .902}\textbf{5.26} \\
    \midrule
    \multirow{3}[2]{*}{Qwen2.5-14B-SFT-Open} & \href{https://github.com/HIT-SCIR/huozi/tree/main/data/huozi-rlhf}{Huozi-RLHF} \citep{huozi}  & 7.59  & 3.98  & 5.78  & 7.80  & 3.76  & 5.78  \\
          & \href{https://huggingface.co/datasets/wenbopan/Chinese-dpo-pairs}{Chinese-DPO-Pairs} & 7.80  & 3.93  & 5.86  & 7.81  & 3.74  & 5.78  \\
          & \cellcolor[rgb]{ .906,  .902,  .902}TaP (GPT-4) & \cellcolor[rgb]{ .906,  .902,  .902}\textbf{8.19} & \cellcolor[rgb]{ .906,  .902,  .902}\textbf{4.45} & \cellcolor[rgb]{ .906,  .902,  .902}\textbf{6.32} & \cellcolor[rgb]{ .906,  .902,  .902}\textbf{8.29} & \cellcolor[rgb]{ .906,  .902,  .902}\textbf{4.14} & \cellcolor[rgb]{ .906,  .902,  .902}\textbf{6.21} \\
    \midrule
    \multirow{3}[2]{*}{Qwen2.5-14B-SFT-TaP} & \href{https://github.com/HIT-SCIR/huozi/tree/main/data/huozi-rlhf}{Huozi-RLHF} \citep{huozi}  & 8.29  & 4.54  & 6.41  & 8.60  & \textbf{4.43} & \textbf{6.51} \\
          & \href{https://huggingface.co/datasets/wenbopan/Chinese-dpo-pairs}{Chinese-DPO-Pairs} & \textbf{8.86} & \textbf{4.80} & \textbf{6.83} & 8.75  & 4.16  & 6.46  \\
          & \cellcolor[rgb]{ .906,  .902,  .902}TaP (GPT-4) & \cellcolor[rgb]{ .906,  .902,  .902}8.50  & \cellcolor[rgb]{ .906,  .902,  .902}4.66  & \cellcolor[rgb]{ .906,  .902,  .902}6.58  & \cellcolor[rgb]{ .906,  .902,  .902}\textbf{8.83} & \cellcolor[rgb]{ .906,  .902,  .902}4.04  & \cellcolor[rgb]{ .906,  .902,  .902}6.43  \\
    \bottomrule
    \end{tabular}%
  \caption{Performance comparison of LLMs trained with \textbf{PPO} using different datasets on MT-Bench-zh. The model names include two possible suffixes: ``Open'' and ``TaP.'' The ``Open'' suffix indicates that the LLMs were initialized from models trained via supervised fine-tuning on open-source datasets, whereas ``TaP'' denotes initialization from models trained on a dataset constructed by TaP. Additionally, ``GPT-4o'' and ``DeepSeek-V3'' specify that responses were evaluated using GPT-4o and DeepSeek-V3, respectively.}
  \label{tab:ppo_mt_bench}%
\end{table*}%

\begin{table*}[!t]
  \centering
  \tiny
    \begin{tabular}{c|l|ccc|ccccccc|c}
    \toprule
    \multirow{2}[4]{*}{\textbf{Model}} & \multicolumn{1}{c|}{\multirow{2}[4]{*}{\textbf{Dataset}}} & \multicolumn{3}{c|}{\textbf{Reasoning}} & \multicolumn{7}{c|}{\textbf{Language}}                & \multicolumn{1}{c}{\multirow{2}[4]{*}{\textbf{Overall}}} \\
\cmidrule{3-12}          &       & \multicolumn{1}{c}{\textbf{Math.}} & \multicolumn{1}{c}{\textbf{Logi.}} & \multicolumn{1}{c|}{\textbf{Avg.}} & \multicolumn{1}{c}{\textbf{Pro.}} & \multicolumn{1}{c}{\textbf{Chi.}} & \multicolumn{1}{c}{\textbf{Fund.}} & \multicolumn{1}{c}{\textbf{Writ.}} & \multicolumn{1}{c}{\textbf{Open.}} & \multicolumn{1}{c}{\textbf{Role.}} & \multicolumn{1}{c|}{\textbf{Avg.}} &  \\
    \midrule
    \multirow{9}[2]{*}{Qwen2.5-3B} & \href{https://huggingface.co/datasets/YeungNLP/firefly-train-1.1M}{Firefly} & 3.59  & 2.86  & 3.22  & 4.56  & 4.60  & 4.07  & 4.80  & 3.79  & 4.19  & 4.34  & 3.78  \\
          & \href{https://github.com/Instruction-Tuning-with-GPT-4/GPT-4-LLM/blob/main/data/alpaca_gpt4_data_zh.json}{Alpaca-GPT-4-ZH} \citep{DBLP:journals/corr/abs-2304-03277} & 4.67  & 3.65  & 4.16  & 4.86  & 4.61  & 4.84  & 4.77  & 5.42  & 4.69  & 4.87  & 4.51  \\
          & \href{https://huggingface.co/datasets/BAAI/COIG}{COIG} \citep{DBLP:journals/corr/abs-2304-07987}  & 2.96  & 2.49  & 2.73  & 4.02  & 4.44  & 3.68  & 3.84  & 4.61  & 4.16  & 4.12  & 3.42  \\
          & \href{https://huggingface.co/datasets/fnlp/moss-003-sft-data}{MOSS-SFT} \citep{MIR-2023-12-294} & 4.30  & 3.34  & 3.82  & 4.96  & 4.60  & 4.99  & 4.99  & 5.11  & 5.09  & 4.96  & 4.39  \\
          & \href{https://huggingface.co/datasets/m-a-p/COIG-CQIA}{COIG-CQIA} \citep{DBLP:journals/corr/abs-2403-18058} & 3.79  & 3.42  & 3.61  & 4.15  & 4.28  & 4.51  & 4.57  & 4.21  & 4.56  & 4.38  & 3.99  \\
          & \href{https://huggingface.co/datasets/BAAI/Infinity-Instruct}{Infinity-Instruct} & 4.56  & 3.48  & 4.02  & 5.10  & 4.52  & 4.74  & 5.44  & 5.26  & 5.63  & 5.11  & 4.57  \\
          & \href{https://huggingface.co/datasets/BelleGroup/train_3.5M_CN}{BELLE-SFT} \citep{BELLE} & 4.60  & 3.16  & 3.88  & 5.00  & 4.76  & 4.97  & 5.28  & 5.34  & 5.28  & 5.11  & 4.49  \\
          & \cellcolor[rgb]{ .906,  .902,  .902}TGPDG-SFT (GPT-4) & \cellcolor[rgb]{ .906,  .902,  .902}5.92  & \cellcolor[rgb]{ .906,  .902,  .902}4.05  & \cellcolor[rgb]{ .906,  .902,  .902}4.99  & \cellcolor[rgb]{ .906,  .902,  .902}\textbf{5.59} & \cellcolor[rgb]{ .906,  .902,  .902}\textbf{5.50} & \cellcolor[rgb]{ .906,  .902,  .902}\textbf{5.26} & \cellcolor[rgb]{ .906,  .902,  .902}\textbf{6.28} & \cellcolor[rgb]{ .906,  .902,  .902}\textbf{6.39 } & \cellcolor[rgb]{ .906,  .902,  .902}\textbf{6.07} & \cellcolor[rgb]{ .906,  .902,  .902}\textbf{5.85} & \cellcolor[rgb]{ .906,  .902,  .902}5.42  \\
          & \cellcolor[rgb]{ .906,  .902,  .902}TGPDG-SFT (DeepSeek-V2) & \cellcolor[rgb]{ .906,  .902,  .902}\textbf{6.03} & \cellcolor[rgb]{ .906,  .902,  .902}\textbf{4.48} & \cellcolor[rgb]{ .906,  .902,  .902}\textbf{5.25} & \cellcolor[rgb]{ .906,  .902,  .902}5.48  & \cellcolor[rgb]{ .906,  .902,  .902}5.43  & \cellcolor[rgb]{ .906,  .902,  .902}5.25  & \cellcolor[rgb]{ .906,  .902,  .902}5.92  & \cellcolor[rgb]{ .906,  .902,  .902}6.13  & \cellcolor[rgb]{ .906,  .902,  .902}5.70  & \cellcolor[rgb]{ .906,  .902,  .902}5.65  & \cellcolor[rgb]{ .906,  .902,  .902}\textbf{5.45} \\
    \midrule
    
    \multirow{9}[2]{*}{Qwen2.5-7B} & \href{https://huggingface.co/datasets/YeungNLP/firefly-train-1.1M}{Firefly} & 3.66  & 3.24  & 3.45  & 4.72  & 4.31  & 4.97  & 5.07  & 3.84  & 4.78  & 4.61  & 4.03  \\
          & \href{https://github.com/Instruction-Tuning-with-GPT-4/GPT-4-LLM/blob/main/data/alpaca_gpt4_data_zh.json}{Alpaca-GPT-4-ZH} \citep{DBLP:journals/corr/abs-2304-03277} & 5.49  & 4.30  & 4.90  & 5.50  & 5.21  & 5.44  & 5.24  & 5.13  & 5.34  & 5.31  & 5.10  \\
          & \href{https://huggingface.co/datasets/BAAI/COIG}{COIG} \citep{DBLP:journals/corr/abs-2304-07987}  & 3.04  & 2.61  & 2.82  & 4.65  & 4.68  & 4.10  & 4.01  & 4.55  & 4.45  & 4.41  & 3.62  \\
          & \href{https://huggingface.co/datasets/fnlp/moss-003-sft-data}{MOSS-SFT} \citep{MIR-2023-12-294} & 5.10  & 4.32  & 4.71  & 5.46  & 4.78  & 4.87  & 5.35  & 5.68  & 5.41  & 5.26  & 4.98  \\
          & \href{https://huggingface.co/datasets/m-a-p/COIG-CQIA}{COIG-CQIA} \citep{DBLP:journals/corr/abs-2403-18058} & 5.26  & 4.17  & 4.72  & 5.10  & 5.35  & 5.07  & 5.15  & 5.13  & 4.96  & 5.13  & 4.92  \\
          & \href{https://huggingface.co/datasets/BAAI/Infinity-Instruct}{Infinity-Instruct} & 5.44  & 4.40  & 4.92  & 5.37  & 5.16  & 5.69  & 5.89  & 5.74  & 5.96  & 5.63  & 5.28  \\
          & \href{https://huggingface.co/datasets/BelleGroup/train_3.5M_CN}{BELLE-SFT} \citep{BELLE} & 5.07  & 3.57  & 4.32  & 5.37  & 4.91  & 5.56  & 5.49  & 5.68  & 5.61  & 5.44  & 4.88  \\
          & \cellcolor[rgb]{ .906,  .902,  .902}TGPDG-SFT (GPT-4) & \cellcolor[rgb]{ .906,  .902,  .902}\textbf{7.42} & \cellcolor[rgb]{ .906,  .902,  .902}\textbf{5.83} & \cellcolor[rgb]{ .906,  .902,  .902}\textbf{6.62} & \cellcolor[rgb]{ .906,  .902,  .902}\textbf{6.37} & \cellcolor[rgb]{ .906,  .902,  .902}5.78  & \cellcolor[rgb]{ .906,  .902,  .902}\textbf{5.93} & \cellcolor[rgb]{ .906,  .902,  .902}\textbf{6.41} & \cellcolor[rgb]{ .906,  .902,  .902}\textbf{6.82} & \cellcolor[rgb]{ .906,  .902,  .902}\textbf{6.30} & \cellcolor[rgb]{ .906,  .902,  .902}\textbf{6.27} & \cellcolor[rgb]{ .906,  .902,  .902}\textbf{6.45} \\
          & \cellcolor[rgb]{ .906,  .902,  .902}TGPDG-SFT (DeepSeek-V2) & \cellcolor[rgb]{ .906,  .902,  .902}7.08  & \cellcolor[rgb]{ .906,  .902,  .902}5.41  & \cellcolor[rgb]{ .906,  .902,  .902}6.25  & \cellcolor[rgb]{ .906,  .902,  .902}5.57  & \cellcolor[rgb]{ .906,  .902,  .902}\textbf{5.95} & \cellcolor[rgb]{ .906,  .902,  .902}5.76  & \cellcolor[rgb]{ .906,  .902,  .902}6.28  & \cellcolor[rgb]{ .906,  .902,  .902}6.76  & \cellcolor[rgb]{ .906,  .902,  .902}6.10  & \cellcolor[rgb]{ .906,  .902,  .902}6.07  & \cellcolor[rgb]{ .906,  .902,  .902}6.16  \\
    \midrule
    \multirow{9}[2]{*}{Llama-3.1-8B} & \href{https://huggingface.co/datasets/YeungNLP/firefly-train-1.1M}{Firefly} & 2.41  & 2.38  & 2.40  & 3.87  & 3.86  & 3.97  & 4.53  & 3.53  & 4.09  & 3.97  & 3.19  \\
          & \href{https://github.com/Instruction-Tuning-with-GPT-4/GPT-4-LLM/blob/main/data/alpaca_gpt4_data_zh.json}{Alpaca-GPT-4-ZH} \citep{DBLP:journals/corr/abs-2304-03277} & 2.50  & 3.04  & 2.77  & 4.12  & 3.84  & 4.09  & 4.67  & 4.37  & 4.59  & 4.28  & 3.53  \\
          & \href{https://huggingface.co/datasets/BAAI/COIG}{COIG} \citep{DBLP:journals/corr/abs-2304-07987}  & 1.73  & 1.85  & 1.79  & 2.96  & 2.53  & 2.90  & 2.79  & 3.32  & 3.37  & 2.98  & 2.38  \\
          & \href{https://huggingface.co/datasets/fnlp/moss-003-sft-data}{MOSS-SFT} \citep{MIR-2023-12-294} & 2.63  & 3.05  & 2.84  & 4.23  & 3.66  & 4.34  & 4.62  & 4.97  & 4.66  & 4.41  & 3.63  \\
          & \href{https://huggingface.co/datasets/m-a-p/COIG-CQIA}{COIG-CQIA} \citep{DBLP:journals/corr/abs-2403-18058} & 2.30  & 2.78  & 2.54  & 3.45  & 3.28  & 3.41  & 3.49  & 3.95  & 3.86  & 3.57  & 3.06  \\
          & \href{https://huggingface.co/datasets/BAAI/Infinity-Instruct}{Infinity-Instruct} & 3.23  & 3.07  & 3.15  & 4.76  & 4.21  & 4.79  & \textbf{5.68} & 5.42  & 5.57  & 5.07  & 4.11  \\
          & \href{https://huggingface.co/datasets/BelleGroup/train_3.5M_CN}{BELLE-SFT} \citep{BELLE} & 3.57  & 3.38  & 3.47  & \textbf{4.92} & \textbf{4.24} & 4.81  & 5.31  & 5.55  & 5.48  & 5.05  & 4.26  \\
          & \cellcolor[rgb]{ .906,  .902,  .902}TGPDG-SFT (GPT-4) & \cellcolor[rgb]{ .906,  .902,  .902}\textbf{3.83} & \cellcolor[rgb]{ .906,  .902,  .902}\textbf{3.90} & \cellcolor[rgb]{ .906,  .902,  .902}\textbf{3.87} & \cellcolor[rgb]{ .906,  .902,  .902}4.90  & \cellcolor[rgb]{ .906,  .902,  .902}3.83  & \cellcolor[rgb]{ .906,  .902,  .902}\textbf{4.85} & \cellcolor[rgb]{ .906,  .902,  .902}5.52  & \cellcolor[rgb]{ .906,  .902,  .902}\textbf{6.05} & \cellcolor[rgb]{ .906,  .902,  .902}\textbf{5.71} & \cellcolor[rgb]{ .906,  .902,  .902}\textbf{5.14} & \cellcolor[rgb]{ .906,  .902,  .902}\textbf{4.50} \\
          & \cellcolor[rgb]{ .906,  .902,  .902}TGPDG-SFT (DeepSeek-V2) & \cellcolor[rgb]{ .906,  .902,  .902}3.48  & \cellcolor[rgb]{ .906,  .902,  .902}3.73  & \cellcolor[rgb]{ .906,  .902,  .902}3.61  & \cellcolor[rgb]{ .906,  .902,  .902}3.98  & \cellcolor[rgb]{ .906,  .902,  .902}3.64  & \cellcolor[rgb]{ .906,  .902,  .902}4.60  & \cellcolor[rgb]{ .906,  .902,  .902}5.41  & \cellcolor[rgb]{ .906,  .902,  .902}5.58  & \cellcolor[rgb]{ .906,  .902,  .902}5.28  & \cellcolor[rgb]{ .906,  .902,  .902}4.75  & \cellcolor[rgb]{ .906,  .902,  .902}4.18  \\
    \midrule
    \multirow{9}[2]{*}{Gemma-2-9B} & \href{https://huggingface.co/datasets/YeungNLP/firefly-train-1.1M}{Firefly} & 2.21  & 2.50  & 2.36  & 3.69  & 3.69  & 4.19  & 4.61  & 3.32  & 4.21  & 3.95  & 3.15  \\
          & \href{https://github.com/Instruction-Tuning-with-GPT-4/GPT-4-LLM/blob/main/data/alpaca_gpt4_data_zh.json}{Alpaca-GPT-4-ZH} \citep{DBLP:journals/corr/abs-2304-03277} & 3.18  & 3.16  & 3.17  & 4.32  & 3.36  & 4.18  & 4.40  & 4.47  & 4.59  & 4.22  & 3.70  \\
          & \href{https://huggingface.co/datasets/BAAI/COIG}{COIG} \citep{DBLP:journals/corr/abs-2304-07987}  & 1.47  & 1.92  & 1.70  & 2.94  & 2.38  & 2.40  & 2.56  & 3.26  & 3.30  & 2.81  & 2.25  \\
          & \href{https://huggingface.co/datasets/fnlp/moss-003-sft-data}{MOSS-SFT} \citep{MIR-2023-12-294} & 2.72  & 2.96  & 2.84  & 4.00  & 3.38  & 3.89  & 4.55  & 5.16  & 4.30  & 4.21  & 3.53  \\
          & \href{https://huggingface.co/datasets/m-a-p/COIG-CQIA}{COIG-CQIA} \citep{DBLP:journals/corr/abs-2403-18058} & 2.55  & 2.83  & 2.69  & 3.25  & 3.03  & 3.76  & 3.53  & 3.87  & 4.05  & 3.58  & 3.14  \\
          & \href{https://huggingface.co/datasets/BAAI/Infinity-Instruct}{Infinity-Instruct} & 3.92  & 3.52  & 3.72  & 5.06  & 4.19  & 5.12  & \textbf{5.47} & 5.61  & 5.83  & \textbf{5.21} & 4.47  \\
          & \href{https://huggingface.co/datasets/BelleGroup/train_3.5M_CN}{BELLE-SFT} \citep{BELLE} & 3.97  & 3.47  & 3.72  & \textbf{5.11} & \textbf{4.33} & \textbf{5.26} & 5.36  & 5.53  & 5.34  & 5.16  & 4.44  \\
          & \cellcolor[rgb]{ .906,  .902,  .902}TGPDG-SFT (GPT-4) & \cellcolor[rgb]{ .906,  .902,  .902}\textbf{4.50} & \cellcolor[rgb]{ .906,  .902,  .902}\textbf{3.71} & \cellcolor[rgb]{ .906,  .902,  .902}\textbf{4.10} & \cellcolor[rgb]{ .906,  .902,  .902}5.07  & \cellcolor[rgb]{ .906,  .902,  .902}3.84  & \cellcolor[rgb]{ .906,  .902,  .902}5.00  & \cellcolor[rgb]{ .906,  .902,  .902}5.41  & \cellcolor[rgb]{ .906,  .902,  .902}\textbf{5.79} & \cellcolor[rgb]{ .906,  .902,  .902}\textbf{5.91} & \cellcolor[rgb]{ .906,  .902,  .902}5.17  & \cellcolor[rgb]{ .906,  .902,  .902}\textbf{4.64} \\
          & \cellcolor[rgb]{ .906,  .902,  .902}TGPDG-SFT (DeepSeek-V2) & \cellcolor[rgb]{ .906,  .902,  .902}4.26  & \cellcolor[rgb]{ .906,  .902,  .902}3.66  & \cellcolor[rgb]{ .906,  .902,  .902}3.96  & \cellcolor[rgb]{ .906,  .902,  .902}4.52  & \cellcolor[rgb]{ .906,  .902,  .902}4.00  & \cellcolor[rgb]{ .906,  .902,  .902}4.87  & \cellcolor[rgb]{ .906,  .902,  .902}5.37  & \cellcolor[rgb]{ .906,  .902,  .902}5.37  & \cellcolor[rgb]{ .906,  .902,  .902}5.25  & \cellcolor[rgb]{ .906,  .902,  .902}4.90  & \cellcolor[rgb]{ .906,  .902,  .902}4.43  \\
    \midrule
    \multirow{9}[2]{*}{Qwen2.5-14B} & \href{https://huggingface.co/datasets/YeungNLP/firefly-train-1.1M}{Firefly} & 4.21  & 3.66  & 3.94  & 4.87  & 5.33  & 5.49  & 4.97  & 4.18  & 4.82  & 4.94  & 4.44  \\
          & \href{https://github.com/Instruction-Tuning-with-GPT-4/GPT-4-LLM/blob/main/data/alpaca_gpt4_data_zh.json}{Alpaca-GPT-4-ZH} \citep{DBLP:journals/corr/abs-2304-03277} & 6.14  & 5.13  & 5.64  & 5.94  & 5.53  & 6.12  & 5.37  & 5.84  & 5.59  & 5.73  & 5.68  \\
          & \href{https://huggingface.co/datasets/BAAI/COIG}{COIG} \citep{DBLP:journals/corr/abs-2304-07987}  & 3.21  & 3.39  & 3.30  & 4.92  & 5.04  & 4.94  & 4.43  & 5.18  & 4.55  & 4.84  & 4.07  \\
          & \href{https://huggingface.co/datasets/fnlp/moss-003-sft-data}{MOSS-SFT} \citep{MIR-2023-12-294} & 5.30  & 5.20  & 5.25  & 5.57  & 5.05  & 5.75  & 5.75  & 5.74  & 5.66  & 5.59  & 5.42  \\
          & \href{https://huggingface.co/datasets/m-a-p/COIG-CQIA}{COIG-CQIA} \citep{DBLP:journals/corr/abs-2403-18058} & 5.55  & 4.73  & 5.14  & 5.44  & 5.36  & 5.52  & 5.25  & 4.97  & 5.04  & 5.27  & 5.20  \\
          & \href{https://huggingface.co/datasets/BAAI/Infinity-Instruct}{Infinity-Instruct} & 5.96  & 4.55  & 5.26  & 5.86  & 5.37  & 6.03  & 6.19  & 6.03  & 6.16  & 5.94  & 5.60  \\
          & \href{https://huggingface.co/datasets/BelleGroup/train_3.5M_CN}{BELLE-SFT} \citep{BELLE} & 4.93  & 4.47  & 4.70  & 5.51  & 5.33  & 6.12  & 5.59  & 5.95  & 5.72  & 5.70  & 5.20  \\
          & \cellcolor[rgb]{ .906,  .902,  .902}TGPDG-SFT (GPT-4) & \cellcolor[rgb]{ .906,  .902,  .902}\textbf{7.50} & \cellcolor[rgb]{ .906,  .902,  .902}\textbf{6.02} & \cellcolor[rgb]{ .906,  .902,  .902}\textbf{6.76} & \cellcolor[rgb]{ .906,  .902,  .902}\textbf{6.87} & \cellcolor[rgb]{ .906,  .902,  .902}\textbf{5.76} & \cellcolor[rgb]{ .906,  .902,  .902}\textbf{6.54} & \cellcolor[rgb]{ .906,  .902,  .902}\textbf{6.65} & \cellcolor[rgb]{ .906,  .902,  .902}\textbf{6.71} & \cellcolor[rgb]{ .906,  .902,  .902}\textbf{6.53} & \cellcolor[rgb]{ .906,  .902,  .902}\textbf{6.51} & \cellcolor[rgb]{ .906,  .902,  .902}\textbf{6.64} \\
          & \cellcolor[rgb]{ .906,  .902,  .902}TGPDG-SFT (DeepSeek-V2) & \cellcolor[rgb]{ .906,  .902,  .902}7.21  & \cellcolor[rgb]{ .906,  .902,  .902}5.47  & \cellcolor[rgb]{ .906,  .902,  .902}6.34  & \cellcolor[rgb]{ .906,  .902,  .902}5.99  & \cellcolor[rgb]{ .906,  .902,  .902}5.57  & \cellcolor[rgb]{ .906,  .902,  .902}6.07  & \cellcolor[rgb]{ .906,  .902,  .902}5.87  & \cellcolor[rgb]{ .906,  .902,  .902}6.16  & \cellcolor[rgb]{ .906,  .902,  .902}5.85  & \cellcolor[rgb]{ .906,  .902,  .902}5.92  & \cellcolor[rgb]{ .906,  .902,  .902}6.13  \\
    \bottomrule
    \end{tabular}%
  \caption{Performance comparison of five LLMs on Alignbench after \textbf{supervised fine-tuning} with various datasets. The evaluation is conducted using \textbf{GPT-4o}, which scores the models' responses. The ``Fund.'' column denotes Fundamental Language Ability, ``Chi.'' denotes Advanced Chinese Understanding, ``Open.'' denotes Open-ended Questions, ``Writ.'' denotes Writing Ability, ``Role.'' denotes Task-oriented Role Play, ``Pro'' denotes ``Professional Knowledge'', ``Math.'' denotes Mathematics, and ``Logic.'' denotes Logical Reasoning.}
  \label{tab:sft_alignbench}%
\end{table*}%

\begin{table*}[!t]
  \centering
  \tiny
    \begin{tabular}{c|l|ccc|ccccccc|c}
    \toprule
    \multirow{2}[4]{*}{\textbf{Model}} & \multicolumn{1}{c|}{\multirow{2}[4]{*}{\textbf{Dataset}}} & \multicolumn{3}{c|}{\textbf{Reasoning}} & \multicolumn{7}{c|}{\textbf{Language}}                & \multicolumn{1}{c}{\multirow{2}[4]{*}{\textbf{Overall}}} \\
\cmidrule{3-12}          &       & \multicolumn{1}{c}{\textbf{Math.}} & \multicolumn{1}{c}{\textbf{Logi.}} & \multicolumn{1}{c|}{\textbf{Avg.}} & \multicolumn{1}{c}{\textbf{Pro.}} & \multicolumn{1}{c}{\textbf{Chi.}} & \multicolumn{1}{c}{\textbf{Fund.}} & \multicolumn{1}{c}{\textbf{Writ.}} & \multicolumn{1}{c}{\textbf{Open.}} & \multicolumn{1}{c}{\textbf{Role.}} & \multicolumn{1}{l|}{\textbf{Avg.}} &  \\
    \midrule
    \multirow{9}[2]{*}{Qwen2.5-3B} & \href{https://huggingface.co/datasets/YeungNLP/firefly-train-1.1M}{Firefly} & 3.84  & 2.92  & 3.38  & 4.31  & 4.50  & 4.21  & 4.61  & 4.08  & 4.49  & 4.37  & 3.87  \\
          & \href{https://github.com/Instruction-Tuning-with-GPT-4/GPT-4-LLM/blob/main/data/alpaca_gpt4_data_zh.json}{Alpaca-GPT-4-ZH} \citep{DBLP:journals/corr/abs-2304-03277} & 4.53  & 3.45  & 3.99  & 4.54  & 4.21  & 4.70  & 4.77  & 5.24  & 5.03  & 4.75  & 4.37  \\
          & \href{https://huggingface.co/datasets/BAAI/COIG}{COIG} \citep{DBLP:journals/corr/abs-2304-07987}  & 2.87  & 2.80  & 2.84  & 3.82  & 3.84  & 3.76  & 3.65  & 4.79  & 4.30  & 4.03  & 3.43  \\
          & \href{https://huggingface.co/datasets/fnlp/moss-003-sft-data}{MOSS-SFT} \citep{MIR-2023-12-294} & 4.24  & 3.22  & 3.73  & 4.78  & 4.24  & 4.66  & 5.00  & 5.13  & 5.09  & 4.82  & 4.27  \\
          & \href{https://huggingface.co/datasets/m-a-p/COIG-CQIA}{COIG-CQIA} \citep{DBLP:journals/corr/abs-2403-18058} & 3.81  & 3.49  & 3.65  & 3.99  & 4.12  & 4.43  & 4.41  & 4.42  & 4.51  & 4.31  & 3.98  \\
          & \href{https://huggingface.co/datasets/BAAI/Infinity-Instruct}{Infinity-Instruct} & 4.45  & 3.38  & 3.91  & 5.01  & 3.93  & 4.50  & 5.45  & 5.16  & 5.45  & 4.92  & 4.41  \\
          & \href{https://huggingface.co/datasets/BelleGroup/train_3.5M_CN}{BELLE-SFT} \citep{BELLE} & 4.29  & 3.34  & 3.82  & 4.73  & 4.28  & 4.76  & 5.37  & 5.47  & 5.35  & 4.99  & 4.41  \\
          & \cellcolor[rgb]{ .906,  .902,  .902}TaP-SFT (GPT-4) & \cellcolor[rgb]{ .906,  .902,  .902}5.41  & \cellcolor[rgb]{ .906,  .902,  .902}3.74  & \cellcolor[rgb]{ .906,  .902,  .902}4.57  & \cellcolor[rgb]{ .906,  .902,  .902}\textbf{5.24} & \cellcolor[rgb]{ .906,  .902,  .902}\textbf{4.93} & \cellcolor[rgb]{ .906,  .902,  .902}\textbf{5.19} & \cellcolor[rgb]{ .906,  .902,  .902}5.67  & \cellcolor[rgb]{ .906,  .902,  .902}\textbf{5.89} & \cellcolor[rgb]{ .906,  .902,  .902}\textbf{5.82} & \cellcolor[rgb]{ .906,  .902,  .902}\textbf{5.46} & \cellcolor[rgb]{ .906,  .902,  .902}5.02  \\
          & \cellcolor[rgb]{ .906,  .902,  .902}TaP-SFT (DeepSeek-V2) & \cellcolor[rgb]{ .906,  .902,  .902}\textbf{5.58} & \cellcolor[rgb]{ .906,  .902,  .902}\textbf{4.20} & \cellcolor[rgb]{ .906,  .902,  .902}\textbf{4.89} & \cellcolor[rgb]{ .906,  .902,  .902}5.15  & \cellcolor[rgb]{ .906,  .902,  .902}4.86  & \cellcolor[rgb]{ .906,  .902,  .902}4.90  & \cellcolor[rgb]{ .906,  .902,  .902}\textbf{5.75} & \cellcolor[rgb]{ .906,  .902,  .902}5.84  & \cellcolor[rgb]{ .906,  .902,  .902}\textbf{5.82} & \cellcolor[rgb]{ .906,  .902,  .902}5.39  & \cellcolor[rgb]{ .906,  .902,  .902}\textbf{5.14} \\
    \midrule
    \multirow{9}[2]{*}{Qwen2.5-7B} & \href{https://huggingface.co/datasets/YeungNLP/firefly-train-1.1M}{Firefly} & 3.91  & 3.34  & 3.62  & 4.67  & 4.41  & 5.00  & 4.92  & 3.95  & 4.89  & 4.64  & 4.13  \\
          & \href{https://github.com/Instruction-Tuning-with-GPT-4/GPT-4-LLM/blob/main/data/alpaca_gpt4_data_zh.json}{Alpaca-GPT-4-ZH} \citep{DBLP:journals/corr/abs-2304-03277} & 5.07  & 3.99  & 4.53  & 5.35  & 4.60  & 5.31  & 5.11  & 5.11  & 5.35  & 5.14  & 4.83  \\
          & \href{https://huggingface.co/datasets/BAAI/COIG}{COIG} \citep{DBLP:journals/corr/abs-2304-07987}  & 3.35  & 2.87  & 3.11  & 4.57  & 4.24  & 4.13  & 3.87  & 4.79  & 4.70  & 4.38  & 3.75  \\
          & \href{https://huggingface.co/datasets/fnlp/moss-003-sft-data}{MOSS-SFT} \citep{MIR-2023-12-294} & 5.09  & 4.01  & 4.55  & 5.30  & 4.47  & 4.85  & 5.16  & 5.76  & 5.50  & 5.17  & 4.86  \\
          & \href{https://huggingface.co/datasets/m-a-p/COIG-CQIA}{COIG-CQIA} \citep{DBLP:journals/corr/abs-2403-18058} & 5.12  & 3.90  & 4.51  & 4.98  & 5.09  & 4.81  & 4.96  & 5.26  & 5.01  & 5.02  & 4.76  \\
          & \href{https://huggingface.co/datasets/BAAI/Infinity-Instruct}{Infinity-Instruct} & 4.99  & 4.41  & 4.70  & 5.22  & 5.00  & 5.71  & 5.67  & 5.79  & 5.87  & 5.54  & 5.12  \\
          & \href{https://huggingface.co/datasets/BelleGroup/train_3.5M_CN}{BELLE-SFT} \citep{BELLE} & 4.73  & 3.59  & 4.16  & 5.10  & 4.74  & 5.59  & 5.37  & 5.68  & 5.64  & 5.35  & 4.76  \\
          & \cellcolor[rgb]{ .906,  .902,  .902}TaP-SFT (GPT-4) & \cellcolor[rgb]{ .906,  .902,  .902}\textbf{6.84} & \cellcolor[rgb]{ .906,  .902,  .902}\textbf{5.36} & \cellcolor[rgb]{ .906,  .902,  .902}\textbf{6.10} & \cellcolor[rgb]{ .906,  .902,  .902}\textbf{6.06} & \cellcolor[rgb]{ .906,  .902,  .902}5.14  & \cellcolor[rgb]{ .906,  .902,  .902}5.58  & \cellcolor[rgb]{ .906,  .902,  .902}6.04  & \cellcolor[rgb]{ .906,  .902,  .902}6.05  & \cellcolor[rgb]{ .906,  .902,  .902}6.09  & \cellcolor[rgb]{ .906,  .902,  .902}5.83  & \cellcolor[rgb]{ .906,  .902,  .902}\textbf{5.96} \\
          & \cellcolor[rgb]{ .906,  .902,  .902}TaP-SFT (DeepSeek-V2) & \cellcolor[rgb]{ .906,  .902,  .902}6.63  & \cellcolor[rgb]{ .906,  .902,  .902}4.91  & \cellcolor[rgb]{ .906,  .902,  .902}5.77  & \cellcolor[rgb]{ .906,  .902,  .902}5.97  & \cellcolor[rgb]{ .906,  .902,  .902}\textbf{5.53} & \cellcolor[rgb]{ .906,  .902,  .902}\textbf{5.76} & \cellcolor[rgb]{ .906,  .902,  .902}\textbf{6.07} & \cellcolor[rgb]{ .906,  .902,  .902}\textbf{6.11} & \cellcolor[rgb]{ .906,  .902,  .902}\textbf{6.13} & \cellcolor[rgb]{ .906,  .902,  .902}\textbf{5.93} & \cellcolor[rgb]{ .906,  .902,  .902}5.85  \\
    \midrule
    \multirow{9}[2]{*}{Llama-3.1-8B} & \href{https://huggingface.co/datasets/YeungNLP/firefly-train-1.1M}{Firefly} & 2.30  & 2.50  & 2.40  & 3.90  & 3.76  & 4.09  & 4.71  & 3.79  & 4.21  & 4.07  & 3.24  \\
          & \href{https://github.com/Instruction-Tuning-with-GPT-4/GPT-4-LLM/blob/main/data/alpaca_gpt4_data_zh.json}{Alpaca-GPT-4-ZH} \citep{DBLP:journals/corr/abs-2304-03277} & 2.35  & 2.93  & 2.64  & 3.95  & 3.50  & 3.88  & 4.49  & 4.34  & 4.56  & 4.12  & 3.38  \\
          & \href{https://huggingface.co/datasets/BAAI/COIG}{COIG} \citep{DBLP:journals/corr/abs-2304-07987}  & 1.68  & 2.08  & 1.88  & 3.01  & 2.38  & 2.78  & 2.84  & 3.71  & 3.52  & 3.04  & 2.46  \\
          & \href{https://huggingface.co/datasets/fnlp/moss-003-sft-data}{MOSS-SFT} \citep{MIR-2023-12-294} & 2.51  & 2.87  & 2.69  & 4.10  & 3.09  & 4.04  & 4.57  & 5.37  & 4.74  & 4.32  & 3.50  \\
          & \href{https://huggingface.co/datasets/m-a-p/COIG-CQIA}{COIG-CQIA} \citep{DBLP:journals/corr/abs-2403-18058} & 2.24  & 2.53  & 2.39  & 3.40  & 2.93  & 3.43  & 3.45  & 4.18  & 3.97  & 3.56  & 2.97  \\
          & \href{https://huggingface.co/datasets/BAAI/Infinity-Instruct}{Infinity-Instruct} & 2.88  & 3.14  & 3.01  & 4.71  & \textbf{4.24} & 4.38  & \textbf{5.48} & 5.26  & \textbf{5.60} & \textbf{4.95} & 3.98  \\
          & \href{https://huggingface.co/datasets/BelleGroup/train_3.5M_CN}{BELLE-SFT} \citep{BELLE} & 3.06  & 3.41  & 3.24  & \textbf{4.78} & 3.97  & \textbf{4.62} & 5.04  & \textbf{5.55} & 5.34  & 4.88  & 4.06  \\
          & \cellcolor[rgb]{ .906,  .902,  .902}TaP-SFT (GPT-4) & \cellcolor[rgb]{ .906,  .902,  .902}3.31  & \cellcolor[rgb]{ .906,  .902,  .902}3.48  & \cellcolor[rgb]{ .906,  .902,  .902}3.40  & \cellcolor[rgb]{ .906,  .902,  .902}4.73  & \cellcolor[rgb]{ .906,  .902,  .902}3.14  & \cellcolor[rgb]{ .906,  .902,  .902}4.46  & \cellcolor[rgb]{ .906,  .902,  .902}5.32  & \cellcolor[rgb]{ .906,  .902,  .902}5.34  & \cellcolor[rgb]{ .906,  .902,  .902}5.37  & \cellcolor[rgb]{ .906,  .902,  .902}4.73  & \cellcolor[rgb]{ .906,  .902,  .902}\textbf{4.06} \\
          & \cellcolor[rgb]{ .906,  .902,  .902}TaP-SFT (DeepSeek-V2) & \cellcolor[rgb]{ .906,  .902,  .902}\textbf{3.44} & \cellcolor[rgb]{ .906,  .902,  .902}\textbf{3.50} & \cellcolor[rgb]{ .906,  .902,  .902}\textbf{3.47} & \cellcolor[rgb]{ .906,  .902,  .902}4.40  & \cellcolor[rgb]{ .906,  .902,  .902}3.12  & \cellcolor[rgb]{ .906,  .902,  .902}4.49  & \cellcolor[rgb]{ .906,  .902,  .902}5.09  & \cellcolor[rgb]{ .906,  .902,  .902}5.21  & \cellcolor[rgb]{ .906,  .902,  .902}5.34  & \cellcolor[rgb]{ .906,  .902,  .902}4.61  & \cellcolor[rgb]{ .906,  .902,  .902}4.04  \\
    \midrule
    \multirow{9}[2]{*}{Gemma-2-9B} & \href{https://huggingface.co/datasets/YeungNLP/firefly-train-1.1M}{Firefly} & 2.29  & 2.59  & 2.44  & 3.81  & 3.64  & 4.24  & 4.77  & 3.53  & 4.40  & 4.06  & 3.25  \\
          & \href{https://github.com/Instruction-Tuning-with-GPT-4/GPT-4-LLM/blob/main/data/alpaca_gpt4_data_zh.json}{Alpaca-GPT-4-ZH} \citep{DBLP:journals/corr/abs-2304-03277} & 3.18  & 3.21  & 3.19  & 4.24  & 2.76  & 4.25  & 4.47  & 4.74  & 4.59  & 4.17  & 3.68  \\
          & \href{https://huggingface.co/datasets/BAAI/COIG}{COIG} \citep{DBLP:journals/corr/abs-2304-07987}  & 1.64  & 2.26  & 1.95  & 2.90  & 1.95  & 2.46  & 2.56  & 3.42  & 3.53  & 2.80  & 2.38  \\
          & \href{https://huggingface.co/datasets/fnlp/moss-003-sft-data}{MOSS-SFT} \citep{MIR-2023-12-294} & 2.62  & 2.85  & 2.73  & 3.94  & 3.02  & 3.70  & 4.53  & 5.50  & 4.46  & 4.19  & 3.46  \\
          & \href{https://huggingface.co/datasets/m-a-p/COIG-CQIA}{COIG-CQIA} \citep{DBLP:journals/corr/abs-2403-18058} & 2.49  & 2.79  & 2.64  & 3.26  & 2.74  & 3.75  & 3.45  & 4.00  & 4.14  & 3.56  & 3.10  \\
          & \href{https://huggingface.co/datasets/BAAI/Infinity-Instruct}{Infinity-Instruct} & 3.64  & 3.50  & 3.57  & 4.77  & 3.79  & 4.88  & \textbf{5.33} & 5.29  & \textbf{5.85} & 4.99  & 4.28  \\
          & \href{https://huggingface.co/datasets/BelleGroup/train_3.5M_CN}{BELLE-SFT} \citep{BELLE} & 3.80  & \textbf{3.57} & \textbf{3.68} & \textbf{4.96} & \textbf{4.12} & \textbf{4.99} & 5.25  & \textbf{5.53} & 5.55  & \textbf{5.07} & \textbf{4.37} \\
          & \cellcolor[rgb]{ .906,  .902,  .902}TaP-SFT (GPT-4) & \cellcolor[rgb]{ .906,  .902,  .902}\textbf{4.00} & \cellcolor[rgb]{ .906,  .902,  .902}3.20  & \cellcolor[rgb]{ .906,  .902,  .902}3.60  & \cellcolor[rgb]{ .906,  .902,  .902}4.84  & \cellcolor[rgb]{ .906,  .902,  .902}3.33  & \cellcolor[rgb]{ .906,  .902,  .902}4.76  & \cellcolor[rgb]{ .906,  .902,  .902}5.25  & \cellcolor[rgb]{ .906,  .902,  .902}5.26  & \cellcolor[rgb]{ .906,  .902,  .902}5.81  & \cellcolor[rgb]{ .906,  .902,  .902}4.88  & \cellcolor[rgb]{ .906,  .902,  .902}4.24  \\
          & \cellcolor[rgb]{ .906,  .902,  .902}TaP-SFT (DeepSeek-V2) & \cellcolor[rgb]{ .906,  .902,  .902}3.84  & \cellcolor[rgb]{ .906,  .902,  .902}3.38  & \cellcolor[rgb]{ .906,  .902,  .902}3.61  & \cellcolor[rgb]{ .906,  .902,  .902}4.51  & \cellcolor[rgb]{ .906,  .902,  .902}3.16  & \cellcolor[rgb]{ .906,  .902,  .902}4.60  & \cellcolor[rgb]{ .906,  .902,  .902}5.08  & \cellcolor[rgb]{ .906,  .902,  .902}5.03  & \cellcolor[rgb]{ .906,  .902,  .902}5.13  & \cellcolor[rgb]{ .906,  .902,  .902}4.58  & \cellcolor[rgb]{ .906,  .902,  .902}4.10  \\
    \midrule
    \multirow{9}[2]{*}{Qwen2.5-14B} & \href{https://huggingface.co/datasets/YeungNLP/firefly-train-1.1M}{Firefly} & 4.40  & 3.79  & 4.10  & 4.81  & 5.10  & 5.06  & 5.13  & 4.26  & 4.97  & 4.89  & 4.49  \\
          & \href{https://github.com/Instruction-Tuning-with-GPT-4/GPT-4-LLM/blob/main/data/alpaca_gpt4_data_zh.json}{Alpaca-GPT-4-ZH} \citep{DBLP:journals/corr/abs-2304-03277} & 5.89  & 4.79  & 5.34  & 5.77  & 5.41  & 5.88  & 5.36  & 6.05  & 5.90  & 5.73  & 5.54  \\
          & \href{https://huggingface.co/datasets/BAAI/COIG}{COIG} \citep{DBLP:journals/corr/abs-2304-07987}  & 3.50  & 3.25  & 3.38  & 4.91  & 4.62  & 4.95  & 4.41  & 5.61  & 4.68  & 4.86  & 4.12  \\
          & \href{https://huggingface.co/datasets/fnlp/moss-003-sft-data}{MOSS-SFT} \citep{MIR-2023-12-294} & 5.03  & 4.89  & 4.96  & 5.47  & 4.79  & 5.15  & 5.71  & 5.89  & 5.89  & 5.48  & 5.22  \\
          & \href{https://huggingface.co/datasets/m-a-p/COIG-CQIA}{COIG-CQIA} \citep{DBLP:journals/corr/abs-2403-18058} & 5.40  & 4.47  & 4.94  & 5.48  & 5.00  & 5.55  & 5.17  & 5.13  & 5.21  & 5.26  & 5.10  \\
          & \href{https://huggingface.co/datasets/BAAI/Infinity-Instruct}{Infinity-Instruct} & 5.71  & 4.61  & 5.16  & 5.81  & 5.09  & 6.09  & 6.01  & 5.71  & 6.12  & 5.81  & 5.48  \\
          & \href{https://huggingface.co/datasets/BelleGroup/train_3.5M_CN}{BELLE-SFT} \citep{BELLE} & 4.72  & 4.37  & 4.55  & 5.56  & 4.74  & 6.13  & 5.55  & 5.89  & 5.78  & 5.61  & 5.08  \\
          & \cellcolor[rgb]{ .906,  .902,  .902}TaP-SFT (GPT-4) & \cellcolor[rgb]{ .906,  .902,  .902}6.79  & \cellcolor[rgb]{ .906,  .902,  .902}\textbf{5.75} & \cellcolor[rgb]{ .906,  .902,  .902}6.27  & \cellcolor[rgb]{ .906,  .902,  .902}\textbf{6.46} & \cellcolor[rgb]{ .906,  .902,  .902}5.78  & \cellcolor[rgb]{ .906,  .902,  .902}6.49  & \cellcolor[rgb]{ .906,  .902,  .902}\textbf{6.27} & \cellcolor[rgb]{ .906,  .902,  .902}\textbf{6.29} & \cellcolor[rgb]{ .906,  .902,  .902}\textbf{6.37} & \cellcolor[rgb]{ .906,  .902,  .902}\textbf{6.27} & \cellcolor[rgb]{ .906,  .902,  .902}6.27  \\
          & \cellcolor[rgb]{ .906,  .902,  .902}TaP-SFT (DeepSeek-V2) & \cellcolor[rgb]{ .906,  .902,  .902}\textbf{7.36} & \cellcolor[rgb]{ .906,  .902,  .902}5.57  & \cellcolor[rgb]{ .906,  .902,  .902}\textbf{6.46} & \cellcolor[rgb]{ .906,  .902,  .902}6.37  & \cellcolor[rgb]{ .906,  .902,  .902}\textbf{5.81} & \cellcolor[rgb]{ .906,  .902,  .902}\textbf{6.63} & \cellcolor[rgb]{ .906,  .902,  .902}6.17  & \cellcolor[rgb]{ .906,  .902,  .902}6.24  & \cellcolor[rgb]{ .906,  .902,  .902}6.22  & \cellcolor[rgb]{ .906,  .902,  .902}6.24  & \cellcolor[rgb]{ .906,  .902,  .902}\textbf{6.35} \\
    \bottomrule
    \end{tabular}%
  \caption{Performance comparison of five LLMs on Alignbench after \textbf{supervised fine-tuning} with various datasets. The evaluation is conducted using \textbf{DeepSeek-V3}, which scores the models' responses. The ``Fund.'' column denotes Fundamental Language Ability, ``Chi.'' denotes Advanced Chinese Understanding, ``Open.'' denotes Open-ended Questions, ``Writ.'' denotes Writing Ability, ``Role.'' denotes Task-oriented Role Play, ``Pro'' denotes ``Professional Knowledge'', ``Math.'' denotes Mathematics, and ``Logic.'' denotes Logical Reasoning.}
  \label{tab:sft_alignbench_deepseek}%
\end{table*}%

\begin{table*}[!t]
  \centering
  \tiny
    \begin{tabular}{c|l|ccc|ccccccc|c}
    \toprule
    \multirow{2}[4]{*}{\textbf{Model}} & \multicolumn{1}{c|}{\multirow{2}[4]{*}{\textbf{Dataset}}} & \multicolumn{3}{c|}{\textbf{Reasoning}} & \multicolumn{7}{c|}{\textbf{Language}}                & \multicolumn{1}{c}{\multirow{2}[4]{*}{\textbf{Overall}}} \\
\cmidrule{3-12}          &       & \multicolumn{1}{c}{\textbf{Math.}} & \multicolumn{1}{c}{\textbf{Logi.}} & \multicolumn{1}{c|}{\textbf{Avg.}} & \multicolumn{1}{c}{\textbf{Pro.}} & \multicolumn{1}{c}{\textbf{Chi.}} & \multicolumn{1}{c}{\textbf{Fund.}} & \multicolumn{1}{c}{\textbf{Writ.}} & \multicolumn{1}{c}{\textbf{Open.}} & \multicolumn{1}{c}{\textbf{Role.}} & \multicolumn{1}{c|}{\textbf{Avg.}} &  \\
    \midrule
    \multirow{3}[2]{*}{Qwen2.5-3B-SFT-Open} & \href{https://github.com/HIT-SCIR/huozi/tree/main/data/huozi-rlhf}{Huozi-RLHF} \citep{huozi}  & 4.63  & 3.42  & 4.02  & 4.86  & 4.74  & 4.87  & 5.55  & 5.61  & 5.74  & 5.23  & 4.63  \\
          & \href{https://huggingface.co/datasets/wenbopan/Chinese-dpo-pairs}{Chinese-DPO-Pairs} & 5.09  & 3.84  & 4.46  & 5.08  & \textbf{5.26} & 4.97  & 6.01  & 6.00  & 5.78  & 5.52  & 4.99  \\
          & \cellcolor[rgb]{ .906,  .902,  .902}TaP (GPT-4) & \cellcolor[rgb]{ .906,  .902,  .902}\textbf{5.78} & \cellcolor[rgb]{ .906,  .902,  .902}\textbf{4.60} & \cellcolor[rgb]{ .906,  .902,  .902}\textbf{5.19} & \cellcolor[rgb]{ .906,  .902,  .902}\textbf{5.71} & \cellcolor[rgb]{ .906,  .902,  .902}5.25  & \cellcolor[rgb]{ .906,  .902,  .902}\textbf{5.27} & \cellcolor[rgb]{ .906,  .902,  .902}\textbf{6.51} & \cellcolor[rgb]{ .906,  .902,  .902}\textbf{6.29} & \cellcolor[rgb]{ .906,  .902,  .902}\textbf{6.41} & \cellcolor[rgb]{ .906,  .902,  .902}\textbf{5.90} & \cellcolor[rgb]{ .906,  .902,  .902}\textbf{5.55} \\
    \midrule
    \multirow{3}[2]{*}{Qwen2.5-3B-SFT-TaP} & \href{https://github.com/HIT-SCIR/huozi/tree/main/data/huozi-rlhf}{Huozi-RLHF} \citep{huozi}  & 6.01  & 4.29  & 5.15  & 5.64  & 5.49  & 5.37  & 6.23  & 6.21  & 5.98  & 5.82  & 5.49  \\
          & \href{https://huggingface.co/datasets/wenbopan/Chinese-dpo-pairs}{Chinese-DPO-Pairs} & 6.09  & 4.41  & 5.25  & 5.74  & \textbf{5.62} & \textbf{5.47} & 6.45  & \textbf{6.55} & 6.20  & 6.01  & 5.63  \\
          & \cellcolor[rgb]{ .906,  .902,  .902}TaP (GPT-4) & \cellcolor[rgb]{ .906,  .902,  .902}\textbf{6.39} & \cellcolor[rgb]{ .906,  .902,  .902}\textbf{4.74} & \cellcolor[rgb]{ .906,  .902,  .902}\textbf{5.57} & \cellcolor[rgb]{ .906,  .902,  .902}\textbf{5.77} & \cellcolor[rgb]{ .906,  .902,  .902}5.40  & \cellcolor[rgb]{ .906,  .902,  .902}5.46  & \cellcolor[rgb]{ .906,  .902,  .902}\textbf{6.68} & \cellcolor[rgb]{ .906,  .902,  .902}6.53  & \cellcolor[rgb]{ .906,  .902,  .902}\textbf{6.54} & \cellcolor[rgb]{ .906,  .902,  .902}\textbf{6.06} & \cellcolor[rgb]{ .906,  .902,  .902}\textbf{5.81} \\
    \midrule

    \multirow{3}[2]{*}{Qwen2.5-7B-SFT-Open} & \href{https://github.com/HIT-SCIR/huozi/tree/main/data/huozi-rlhf}{Huozi-RLHF} \citep{huozi}  & 5.04  & 4.01  & 4.53  & 5.48  & 5.02  & 5.24  & 5.43  & 5.68  & 5.50  & 5.39  & 4.96  \\
          & \href{https://huggingface.co/datasets/wenbopan/Chinese-dpo-pairs}{Chinese-DPO-Pairs} & 5.36  & 4.40  & 4.88  & 5.54  & 5.18  & 5.36  & 5.75  & \textbf{5.89} & 5.49  & 5.53  & 5.21  \\
          & \cellcolor[rgb]{ .906,  .902,  .902}TaP (GPT-4) & \cellcolor[rgb]{ .906,  .902,  .902}\textbf{5.64} & \cellcolor[rgb]{ .906,  .902,  .902}\textbf{4.59} & \cellcolor[rgb]{ .906,  .902,  .902}\textbf{5.11} & \cellcolor[rgb]{ .906,  .902,  .902}\textbf{5.63} & \cellcolor[rgb]{ .906,  .902,  .902}\textbf{5.28} & \cellcolor[rgb]{ .906,  .902,  .902}\textbf{5.46} & \cellcolor[rgb]{ .906,  .902,  .902}\textbf{6.23} & \cellcolor[rgb]{ .906,  .902,  .902}5.79  & \cellcolor[rgb]{ .906,  .902,  .902}\textbf{6.05} & \cellcolor[rgb]{ .906,  .902,  .902}\textbf{5.74} & \cellcolor[rgb]{ .906,  .902,  .902}\textbf{5.43} \\
    \midrule
    \multirow{3}[2]{*}{Qwen2.5-7B-SFT-TaP} & \href{https://github.com/HIT-SCIR/huozi/tree/main/data/huozi-rlhf}{Huozi-RLHF} \citep{huozi}  & 7.44  & 5.04  & 6.24  & 6.31  & 5.78  & 6.34  & 6.41  & 6.79  & 6.39  & 6.34  & 6.29  \\
          & \href{https://huggingface.co/datasets/wenbopan/Chinese-dpo-pairs}{Chinese-DPO-Pairs} & 7.46  & \textbf{5.53} & 6.49  & 6.52  & 5.91  & 6.15  & 6.64  & 6.79  & 6.53  & 6.42  & 6.46  \\
          & \cellcolor[rgb]{ .906,  .902,  .902}TaP (GPT-4) & \cellcolor[rgb]{ .906,  .902,  .902}\textbf{7.66} & \cellcolor[rgb]{ .906,  .902,  .902}5.52  & \cellcolor[rgb]{ .906,  .902,  .902}\textbf{6.59} & \cellcolor[rgb]{ .906,  .902,  .902}\textbf{6.79} & \cellcolor[rgb]{ .906,  .902,  .902}\textbf{6.00} & \cellcolor[rgb]{ .906,  .902,  .902}\textbf{6.40} & \cellcolor[rgb]{ .906,  .902,  .902}\textbf{6.88} & \cellcolor[rgb]{ .906,  .902,  .902}\textbf{6.95} & \cellcolor[rgb]{ .906,  .902,  .902}\textbf{6.93} & \cellcolor[rgb]{ .906,  .902,  .902}\textbf{6.66} & \cellcolor[rgb]{ .906,  .902,  .902}\textbf{6.62} \\
    \midrule
    \multirow{3}[2]{*}{Llama-3.1-8B-SFT-Open} & \href{https://github.com/HIT-SCIR/huozi/tree/main/data/huozi-rlhf}{Huozi-RLHF} \citep{huozi}  & 3.68  & 3.37  & 3.52  & 4.98  & 4.43  & 4.78  & 5.40  & 5.74  & 5.83  & 5.19  & 4.36  \\
          & \href{https://huggingface.co/datasets/wenbopan/Chinese-dpo-pairs}{Chinese-DPO-Pairs} & 3.81  & 3.25  & 3.53  & 5.06  & 4.45  & \textbf{5.12} & 6.15  & \textbf{6.58} & 6.04  & 5.57  & 4.55  \\
          & \cellcolor[rgb]{ .906,  .902,  .902}TaP (GPT-4) & \cellcolor[rgb]{ .906,  .902,  .902}\textbf{4.25} & \cellcolor[rgb]{ .906,  .902,  .902}\textbf{3.67} & \cellcolor[rgb]{ .906,  .902,  .902}\textbf{3.96} & \cellcolor[rgb]{ .906,  .902,  .902}\textbf{5.13} & \cellcolor[rgb]{ .906,  .902,  .902}\textbf{4.72} & \cellcolor[rgb]{ .906,  .902,  .902}4.94  & \cellcolor[rgb]{ .906,  .902,  .902}\textbf{6.40} & \cellcolor[rgb]{ .906,  .902,  .902}6.03  & \cellcolor[rgb]{ .906,  .902,  .902}\textbf{6.48} & \cellcolor[rgb]{ .906,  .902,  .902}\textbf{5.62} & \cellcolor[rgb]{ .906,  .902,  .902}\textbf{4.79} \\
    \midrule
    \multirow{3}[2]{*}{Llama-3.1-8B-SFT-TaP} & \href{https://github.com/HIT-SCIR/huozi/tree/main/data/huozi-rlhf}{Huozi-RLHF} \citep{huozi}  & 3.47  & 3.93  & 3.70  & 4.84  & 3.93  & 4.87  & 5.71  & 6.18  & 5.78  & 5.22  & 4.46  \\
          & \href{https://huggingface.co/datasets/wenbopan/Chinese-dpo-pairs}{Chinese-DPO-Pairs} & \textbf{4.19} & 3.95  & 4.07  & \textbf{5.05} & 3.93  & 4.99  & 6.00  & 6.05  & 6.14  & 5.36  & 4.71  \\
          & \cellcolor[rgb]{ .906,  .902,  .902}TaP (GPT-4) & \cellcolor[rgb]{ .906,  .902,  .902}4.14  & \cellcolor[rgb]{ .906,  .902,  .902}\textbf{4.07} & \cellcolor[rgb]{ .906,  .902,  .902}\textbf{4.10} & \cellcolor[rgb]{ .906,  .902,  .902}5.02  & \cellcolor[rgb]{ .906,  .902,  .902}\textbf{3.95} & \cellcolor[rgb]{ .906,  .902,  .902}\textbf{5.25} & \cellcolor[rgb]{ .906,  .902,  .902}\textbf{6.23} & \cellcolor[rgb]{ .906,  .902,  .902}\textbf{6.34} & \cellcolor[rgb]{ .906,  .902,  .902}\textbf{6.23} & \cellcolor[rgb]{ .906,  .902,  .902}\textbf{5.50} & \cellcolor[rgb]{ .906,  .902,  .902}\textbf{4.80} \\
    \midrule
    \multirow{3}[2]{*}{Gemma-2-9B-SFT-Open} & \href{https://github.com/HIT-SCIR/huozi/tree/main/data/huozi-rlhf}{Huozi-RLHF} \citep{huozi}  & 4.36  & 3.51  & 3.93  & 5.01  & 4.10  & 5.15  & 5.57  & 5.55  & 5.93  & 5.22  & 4.58  \\
          & \href{https://huggingface.co/datasets/wenbopan/Chinese-dpo-pairs}{Chinese-DPO-Pairs} & 4.18  & 3.54  & 3.86  & 5.17  & 4.29  & 5.10  & 5.67  & 5.74  & 5.97  & 5.32  & 4.59  \\
          & \cellcolor[rgb]{ .906,  .902,  .902}TaP (GPT-4) & \cellcolor[rgb]{ .906,  .902,  .902}\textbf{4.80} & \cellcolor[rgb]{ .906,  .902,  .902}\textbf{4.29} & \cellcolor[rgb]{ .906,  .902,  .902}\textbf{4.55} & \cellcolor[rgb]{ .906,  .902,  .902}\textbf{5.52} & \cellcolor[rgb]{ .906,  .902,  .902}\textbf{4.69} & \cellcolor[rgb]{ .906,  .902,  .902}\textbf{5.72} & \cellcolor[rgb]{ .906,  .902,  .902}\textbf{6.28} & \cellcolor[rgb]{ .906,  .902,  .902}\textbf{6.50} & \cellcolor[rgb]{ .906,  .902,  .902}\textbf{6.67} & \cellcolor[rgb]{ .906,  .902,  .902}\textbf{5.90} & \cellcolor[rgb]{ .906,  .902,  .902}\textbf{5.22} \\
    \midrule
    \multirow{3}[2]{*}{Gemma-2-9B-SFT-TaP} & \href{https://github.com/HIT-SCIR/huozi/tree/main/data/huozi-rlhf}{Huozi-RLHF} \citep{huozi}  & 4.78  & 3.59  & 4.18  & 4.94  & 3.83  & 5.09  & 5.71  & 5.95  & 5.84  & 5.23  & 4.70  \\
          & \href{https://huggingface.co/datasets/wenbopan/Chinese-dpo-pairs}{Chinese-DPO-Pairs} & 5.12  & \textbf{4.22} & 4.67  & 5.23  & \textbf{4.00} & \textbf{5.46} & 6.11  & \textbf{6.11} & 6.23  & 5.52  & 5.09  \\
          & \cellcolor[rgb]{ .906,  .902,  .902}TaP (GPT-4) & \cellcolor[rgb]{ .906,  .902,  .902}\textbf{5.31} & \cellcolor[rgb]{ .906,  .902,  .902}4.12  & \cellcolor[rgb]{ .906,  .902,  .902}\textbf{4.72} & \cellcolor[rgb]{ .906,  .902,  .902}\textbf{5.35} & \cellcolor[rgb]{ .906,  .902,  .902}3.98  & \cellcolor[rgb]{ .906,  .902,  .902}\textbf{5.46} & \cellcolor[rgb]{ .906,  .902,  .902}\textbf{6.43} & \cellcolor[rgb]{ .906,  .902,  .902}\textbf{6.11} & \cellcolor[rgb]{ .906,  .902,  .902}\textbf{6.34} & \cellcolor[rgb]{ .906,  .902,  .902}\textbf{5.61} & \cellcolor[rgb]{ .906,  .902,  .902}\textbf{5.16} \\
    \midrule
    \multirow{3}[2]{*}{Qwen2.5-14B-SFT-Open} & \href{https://github.com/HIT-SCIR/huozi/tree/main/data/huozi-rlhf}{Huozi-RLHF} \citep{huozi}  & 6.17  & 5.34  & 5.75  & 6.19  & 5.74  & 6.18  & 5.61  & 5.76  & 5.90  & 5.90  & 5.82  \\
          & \href{https://huggingface.co/datasets/wenbopan/Chinese-dpo-pairs}{Chinese-DPO-Pairs} & 6.21  & 5.49  & 5.85  & 6.27  & 5.78  & 6.03  & 5.80  & 6.05  & 5.84  & 5.96  & 5.90  \\
          & \cellcolor[rgb]{ .906,  .902,  .902}TaP (GPT-4) & \cellcolor[rgb]{ .906,  .902,  .902}\textbf{7.11} & \cellcolor[rgb]{ .906,  .902,  .902}\textbf{5.78} & \cellcolor[rgb]{ .906,  .902,  .902}\textbf{6.44} & \cellcolor[rgb]{ .906,  .902,  .902}\textbf{6.67} & \cellcolor[rgb]{ .906,  .902,  .902}\textbf{6.19} & \cellcolor[rgb]{ .906,  .902,  .902}\textbf{6.38} & \cellcolor[rgb]{ .906,  .902,  .902}\textbf{6.24} & \cellcolor[rgb]{ .906,  .902,  .902}\textbf{6.39} & \cellcolor[rgb]{ .906,  .902,  .902}\textbf{6.44} & \cellcolor[rgb]{ .906,  .902,  .902}\textbf{6.39} & \cellcolor[rgb]{ .906,  .902,  .902}\textbf{6.42} \\
    \midrule
    \multirow{3}[2]{*}{Qwen2.5-14B-SFT-TaP} & \href{https://github.com/HIT-SCIR/huozi/tree/main/data/huozi-rlhf}{Huozi-RLHF} \citep{huozi}  & 7.54  & 6.35  & 6.94  & 6.95  & 6.38  & 6.75  & 6.55  & 6.68  & 6.51  & 6.64  & 6.79  \\
          & \href{https://huggingface.co/datasets/wenbopan/Chinese-dpo-pairs}{Chinese-DPO-Pairs} & 7.93  & 6.05  & 6.99  & \textbf{7.14} & 6.31  & \textbf{6.85} & 6.93  & 7.05  & 6.84  & 6.85  & 6.92  \\
          & \cellcolor[rgb]{ .906,  .902,  .902}TaP (GPT-4) & \cellcolor[rgb]{ .906,  .902,  .902}\textbf{8.03} & \cellcolor[rgb]{ .906,  .902,  .902}\textbf{6.51} & \cellcolor[rgb]{ .906,  .902,  .902}\textbf{7.27} & \cellcolor[rgb]{ .906,  .902,  .902}7.09  & \cellcolor[rgb]{ .906,  .902,  .902}\textbf{6.83} & \cellcolor[rgb]{ .906,  .902,  .902}6.59  & \cellcolor[rgb]{ .906,  .902,  .902}\textbf{7.13} & \cellcolor[rgb]{ .906,  .902,  .902}\textbf{7.55} & \cellcolor[rgb]{ .906,  .902,  .902}\textbf{7.29} & \cellcolor[rgb]{ .906,  .902,  .902}\textbf{7.08} & \cellcolor[rgb]{ .906,  .902,  .902}\textbf{7.17} \\
    \bottomrule
    \end{tabular}%
  \caption{Performance comparison of LLMs trained with \textbf{DPO} using different datasets on AlignBench. The model names include two possible suffixes: ``Open'' and ``TaP.'' The ``Open'' suffix indicates that the LLMs were initialized from models trained via supervised fine-tuning on open-source datasets, whereas ``TaP'' denotes initialization from models trained on a dataset constructed by TaP. The evaluation is conducted using \textbf{GPT-4o}, which scores the models’ responses. The ``Fund.'' column denotes Fundamental Language Ability, ``Chi.'' denotes Advanced Chinese Understanding, ``Open.'' denotes Open-ended Questions, ``Writ.'' denotes Writing Ability, ``Role.'' denotes Task-oriented Role Play, ``Pro'' denotes ``Professional Knowledge'', ``Math.'' denotes Mathematics, and ``Logic.'' denotes Logical Reasoning.}
  \label{tab:dpo_alignbench}%
\end{table*}%

\begin{table*}[!t]
  \centering
  \tiny
    \begin{tabular}{c|l|ccc|ccccccc|c}
    \toprule
    \multirow{2}[4]{*}{\textbf{Model}} & \multicolumn{1}{c|}{\multirow{2}[4]{*}{\textbf{Dataset}}} & \multicolumn{3}{c|}{\textbf{Reasoning}} & \multicolumn{7}{c|}{\textbf{Language}}                & \multicolumn{1}{c}{\multirow{2}[4]{*}{\textbf{Overall}}} \\
\cmidrule{3-12}          &       & \multicolumn{1}{c}{\textbf{Math.}} & \multicolumn{1}{c}{\textbf{Logi.}} & \multicolumn{1}{c|}{\textbf{Avg.}} & \multicolumn{1}{c}{\textbf{Pro.}} & \multicolumn{1}{c}{\textbf{Chi.}} & \multicolumn{1}{c}{\textbf{Fund.}} & \multicolumn{1}{c}{\textbf{Writ.}} & \multicolumn{1}{c}{\textbf{Open.}} & \multicolumn{1}{c}{\textbf{Role.}} & \multicolumn{1}{c|}{\textbf{Avg.}} &  \\
    \midrule
    \multirow{3}[2]{*}{Qwen2.5-3B-SFT-Open} & \href{https://github.com/HIT-SCIR/huozi/tree/main/data/huozi-rlhf}{Huozi-RLHF} \citep{huozi}  & 4.45  & 3.30  & 3.88  & 4.75  & 4.21  & 4.72  & 5.48  & 5.45  & 5.64  & 5.04  & 4.46  \\
          & \href{https://huggingface.co/datasets/wenbopan/Chinese-dpo-pairs}{Chinese-DPO-Pairs} & 4.69  & 3.71  & 4.20  & 4.77  & \textbf{4.83} & 4.71  & 5.67  & 5.66  & 5.82  & 5.24  & 4.72  \\
          & \cellcolor[rgb]{ .906,  .902,  .902}TaP (GPT-4) & \cellcolor[rgb]{ .906,  .902,  .902}\textbf{5.10} & \cellcolor[rgb]{ .906,  .902,  .902}\textbf{4.49} & \cellcolor[rgb]{ .906,  .902,  .902}\textbf{4.79} & \cellcolor[rgb]{ .906,  .902,  .902}\textbf{5.56} & \cellcolor[rgb]{ .906,  .902,  .902}4.69  & \cellcolor[rgb]{ .906,  .902,  .902}\textbf{5.15} & \cellcolor[rgb]{ .906,  .902,  .902}\textbf{6.28} & \cellcolor[rgb]{ .906,  .902,  .902}\textbf{5.87} & \cellcolor[rgb]{ .906,  .902,  .902}\textbf{6.20} & \cellcolor[rgb]{ .906,  .902,  .902}\textbf{5.62} & \cellcolor[rgb]{ .906,  .902,  .902}\textbf{5.21} \\
    \midrule
    \multirow{3}[2]{*}{Qwen2.5-3B-SFT-TaP} & \href{https://github.com/HIT-SCIR/huozi/tree/main/data/huozi-rlhf}{Huozi-RLHF} \citep{huozi}  & 5.45  & 3.85  & 4.65  & \textbf{5.45} & \textbf{5.09} & \textbf{5.33} & 5.87  & 5.55  & 5.74  & 5.50  & 5.08  \\
          & \href{https://huggingface.co/datasets/wenbopan/Chinese-dpo-pairs}{Chinese-DPO-Pairs} & 5.61  & \textbf{4.20} & \textbf{4.90} & 5.44  & 4.64  & 5.25  & 5.89  & 5.63  & \textbf{6.18} & 5.51  & 5.20  \\
          & \cellcolor[rgb]{ .906,  .902,  .902}TaP (GPT-4) & \cellcolor[rgb]{ .906,  .902,  .902}\textbf{5.79} & \cellcolor[rgb]{ .906,  .902,  .902}4.02  & \cellcolor[rgb]{ .906,  .902,  .902}\textbf{4.90} & \cellcolor[rgb]{ .906,  .902,  .902}5.27  & \cellcolor[rgb]{ .906,  .902,  .902}4.88  & \cellcolor[rgb]{ .906,  .902,  .902}5.28  & \cellcolor[rgb]{ .906,  .902,  .902}\textbf{6.40} & \cellcolor[rgb]{ .906,  .902,  .902}\textbf{5.95} & \cellcolor[rgb]{ .906,  .902,  .902}6.14  & \cellcolor[rgb]{ .906,  .902,  .902}\textbf{5.65} & \cellcolor[rgb]{ .906,  .902,  .902}\textbf{5.28} \\
    \midrule
    
    \multirow{3}[2]{*}{Qwen2.5-7B-SFT-Open} & \href{https://github.com/HIT-SCIR/huozi/tree/main/data/huozi-rlhf}{Huozi-RLHF} \citep{huozi}  & 4.86  & 3.88  & 4.37  & \textbf{5.52} & 4.66  & 5.19  & 5.43  & 5.84  & 5.40  & 5.34  & 4.85  \\
          & \href{https://huggingface.co/datasets/wenbopan/Chinese-dpo-pairs}{Chinese-DPO-Pairs} & 5.15  & 4.13  & 4.64  & 5.47  & 4.57  & 5.27  & 5.75  & 5.84  & 5.51  & 5.40  & 5.02  \\
          & \cellcolor[rgb]{ .906,  .902,  .902}TaP (GPT-4) & \cellcolor[rgb]{ .906,  .902,  .902}\textbf{5.42} & \cellcolor[rgb]{ .906,  .902,  .902}\textbf{4.32} & \cellcolor[rgb]{ .906,  .902,  .902}\textbf{4.87} & \cellcolor[rgb]{ .906,  .902,  .902}5.35  & \cellcolor[rgb]{ .906,  .902,  .902}\textbf{5.09} & \cellcolor[rgb]{ .906,  .902,  .902}\textbf{5.28} & \cellcolor[rgb]{ .906,  .902,  .902}\textbf{5.91} & \cellcolor[rgb]{ .906,  .902,  .902}\textbf{5.92} & \cellcolor[rgb]{ .906,  .902,  .902}\textbf{5.92} & \cellcolor[rgb]{ .906,  .902,  .902}\textbf{5.58} & \cellcolor[rgb]{ .906,  .902,  .902}\textbf{5.22} \\
    \midrule
    \multirow{3}[2]{*}{Qwen2.5-7B-SFT-TaP} & \href{https://github.com/HIT-SCIR/huozi/tree/main/data/huozi-rlhf}{Huozi-RLHF} \citep{huozi}  & 6.79  & 4.88  & 5.83  & 5.86  & 5.36  & \textbf{6.34} & 6.09  & 6.11  & 6.26  & 6.00  & 5.92  \\
          & \href{https://huggingface.co/datasets/wenbopan/Chinese-dpo-pairs}{Chinese-DPO-Pairs} & 6.87  & 5.01  & 5.94  & 6.03  & \textbf{5.52} & 6.16  & 6.13  & 6.16  & 6.51  & 6.09  & 6.01  \\
          & \cellcolor[rgb]{ .906,  .902,  .902}TaP (GPT-4) & \cellcolor[rgb]{ .906,  .902,  .902}\textbf{6.92} & \cellcolor[rgb]{ .906,  .902,  .902}\textbf{5.14} & \cellcolor[rgb]{ .906,  .902,  .902}\textbf{6.03} & \cellcolor[rgb]{ .906,  .902,  .902}\textbf{6.28} & \cellcolor[rgb]{ .906,  .902,  .902}5.26  & \cellcolor[rgb]{ .906,  .902,  .902}6.24  & \cellcolor[rgb]{ .906,  .902,  .902}\textbf{6.43} & \cellcolor[rgb]{ .906,  .902,  .902}\textbf{6.32} & \cellcolor[rgb]{ .906,  .902,  .902}\textbf{6.59} & \cellcolor[rgb]{ .906,  .902,  .902}\textbf{6.18} & \cellcolor[rgb]{ .906,  .902,  .902}\textbf{6.11} \\
    \midrule
    \multirow{3}[2]{*}{Llama-3.1-8B-SFT-Open} & \href{https://github.com/HIT-SCIR/huozi/tree/main/data/huozi-rlhf}{Huozi-RLHF} \citep{huozi}  & 3.29  & 3.29  & 3.29  & 4.71  & 3.97  & 4.65  & 5.52  & 5.63  & 5.88  & 5.06  & 4.17  \\
          & \href{https://huggingface.co/datasets/wenbopan/Chinese-dpo-pairs}{Chinese-DPO-Pairs} & 3.49  & 3.10  & 3.29  & 4.91  & 4.05  & \textbf{4.87} & 5.85  & \textbf{6.00} & 6.08  & 5.29  & 4.29  \\
          & \cellcolor[rgb]{ .906,  .902,  .902}TaP (GPT-4) & \cellcolor[rgb]{ .906,  .902,  .902}\textbf{3.58} & \cellcolor[rgb]{ .906,  .902,  .902}\textbf{3.41} & \cellcolor[rgb]{ .906,  .902,  .902}\textbf{3.50} & \cellcolor[rgb]{ .906,  .902,  .902}\textbf{4.95} & \cellcolor[rgb]{ .906,  .902,  .902}\textbf{4.29} & \cellcolor[rgb]{ .906,  .902,  .902}4.76  & \cellcolor[rgb]{ .906,  .902,  .902}\textbf{6.28} & \cellcolor[rgb]{ .906,  .902,  .902}5.66  & \cellcolor[rgb]{ .906,  .902,  .902}\textbf{6.28} & \cellcolor[rgb]{ .906,  .902,  .902}\textbf{5.37} & \cellcolor[rgb]{ .906,  .902,  .902}\textbf{4.43} \\
    \midrule
    \multirow{3}[2]{*}{Llama-3.1-8B-SFT-TaP} & \href{https://github.com/HIT-SCIR/huozi/tree/main/data/huozi-rlhf}{Huozi-RLHF} \citep{huozi}  & 3.01  & \textbf{3.64} & 3.33  & 4.61  & 3.22  & 4.65  & 5.39  & 5.61  & 5.71  & 4.86  & 4.09  \\
          & \href{https://huggingface.co/datasets/wenbopan/Chinese-dpo-pairs}{Chinese-DPO-Pairs} & \textbf{3.54} & 3.63  & \textbf{3.59} & \textbf{4.73} & \textbf{3.69} & \textbf{4.71} & 5.59  & 5.66  & 6.08  & \textbf{5.07} & \textbf{4.33} \\
          & \cellcolor[rgb]{ .906,  .902,  .902}TaP (GPT-4) & \cellcolor[rgb]{ .906,  .902,  .902}3.46  & \cellcolor[rgb]{ .906,  .902,  .902}3.58  & \cellcolor[rgb]{ .906,  .902,  .902}3.52  & \cellcolor[rgb]{ .906,  .902,  .902}4.52  & \cellcolor[rgb]{ .906,  .902,  .902}3.26  & \cellcolor[rgb]{ .906,  .902,  .902}4.49  & \cellcolor[rgb]{ .906,  .902,  .902}\textbf{5.73} & \cellcolor[rgb]{ .906,  .902,  .902}\textbf{5.82} & \cellcolor[rgb]{ .906,  .902,  .902}\textbf{6.10} & \cellcolor[rgb]{ .906,  .902,  .902}4.99  & \cellcolor[rgb]{ .906,  .902,  .902}4.25  \\
    \midrule
    \multirow{3}[2]{*}{Gemma-2-9B-SFT-Open} & \href{https://github.com/HIT-SCIR/huozi/tree/main/data/huozi-rlhf}{Huozi-RLHF} \citep{huozi}  & 3.72  & 3.35  & 3.54  & 4.81  & 3.72  & 4.99  & 5.49  & 5.39  & 5.90  & 5.05  & 4.29  \\
          & \href{https://huggingface.co/datasets/wenbopan/Chinese-dpo-pairs}{Chinese-DPO-Pairs} & 3.77  & 3.46  & 3.61  & 5.09  & 3.83  & 5.09  & 5.60  & 5.76  & 5.87  & 5.21  & 4.41  \\
          & \cellcolor[rgb]{ .906,  .902,  .902}TaP (GPT-4) & \cellcolor[rgb]{ .906,  .902,  .902}\textbf{4.16} & \cellcolor[rgb]{ .906,  .902,  .902}\textbf{4.01} & \cellcolor[rgb]{ .906,  .902,  .902}\textbf{4.09} & \cellcolor[rgb]{ .906,  .902,  .902}\textbf{5.17} & \cellcolor[rgb]{ .906,  .902,  .902}\textbf{3.98} & \cellcolor[rgb]{ .906,  .902,  .902}\textbf{5.32} & \cellcolor[rgb]{ .906,  .902,  .902}\textbf{5.96} & \cellcolor[rgb]{ .906,  .902,  .902}\textbf{5.87} & \cellcolor[rgb]{ .906,  .902,  .902}\textbf{6.33} & \cellcolor[rgb]{ .906,  .902,  .902}\textbf{5.44} & \cellcolor[rgb]{ .906,  .902,  .902}\textbf{4.76} \\
    \midrule
    \multirow{3}[2]{*}{Gemma-2-9B-SFT-TaP} & \href{https://github.com/HIT-SCIR/huozi/tree/main/data/huozi-rlhf}{Huozi-RLHF} \citep{huozi}  & 4.37  & 3.17  & 3.77  & 4.69  & \textbf{3.43} & 5.25  & 5.32  & 5.34  & 5.56  & 4.93  & 4.35  \\
          & \href{https://huggingface.co/datasets/wenbopan/Chinese-dpo-pairs}{Chinese-DPO-Pairs} & \textbf{4.54} & \textbf{3.64} & \textbf{4.09} & \textbf{5.15} & \textbf{3.43} & \textbf{5.41} & 5.57  & 5.42  & \textbf{6.24} & \textbf{5.21} & \textbf{4.65} \\
          & \cellcolor[rgb]{ .906,  .902,  .902}TaP (GPT-4) & \cellcolor[rgb]{ .906,  .902,  .902}4.49  & \cellcolor[rgb]{ .906,  .902,  .902}3.61  & \cellcolor[rgb]{ .906,  .902,  .902}4.05  & \cellcolor[rgb]{ .906,  .902,  .902}4.85  & \cellcolor[rgb]{ .906,  .902,  .902}3.41  & \cellcolor[rgb]{ .906,  .902,  .902}5.18  & \cellcolor[rgb]{ .906,  .902,  .902}\textbf{5.88} & \cellcolor[rgb]{ .906,  .902,  .902}\textbf{5.55} & \cellcolor[rgb]{ .906,  .902,  .902}5.99  & \cellcolor[rgb]{ .906,  .902,  .902}5.14  & \cellcolor[rgb]{ .906,  .902,  .902}4.60  \\
    \midrule
    \multirow{3}[2]{*}{Qwen2.5-14B-SFT-Open} & \href{https://github.com/HIT-SCIR/huozi/tree/main/data/huozi-rlhf}{Huozi-RLHF} \citep{huozi}  & 6.12  & 5.17  & 5.64  & 5.98  & 5.64  & 5.90  & 5.51  & 5.87  & 5.97  & 5.81  & 5.73  \\
          & \href{https://huggingface.co/datasets/wenbopan/Chinese-dpo-pairs}{Chinese-DPO-Pairs} & 6.05  & 4.89  & 5.47  & 6.04  & 5.43  & 5.88  & 5.63  & 5.87  & 5.91  & 5.79  & 5.63  \\
          & \cellcolor[rgb]{ .906,  .902,  .902}TaP (GPT-4) & \cellcolor[rgb]{ .906,  .902,  .902}\textbf{6.77} & \cellcolor[rgb]{ .906,  .902,  .902}\textbf{5.34} & \cellcolor[rgb]{ .906,  .902,  .902}\textbf{6.05} & \cellcolor[rgb]{ .906,  .902,  .902}\textbf{6.37} & \cellcolor[rgb]{ .906,  .902,  .902}\textbf{5.74} & \cellcolor[rgb]{ .906,  .902,  .902}\textbf{6.26} & \cellcolor[rgb]{ .906,  .902,  .902}\textbf{6.04} & \cellcolor[rgb]{ .906,  .902,  .902}\textbf{6.21} & \cellcolor[rgb]{ .906,  .902,  .902}\textbf{6.41} & \cellcolor[rgb]{ .906,  .902,  .902}\textbf{6.17} & \cellcolor[rgb]{ .906,  .902,  .902}\textbf{6.11} \\
    \midrule
    \multirow{3}[2]{*}{Qwen2.5-14B-SFT-TaP} & \href{https://github.com/HIT-SCIR/huozi/tree/main/data/huozi-rlhf}{Huozi-RLHF} \citep{huozi}  & 7.19  & 5.64  & 6.41  & 6.41  & 5.83  & 6.60  & 6.19  & 6.24  & 6.43  & 6.28  & 6.35  \\
          & \href{https://huggingface.co/datasets/wenbopan/Chinese-dpo-pairs}{Chinese-DPO-Pairs} & \textbf{7.44} & 5.38  & 6.41  & \textbf{6.57} & 5.79  & \textbf{6.65} & 6.51  & 6.32  & 6.74  & 6.43  & 6.42  \\
          & \cellcolor[rgb]{ .906,  .902,  .902}TaP (GPT-4) & \cellcolor[rgb]{ .906,  .902,  .902}7.31  & \cellcolor[rgb]{ .906,  .902,  .902}\textbf{5.73} & \cellcolor[rgb]{ .906,  .902,  .902}\textbf{6.52} & \cellcolor[rgb]{ .906,  .902,  .902}6.43  & \cellcolor[rgb]{ .906,  .902,  .902}\textbf{5.91} & \cellcolor[rgb]{ .906,  .902,  .902}6.63  & \cellcolor[rgb]{ .906,  .902,  .902}\textbf{6.77} & \cellcolor[rgb]{ .906,  .902,  .902}\textbf{6.42} & \cellcolor[rgb]{ .906,  .902,  .902}\textbf{6.96} & \cellcolor[rgb]{ .906,  .902,  .902}\textbf{6.52} & \cellcolor[rgb]{ .906,  .902,  .902}\textbf{6.52} \\
    \bottomrule
    \end{tabular}%
  \caption{Performance comparison of LLMs trained with \textbf{DPO} using different datasets on AlignBench. The model names include two possible suffixes: ``Open'' and ``TaP.'' The ``Open'' suffix indicates that the LLMs were initialized from models trained via supervised fine-tuning on open-source datasets, whereas ``TaP'' denotes initialization from models trained on a dataset constructed by TaP. The evaluation is conducted using \textbf{DeepSeek-V3}, which scores the models’ responses. The ``Fund.'' column denotes Fundamental Language Ability, ``Chi.'' denotes Advanced Chinese Understanding, ``Open.'' denotes Open-ended Questions, ``Writ.'' denotes Writing Ability, ``Role.'' denotes Task-oriented Role Play, ``Pro'' denotes ``Professional Knowledge'', ``Math.'' denotes Mathematics, and ``Logic.'' denotes Logical Reasoning.}
  \label{tab:dpo_alignbench_deepseek}%
\end{table*}%

\begin{table*}[!t]
  \centering
  \tiny
    \begin{tabular}{c|l|ccc|ccccccc|c}
    \toprule
    \multirow{2}[4]{*}{\textbf{Model}} & \multicolumn{1}{c|}{\multirow{2}[4]{*}{\textbf{Dataset}}} & \multicolumn{3}{c|}{\textbf{Reasoning}} & \multicolumn{7}{c|}{\textbf{Language}}                & \multicolumn{1}{c}{\multirow{2}[4]{*}{\textbf{Overall}}} \\
    \cmidrule{3-12}          &       & \multicolumn{1}{c}{\textbf{Math.}} & \multicolumn{1}{c}{\textbf{Logi.}} & \multicolumn{1}{c|}{\textbf{Avg.}} & \multicolumn{1}{c}{\textbf{Pro.}} & \multicolumn{1}{c}{\textbf{Chi.}} & \multicolumn{1}{c}{\textbf{Fund.}} & \multicolumn{1}{c}{\textbf{Writ.}} & \multicolumn{1}{c}{\textbf{Open.}} & \multicolumn{1}{c}{\textbf{Role.}} & \multicolumn{1}{c|}{\textbf{Avg.}} &  \\
    \midrule
    \multirow{3}[2]{*}{Qwen2.5-3B-SFT-Open} & \href{https://github.com/HIT-SCIR/huozi/tree/main/data/huozi-rlhf}{Huozi-RLHF} \citep{huozi}  & 4.62  & 3.65  & 4.13  & 4.45  & 4.71  & 4.57  & 5.03  & 5.18  & 5.36  & 4.88  & 4.51  \\
          & \href{https://huggingface.co/datasets/wenbopan/Chinese-dpo-pairs}{Chinese-DPO-Pairs} & 5.82  & \textbf{4.42} & 5.12  & \textbf{5.83} & 5.22  & 5.07  & 6.43  & \textbf{6.89} & 6.41  & 5.98  & 5.55  \\
          & \cellcolor[rgb]{ .906,  .902,  .902}TaP (GPT-4) & \cellcolor[rgb]{ .906,  .902,  .902}\textbf{6.24} & \cellcolor[rgb]{ .906,  .902,  .902}4.16  & \cellcolor[rgb]{ .906,  .902,  .902}\textbf{5.20} & \cellcolor[rgb]{ .906,  .902,  .902}5.73  & \cellcolor[rgb]{ .906,  .902,  .902}\textbf{5.50} & \cellcolor[rgb]{ .906,  .902,  .902}\textbf{5.61} & \cellcolor[rgb]{ .906,  .902,  .902}\textbf{6.49} & \cellcolor[rgb]{ .906,  .902,  .902}6.71  & \cellcolor[rgb]{ .906,  .902,  .902}\textbf{6.60} & \cellcolor[rgb]{ .906,  .902,  .902}\textbf{6.11} & \cellcolor[rgb]{ .906,  .902,  .902}\textbf{5.66} \\
    \midrule
    \multirow{3}[2]{*}{Qwen2.5-3B-SFT-TaP} & \href{https://github.com/HIT-SCIR/huozi/tree/main/data/huozi-rlhf}{Huozi-RLHF} \citep{huozi}  & 5.59  & 3.63  & 4.61  & 4.97  & 4.79  & 5.07  & 5.48  & 5.79  & 5.53  & 5.27  & 4.94  \\
          & \href{https://huggingface.co/datasets/wenbopan/Chinese-dpo-pairs}{Chinese-DPO-Pairs} & 5.82  & 3.93  & 4.88  & 5.17  & 5.41  & 5.13  & 5.68  & 5.97  & 5.71  & 5.51  & 5.20  \\
          & \cellcolor[rgb]{ .906,  .902,  .902}TaP (GPT-4) & \cellcolor[rgb]{ .906,  .902,  .902}\textbf{6.49} & \cellcolor[rgb]{ .906,  .902,  .902}\textbf{4.89} & \cellcolor[rgb]{ .906,  .902,  .902}\textbf{5.69} & \cellcolor[rgb]{ .906,  .902,  .902}\textbf{5.93} & \cellcolor[rgb]{ .906,  .902,  .902}\textbf{6.12} & \cellcolor[rgb]{ .906,  .902,  .902}\textbf{5.53} & \cellcolor[rgb]{ .906,  .902,  .902}\textbf{6.55} & \cellcolor[rgb]{ .906,  .902,  .902}\textbf{6.89} & \cellcolor[rgb]{ .906,  .902,  .902}\textbf{6.59} & \cellcolor[rgb]{ .906,  .902,  .902}\textbf{6.27} & \cellcolor[rgb]{ .906,  .902,  .902}\textbf{5.98} \\
    \midrule
    
    \multirow{3}[2]{*}{Qwen2.5-7B-SFT-Open} & \href{https://github.com/HIT-SCIR/huozi/tree/main/data/huozi-rlhf}{Huozi-RLHF} \citep{huozi}  & 5.04  & 4.39  & 4.71  & 5.60  & 5.00  & \textbf{5.68} & 5.61  & 5.92  & 5.53  & 5.56  & 5.13  \\
          & \href{https://huggingface.co/datasets/wenbopan/Chinese-dpo-pairs}{Chinese-DPO-Pairs} & 5.03  & 3.79  & 4.41  & 5.15  & 5.18  & 5.28  & 5.72  & 5.32  & 5.34  & 5.33  & 4.87  \\
          & \cellcolor[rgb]{ .906,  .902,  .902}TaP (GPT-4) & \cellcolor[rgb]{ .906,  .902,  .902}\textbf{5.29} & \cellcolor[rgb]{ .906,  .902,  .902}\textbf{4.60} & \cellcolor[rgb]{ .906,  .902,  .902}\textbf{4.94} & \cellcolor[rgb]{ .906,  .902,  .902}\textbf{5.66} & \cellcolor[rgb]{ .906,  .902,  .902}\textbf{5.31} & \cellcolor[rgb]{ .906,  .902,  .902}5.50  & \cellcolor[rgb]{ .906,  .902,  .902}\textbf{6.28} & \cellcolor[rgb]{ .906,  .902,  .902}\textbf{6.29} & \cellcolor[rgb]{ .906,  .902,  .902}\textbf{5.83} & \cellcolor[rgb]{ .906,  .902,  .902}\textbf{5.81} & \cellcolor[rgb]{ .906,  .902,  .902}\textbf{5.38} \\
    \midrule
    \multirow{3}[2]{*}{Qwen2.5-7B-SFT-TaP} & \href{https://github.com/HIT-SCIR/huozi/tree/main/data/huozi-rlhf}{Huozi-RLHF} \citep{huozi}  & \textbf{7.11} & 5.38  & 6.24  & 6.45  & \textbf{6.19} & 6.26  & 6.40  & 6.89  & 6.46  & 6.44  & 6.34  \\
          & \href{https://huggingface.co/datasets/wenbopan/Chinese-dpo-pairs}{Chinese-DPO-Pairs} & 6.82  & 4.75  & 5.79  & 5.73  & 5.67  & 5.53  & 5.76  & 7.11  & 6.34  & 6.02  & 5.90  \\
          & \cellcolor[rgb]{ .906,  .902,  .902}TaP (GPT-4) & \cellcolor[rgb]{ .906,  .902,  .902}7.05  & \cellcolor[rgb]{ .906,  .902,  .902}\textbf{5.53} & \cellcolor[rgb]{ .906,  .902,  .902}\textbf{6.29} & \cellcolor[rgb]{ .906,  .902,  .902}\textbf{6.77} & \cellcolor[rgb]{ .906,  .902,  .902}5.91  & \cellcolor[rgb]{ .906,  .902,  .902}\textbf{6.35} & \cellcolor[rgb]{ .906,  .902,  .902}\textbf{6.81} & \cellcolor[rgb]{ .906,  .902,  .902}\textbf{7.13} & \cellcolor[rgb]{ .906,  .902,  .902}\textbf{6.75} & \cellcolor[rgb]{ .906,  .902,  .902}\textbf{6.62} & \cellcolor[rgb]{ .906,  .902,  .902}\textbf{6.46} \\
    \midrule
    \multirow{3}[2]{*}{Llama-3.1-8B-SFT-Open} & \href{https://github.com/HIT-SCIR/huozi/tree/main/data/huozi-rlhf}{Huozi-RLHF} \citep{huozi}  & 3.78  & \textbf{3.66 } & 3.72  & 5.02  & 4.36  & \textbf{4.91 } & 5.67  & 5.92  & 5.74  & 5.27  & 4.50  \\
          & \href{https://huggingface.co/datasets/wenbopan/Chinese-dpo-pairs}{Chinese-DPO-Pairs} & 2.76  & 2.83  & 2.79  & 4.26  & 3.48  & 4.46  & 4.83  & 4.82  & 5.29  & 4.52  & 3.66  \\
          & \cellcolor[rgb]{ .906,  .902,  .902}TaP (GPT-4) & \cellcolor[rgb]{ .906,  .902,  .902}\textbf{3.90} & \cellcolor[rgb]{ .906,  .902,  .902}3.65  & \cellcolor[rgb]{ .906,  .902,  .902}\textbf{3.78} & \cellcolor[rgb]{ .906,  .902,  .902}\textbf{5.17} & \cellcolor[rgb]{ .906,  .902,  .902}\textbf{4.67} & \cellcolor[rgb]{ .906,  .902,  .902}4.88  & \cellcolor[rgb]{ .906,  .902,  .902}\textbf{6.31} & \cellcolor[rgb]{ .906,  .902,  .902}\textbf{6.21} & \cellcolor[rgb]{ .906,  .902,  .902}\textbf{6.20} & \cellcolor[rgb]{ .906,  .902,  .902}\textbf{5.57} & \cellcolor[rgb]{ .906,  .902,  .902}\textbf{4.68} \\
    \midrule
    \multirow{3}[2]{*}{Llama-3.1-8B-SFT-TaP} & \href{https://github.com/HIT-SCIR/huozi/tree/main/data/huozi-rlhf}{Huozi-RLHF} \citep{huozi}  & 3.81  & 4.08  & 3.94  & 4.81  & 4.21  & 5.09  & 5.73  & 5.89  & 5.78  & 5.25  & 4.60  \\
          & \href{https://huggingface.co/datasets/wenbopan/Chinese-dpo-pairs}{Chinese-DPO-Pairs} & 3.56  & 3.88  & 3.72  & 5.00  & 4.05  & 4.88  & 5.73  & 5.50  & 5.35  & 5.09  & 4.40  \\
          & \cellcolor[rgb]{ .906,  .902,  .902}TaP (GPT-4) & \cellcolor[rgb]{ .906,  .902,  .902}\textbf{3.92} & \cellcolor[rgb]{ .906,  .902,  .902}\textbf{4.15} & \cellcolor[rgb]{ .906,  .902,  .902}\textbf{4.04} & \cellcolor[rgb]{ .906,  .902,  .902}\textbf{5.05} & \cellcolor[rgb]{ .906,  .902,  .902}\textbf{4.24} & \cellcolor[rgb]{ .906,  .902,  .902}\textbf{5.12} & \cellcolor[rgb]{ .906,  .902,  .902}\textbf{6.33} & \cellcolor[rgb]{ .906,  .902,  .902}\textbf{6.53} & \cellcolor[rgb]{ .906,  .902,  .902}\textbf{6.29} & \cellcolor[rgb]{ .906,  .902,  .902}\textbf{5.59} & \cellcolor[rgb]{ .906,  .902,  .902}\textbf{4.81} \\
    \midrule
    \multirow{3}[2]{*}{Gemma-2-9B-SFT-Open} & \href{https://github.com/HIT-SCIR/huozi/tree/main/data/huozi-rlhf}{Huozi-RLHF} \citep{huozi}  & 4.23  & 3.26  & 3.75  & 5.09  & 3.91  & 4.44  & 4.88  & 5.03  & 5.16  & 4.75  & 4.25  \\
          & \href{https://huggingface.co/datasets/wenbopan/Chinese-dpo-pairs}{Chinese-DPO-Pairs} & 4.38  & 3.73  & 4.05  & 4.89  & 4.63  & 4.97  & 6.15  & 6.24  & 6.01  & 5.48  & 4.77  \\
          & \cellcolor[rgb]{ .906,  .902,  .902}TaP (GPT-4) & \cellcolor[rgb]{ .906,  .902,  .902}\textbf{4.85} & \cellcolor[rgb]{ .906,  .902,  .902}\textbf{4.14} & \cellcolor[rgb]{ .906,  .902,  .902}\textbf{4.49} & \cellcolor[rgb]{ .906,  .902,  .902}\textbf{5.75} & \cellcolor[rgb]{ .906,  .902,  .902}\textbf{4.66} & \cellcolor[rgb]{ .906,  .902,  .902}\textbf{5.45} & \cellcolor[rgb]{ .906,  .902,  .902}\textbf{6.69} & \cellcolor[rgb]{ .906,  .902,  .902}\textbf{6.34} & \cellcolor[rgb]{ .906,  .902,  .902}\textbf{6.45} & \cellcolor[rgb]{ .906,  .902,  .902}\textbf{5.89} & \cellcolor[rgb]{ .906,  .902,  .902}\textbf{5.19} \\
    \midrule
    \multirow{3}[2]{*}{Gemma-2-9B-SFT-TaP} & \href{https://github.com/HIT-SCIR/huozi/tree/main/data/huozi-rlhf}{Huozi-RLHF} \citep{huozi}  & 4.13  & 3.57  & 3.85  & 5.04  & \textbf{4.53} & 5.13  & 5.53  & 5.26  & 5.73  & 5.20  & 4.53  \\
          & \href{https://huggingface.co/datasets/wenbopan/Chinese-dpo-pairs}{Chinese-DPO-Pairs} & 4.52  & 3.77  & 4.14  & 5.02  & 4.12  & \textbf{5.24} & 5.48  & 6.05  & 5.98  & 5.32  & 4.73  \\
          & \cellcolor[rgb]{ .906,  .902,  .902}TaP (GPT-4) & \cellcolor[rgb]{ .906,  .902,  .902}\textbf{4.83} & \cellcolor[rgb]{ .906,  .902,  .902}\textbf{4.00} & \cellcolor[rgb]{ .906,  .902,  .902}\textbf{4.42} & \cellcolor[rgb]{ .906,  .902,  .902}\textbf{5.35} & \cellcolor[rgb]{ .906,  .902,  .902}4.14  & \cellcolor[rgb]{ .906,  .902,  .902}5.15  & \cellcolor[rgb]{ .906,  .902,  .902}\textbf{6.14} & \cellcolor[rgb]{ .906,  .902,  .902}\textbf{6.47} & \cellcolor[rgb]{ .906,  .902,  .902}\textbf{6.24} & \cellcolor[rgb]{ .906,  .902,  .902}\textbf{5.58} & \cellcolor[rgb]{ .906,  .902,  .902}\textbf{5.00} \\
    \midrule
    \multirow{3}[2]{*}{Qwen2.5-14B-SFT-Open} & \href{https://github.com/HIT-SCIR/huozi/tree/main/data/huozi-rlhf}{Huozi-RLHF} \citep{huozi}  & 6.03  & 5.18  & 5.61  & 5.50  & 6.05  & 6.35  & 5.73  & 5.82  & 5.93  & 5.90  & 5.75  \\
          & \href{https://huggingface.co/datasets/wenbopan/Chinese-dpo-pairs}{Chinese-DPO-Pairs} & 6.60  & 5.51  & 6.05  & 6.58  & 5.81  & \textbf{6.46} & 6.27  & 6.42  & 6.24  & 6.30  & 6.18  \\
          & \cellcolor[rgb]{ .906,  .902,  .902}TaP (GPT-4) & \cellcolor[rgb]{ .906,  .902,  .902}\textbf{7.10} & \cellcolor[rgb]{ .906,  .902,  .902}\textbf{5.96} & \cellcolor[rgb]{ .906,  .902,  .902}\textbf{6.53} & \cellcolor[rgb]{ .906,  .902,  .902}\textbf{6.77} & \cellcolor[rgb]{ .906,  .902,  .902}\textbf{6.41} & \cellcolor[rgb]{ .906,  .902,  .902}6.06  & \cellcolor[rgb]{ .906,  .902,  .902}\textbf{6.52} & \cellcolor[rgb]{ .906,  .902,  .902}\textbf{6.74} & \cellcolor[rgb]{ .906,  .902,  .902}\textbf{6.43} & \cellcolor[rgb]{ .906,  .902,  .902}\textbf{6.49} & \cellcolor[rgb]{ .906,  .902,  .902}\textbf{6.51} \\
    \midrule
    \multirow{3}[2]{*}{Qwen2.5-14B-SFT-TaP} & \href{https://github.com/HIT-SCIR/huozi/tree/main/data/huozi-rlhf}{Huozi-RLHF} \citep{huozi}  & 6.79  & 6.43  & 6.61  & 6.07  & 6.38  & 6.22  & 6.59  & 7.03  & 6.69  & 6.50  & 6.55  \\
          & \href{https://huggingface.co/datasets/wenbopan/Chinese-dpo-pairs}{Chinese-DPO-Pairs} & 7.32  & 5.42  & 6.37  & 6.22  & 6.22  & 6.59  & 6.43  & 6.58  & 6.34  & 6.40  & 6.38  \\
          & \cellcolor[rgb]{ .906,  .902,  .902}TaP (GPT-4) & \cellcolor[rgb]{ .906,  .902,  .902}\textbf{8.07} & \cellcolor[rgb]{ .906,  .902,  .902}\textbf{6.76} & \cellcolor[rgb]{ .906,  .902,  .902}\textbf{7.42} & \cellcolor[rgb]{ .906,  .902,  .902}\textbf{7.30} & \cellcolor[rgb]{ .906,  .902,  .902}\textbf{6.74} & \cellcolor[rgb]{ .906,  .902,  .902}\textbf{6.81} & \cellcolor[rgb]{ .906,  .902,  .902}\textbf{7.19} & \cellcolor[rgb]{ .906,  .902,  .902}\textbf{7.42} & \cellcolor[rgb]{ .906,  .902,  .902}\textbf{7.27} & \cellcolor[rgb]{ .906,  .902,  .902}\textbf{7.12} & \cellcolor[rgb]{ .906,  .902,  .902}\textbf{7.27} \\
    \bottomrule
    \end{tabular}%
  \caption{Performance comparison of LLMs trained with \textbf{PPO} using different datasets on AlignBench. The model names include two possible suffixes: ``Open'' and ``TaP.'' The ``Open'' suffix indicates that the LLMs were initialized from models trained via supervised fine-tuning on open-source datasets, whereas ``TaP'' denotes initialization from models trained on a dataset constructed by TaP. The evaluation is conducted using \textbf{GPT-4o}, which scores the models’ responses. The ``Fund.'' column denotes Fundamental Language Ability, ``Chi.'' denotes Advanced Chinese Understanding, ``Open.'' denotes Open-ended Questions, ``Writ.'' denotes Writing Ability, ``Role.'' denotes Task-oriented Role Play, ``Pro'' denotes ``Professional Knowledge'', ``Math.'' denotes Mathematics, and ``Logic.'' denotes Logical Reasoning.}
  \label{tab:ppo_alignbench}%
\end{table*}%

\begin{table*}[!t]
  \centering
  \tiny
    \begin{tabular}{c|l|ccc|ccccccc|c}
    \toprule
    \multirow{2}[4]{*}{\textbf{Model}} & \multicolumn{1}{c|}{\multirow{2}[4]{*}{\textbf{Dataset}}} & \multicolumn{3}{c|}{\textbf{Reasoning}} & \multicolumn{7}{c|}{\textbf{Language}}                & \multicolumn{1}{c}{\multirow{2}[4]{*}{\textbf{Overall}}} \\
\cmidrule{3-12}          &       & \multicolumn{1}{c}{\textbf{Math.}} & \multicolumn{1}{c}{\textbf{Logi.}} & \multicolumn{1}{c|}{\textbf{Avg.}} & \multicolumn{1}{c}{\textbf{Pro.}} & \multicolumn{1}{c}{\textbf{Chi.}} & \multicolumn{1}{c}{\textbf{Fund.}} & \multicolumn{1}{c}{\textbf{Writ.}} & \multicolumn{1}{c}{\textbf{Open.}} & \multicolumn{1}{c}{\textbf{Role.}} & \multicolumn{1}{c|}{\textbf{Avg.}} &  \\
    \midrule
    \multirow{3}[2]{*}{Qwen2.5-3B-SFT-Open} & \href{https://github.com/HIT-SCIR/huozi/tree/main/data/huozi-rlhf}{Huozi-RLHF} \citep{huozi}  & 4.93  & \textbf{3.96} & 4.44  & 5.10  & \textbf{4.81} & 4.72  & 5.68  & 5.76  & 6.07  & 5.36  & 4.90  \\
          & \href{https://huggingface.co/datasets/wenbopan/Chinese-dpo-pairs}{Chinese-DPO-Pairs} & 5.14  & 3.91  & 4.53  & \textbf{5.31} & 4.64  & 4.85  & 6.12  & \textbf{6.13} & 6.06  & 5.52  & 5.02  \\
          & \cellcolor[rgb]{ .906,  .902,  .902}TaP (GPT-4) & \cellcolor[rgb]{ .906,  .902,  .902}\textbf{5.61} & \cellcolor[rgb]{ .906,  .902,  .902}3.82  & \cellcolor[rgb]{ .906,  .902,  .902}\textbf{4.71} & \cellcolor[rgb]{ .906,  .902,  .902}5.29  & \cellcolor[rgb]{ .906,  .902,  .902}4.71  & \cellcolor[rgb]{ .906,  .902,  .902}\textbf{5.21} & \cellcolor[rgb]{ .906,  .902,  .902}\textbf{6.16} & \cellcolor[rgb]{ .906,  .902,  .902}5.97  & \cellcolor[rgb]{ .906,  .902,  .902}\textbf{6.40} & \cellcolor[rgb]{ .906,  .902,  .902}\textbf{5.62} & \cellcolor[rgb]{ .906,  .902,  .902}\textbf{5.17} \\
    \midrule
    \multirow{3}[2]{*}{Qwen2.5-3B-SFT-TaP} & \href{https://github.com/HIT-SCIR/huozi/tree/main/data/huozi-rlhf}{Huozi-RLHF} \citep{huozi}  & 5.33  & 3.90  & 4.62  & 5.28  & 4.95  & \textbf{5.37} & 5.81  & 5.89  & 6.00  & 5.55  & 5.08  \\
          & \href{https://huggingface.co/datasets/wenbopan/Chinese-dpo-pairs}{Chinese-DPO-Pairs} & 5.83  & \textbf{4.30} & \textbf{5.07} & \textbf{5.57} & \textbf{5.43} & 5.25  & \textbf{6.27} & 5.92  & \textbf{6.16} & \textbf{5.77} & \textbf{5.42} \\
          & \cellcolor[rgb]{ .906,  .902,  .902}TaP (GPT-4) & \cellcolor[rgb]{ .906,  .902,  .902}\textbf{5.89} & \cellcolor[rgb]{ .906,  .902,  .902}4.07  & \cellcolor[rgb]{ .906,  .902,  .902}4.98  & \cellcolor[rgb]{ .906,  .902,  .902}5.53  & \cellcolor[rgb]{ .906,  .902,  .902}5.32  & \cellcolor[rgb]{ .906,  .902,  .902}5.18  & \cellcolor[rgb]{ .906,  .902,  .902}5.95  & \cellcolor[rgb]{ .906,  .902,  .902}\textbf{6.03} & \cellcolor[rgb]{ .906,  .902,  .902}6.08  & \cellcolor[rgb]{ .906,  .902,  .902}5.68  & \cellcolor[rgb]{ .906,  .902,  .902}5.33  \\
    \midrule
    \multirow{3}[2]{*}{Qwen2.5-7B-SFT-Open} & \href{https://github.com/HIT-SCIR/huozi/tree/main/data/huozi-rlhf}{Huozi-RLHF} \citep{huozi}  & 4.85  & 4.08  & 4.46  & 5.19  & 4.55  & \textbf{5.60} & 5.67  & \textbf{5.84} & 5.72  & 5.43  & 4.95  \\
          & \href{https://huggingface.co/datasets/wenbopan/Chinese-dpo-pairs}{Chinese-DPO-Pairs} & 4.82  & 3.79  & 4.31  & \textbf{5.48} & \textbf{4.93} & 5.45  & \textbf{5.69} & 5.79  & \textbf{5.78} & \textbf{5.52} & 4.91  \\
          & \cellcolor[rgb]{ .906,  .902,  .902}TaP (GPT-4) & \cellcolor[rgb]{ .906,  .902,  .902}\textbf{5.03} & \cellcolor[rgb]{ .906,  .902,  .902}\textbf{4.25} & \cellcolor[rgb]{ .906,  .902,  .902}\textbf{4.64} & \cellcolor[rgb]{ .906,  .902,  .902}5.39  & \cellcolor[rgb]{ .906,  .902,  .902}4.84  & \cellcolor[rgb]{ .906,  .902,  .902}5.09  & \cellcolor[rgb]{ .906,  .902,  .902}5.68  & \cellcolor[rgb]{ .906,  .902,  .902}5.82  & \cellcolor[rgb]{ .906,  .902,  .902}5.66  & \cellcolor[rgb]{ .906,  .902,  .902}5.41  & \cellcolor[rgb]{ .906,  .902,  .902}\textbf{5.03} \\
    \midrule
    \multirow{3}[2]{*}{Qwen2.5-7B-SFT-TaP} & \href{https://github.com/HIT-SCIR/huozi/tree/main/data/huozi-rlhf}{Huozi-RLHF} \citep{huozi}  & 6.67  & 4.67  & 5.67  & 6.04  & 5.52  & 5.69  & 6.03  & 6.26  & 6.32  & 5.98  & 5.82  \\
          & \href{https://huggingface.co/datasets/wenbopan/Chinese-dpo-pairs}{Chinese-DPO-Pairs} & \textbf{6.87} & 4.92  & \textbf{5.89} & 5.97  & \textbf{5.62} & 5.85  & 6.20  & 6.29  & 6.34  & 6.04  & \textbf{5.97} \\
          & \cellcolor[rgb]{ .906,  .902,  .902}TaP (GPT-4) & \cellcolor[rgb]{ .906,  .902,  .902}6.51  & \cellcolor[rgb]{ .906,  .902,  .902}\textbf{4.95} & \cellcolor[rgb]{ .906,  .902,  .902}5.73  & \cellcolor[rgb]{ .906,  .902,  .902}\textbf{6.36} & \cellcolor[rgb]{ .906,  .902,  .902}5.26  & \cellcolor[rgb]{ .906,  .902,  .902}\textbf{6.21} & \cellcolor[rgb]{ .906,  .902,  .902}\textbf{6.25} & \cellcolor[rgb]{ .906,  .902,  .902}\textbf{6.35} & \cellcolor[rgb]{ .906,  .902,  .902}\textbf{6.49} & \cellcolor[rgb]{ .906,  .902,  .902}\textbf{6.15} & \cellcolor[rgb]{ .906,  .902,  .902}5.94  \\
    \midrule
    \multirow{3}[2]{*}{Llama-3.1-8B-SFT-Open} & \href{https://github.com/HIT-SCIR/huozi/tree/main/data/huozi-rlhf}{Huozi-RLHF} \citep{huozi}  & \textbf{3.54} & \textbf{3.27} & \textbf{3.41} & 4.81  & 3.79  & 4.43  & 5.16  & 5.82  & 5.80  & 4.97  & 4.19  \\
          & \href{https://huggingface.co/datasets/wenbopan/Chinese-dpo-pairs}{Chinese-DPO-Pairs} & 3.10  & 3.11  & 3.10  & 4.91  & 3.83  & \textbf{4.78} & 5.49  & 5.26  & 5.88  & 5.03  & 4.06  \\
          & \cellcolor[rgb]{ .906,  .902,  .902}TaP (GPT-4) & \cellcolor[rgb]{ .906,  .902,  .902}3.43  & \cellcolor[rgb]{ .906,  .902,  .902}3.23  & \cellcolor[rgb]{ .906,  .902,  .902}3.33  & \cellcolor[rgb]{ .906,  .902,  .902}\textbf{4.99} & \cellcolor[rgb]{ .906,  .902,  .902}\textbf{4.10} & \cellcolor[rgb]{ .906,  .902,  .902}4.40  & \cellcolor[rgb]{ .906,  .902,  .902}\textbf{6.09} & \cellcolor[rgb]{ .906,  .902,  .902}\textbf{5.97} & \cellcolor[rgb]{ .906,  .902,  .902}\textbf{6.12} & \cellcolor[rgb]{ .906,  .902,  .902}\textbf{5.28} & \cellcolor[rgb]{ .906,  .902,  .902}\textbf{4.30} \\
    \midrule
    \multirow{3}[2]{*}{Llama-3.1-8B-SFT-TaP} & \href{https://github.com/HIT-SCIR/huozi/tree/main/data/huozi-rlhf}{Huozi-RLHF} \citep{huozi}  & \textbf{3.50} & \textbf{3.73} & \textbf{3.61} & 4.28  & \textbf{3.66} & 4.53  & 5.36  & 5.50  & 5.81  & 4.86  & 4.24  \\
          & \href{https://huggingface.co/datasets/wenbopan/Chinese-dpo-pairs}{Chinese-DPO-Pairs} & 3.13  & 3.47  & 3.30  & 4.47  & 3.48  & \textbf{4.69} & 5.55  & 5.55  & 5.59  & 4.89  & 4.09  \\
          & \cellcolor[rgb]{ .906,  .902,  .902}TaP (GPT-4) & \cellcolor[rgb]{ .906,  .902,  .902}3.17  & \cellcolor[rgb]{ .906,  .902,  .902}3.68  & \cellcolor[rgb]{ .906,  .902,  .902}3.43  & \cellcolor[rgb]{ .906,  .902,  .902}\textbf{4.52} & \cellcolor[rgb]{ .906,  .902,  .902}3.31  & \cellcolor[rgb]{ .906,  .902,  .902}4.63  & \cellcolor[rgb]{ .906,  .902,  .902}\textbf{5.60} & \cellcolor[rgb]{ .906,  .902,  .902}\textbf{5.66} & \cellcolor[rgb]{ .906,  .902,  .902}\textbf{5.90} & \cellcolor[rgb]{ .906,  .902,  .902}\textbf{4.94} & \cellcolor[rgb]{ .906,  .902,  .902}\textbf{4.18} \\
    \midrule
    \multirow{3}[2]{*}{Gemma-2-9B-SFT-Open} & \href{https://github.com/HIT-SCIR/huozi/tree/main/data/huozi-rlhf}{Huozi-RLHF} \citep{huozi}  & 3.98  & 3.63  & 3.81  & 4.98  & 4.00  & 4.99  & 5.25  & 5.92  & 5.71  & 5.14  & 4.47  \\
          & \href{https://huggingface.co/datasets/wenbopan/Chinese-dpo-pairs}{Chinese-DPO-Pairs} & 4.06  & 3.39  & 3.73  & 4.84  & \textbf{4.14} & 4.87  & 5.88  & 5.68  & 6.09  & 5.25  & 4.49  \\
          & \cellcolor[rgb]{ .906,  .902,  .902}TaP (GPT-4) & \cellcolor[rgb]{ .906,  .902,  .902}\textbf{4.26} & \cellcolor[rgb]{ .906,  .902,  .902}\textbf{3.65} & \cellcolor[rgb]{ .906,  .902,  .902}\textbf{3.96} & \cellcolor[rgb]{ .906,  .902,  .902}\textbf{5.59} & \cellcolor[rgb]{ .906,  .902,  .902}3.84  & \cellcolor[rgb]{ .906,  .902,  .902}\textbf{5.37} & \cellcolor[rgb]{ .906,  .902,  .902}\textbf{6.16} & \cellcolor[rgb]{ .906,  .902,  .902}\textbf{6.08} & \cellcolor[rgb]{ .906,  .902,  .902}\textbf{6.24} & \cellcolor[rgb]{ .906,  .902,  .902}\textbf{5.55} & \cellcolor[rgb]{ .906,  .902,  .902}\textbf{4.75} \\
    \midrule
    \multirow{3}[2]{*}{Gemma-2-9B-SFT-TaP} & \href{https://github.com/HIT-SCIR/huozi/tree/main/data/huozi-rlhf}{Huozi-RLHF} \citep{huozi}  & 3.62  & 3.50  & 3.56  & 4.69  & \textbf{3.83} & \textbf{4.96} & 5.17  & 5.50  & 5.61  & 4.96  & 4.26  \\
          & \href{https://huggingface.co/datasets/wenbopan/Chinese-dpo-pairs}{Chinese-DPO-Pairs} & 3.82  & 3.28  & 3.55  & 4.69  & 3.28  & 4.93  & 5.40  & 5.42  & 5.73  & 4.91  & 4.23  \\
          & \cellcolor[rgb]{ .906,  .902,  .902}TaP (GPT-4) & \cellcolor[rgb]{ .906,  .902,  .902}\textbf{4.31} & \cellcolor[rgb]{ .906,  .902,  .902}\textbf{3.51} & \cellcolor[rgb]{ .906,  .902,  .902}\textbf{3.91} & \cellcolor[rgb]{ .906,  .902,  .902}\textbf{4.98} & \cellcolor[rgb]{ .906,  .902,  .902}3.67  & \cellcolor[rgb]{ .906,  .902,  .902}\textbf{4.96} & \cellcolor[rgb]{ .906,  .902,  .902}\textbf{5.80} & \cellcolor[rgb]{ .906,  .902,  .902}\textbf{5.92} & \cellcolor[rgb]{ .906,  .902,  .902}\textbf{6.07} & \cellcolor[rgb]{ .906,  .902,  .902}\textbf{5.23} & \cellcolor[rgb]{ .906,  .902,  .902}\textbf{4.57} \\
    \midrule
    \multirow{3}[2]{*}{Qwen2.5-14B-SFT-Open} & \href{https://github.com/HIT-SCIR/huozi/tree/main/data/huozi-rlhf}{Huozi-RLHF} \citep{huozi}  & 6.53  & 4.86  & 5.69  & 5.83  & 5.71  & 6.25  & 5.64  & 6.03  & 6.10  & 5.93  & 5.81  \\
          & \href{https://huggingface.co/datasets/wenbopan/Chinese-dpo-pairs}{Chinese-DPO-Pairs} & 6.35  & \textbf{5.35} & 5.85  & \textbf{6.22} & 5.62  & \textbf{6.50} & 6.08  & \textbf{6.32} & 6.29  & \textbf{6.17} & 6.01  \\
          & \cellcolor[rgb]{ .906,  .902,  .902}TaP (GPT-4) & \cellcolor[rgb]{ .906,  .902,  .902}\textbf{6.73} & \cellcolor[rgb]{ .906,  .902,  .902}5.32  & \cellcolor[rgb]{ .906,  .902,  .902}\textbf{6.02} & \cellcolor[rgb]{ .906,  .902,  .902}6.13  & \cellcolor[rgb]{ .906,  .902,  .902}\textbf{5.79} & \cellcolor[rgb]{ .906,  .902,  .902}6.00  & \cellcolor[rgb]{ .906,  .902,  .902}\textbf{6.28} & \cellcolor[rgb]{ .906,  .902,  .902}\textbf{6.32} & \cellcolor[rgb]{ .906,  .902,  .902}\textbf{6.43} & \cellcolor[rgb]{ .906,  .902,  .902}6.16  & \cellcolor[rgb]{ .906,  .902,  .902}\textbf{6.09} \\
    \midrule
    \multirow{3}[2]{*}{Qwen2.5-14B-SFT-TaP} & \href{https://github.com/HIT-SCIR/huozi/tree/main/data/huozi-rlhf}{Huozi-RLHF} \citep{huozi}  & 6.94  & 5.90  & 6.42  & 6.55  & 5.88  & 6.58  & 6.17  & 6.34  & 6.49  & 6.34  & 6.38  \\
          & \href{https://huggingface.co/datasets/wenbopan/Chinese-dpo-pairs}{Chinese-DPO-Pairs} & 7.23  & 5.73  & 6.48  & \textbf{6.68} & \textbf{6.14} & \textbf{6.91} & 6.57  & \textbf{6.50} & 6.72  & \textbf{6.59} & 6.53  \\
          & \cellcolor[rgb]{ .906,  .902,  .902}TaP (GPT-4) & \cellcolor[rgb]{ .906,  .902,  .902}\textbf{7.38} & \cellcolor[rgb]{ .906,  .902,  .902}\textbf{6.29} & \cellcolor[rgb]{ .906,  .902,  .902}\textbf{6.84} & \cellcolor[rgb]{ .906,  .902,  .902}6.67  & \cellcolor[rgb]{ .906,  .902,  .902}6.05  & \cellcolor[rgb]{ .906,  .902,  .902}6.35  & \cellcolor[rgb]{ .906,  .902,  .902}\textbf{6.64} & \cellcolor[rgb]{ .906,  .902,  .902}6.45  & \cellcolor[rgb]{ .906,  .902,  .902}\textbf{6.81} & \cellcolor[rgb]{ .906,  .902,  .902}6.50  & \cellcolor[rgb]{ .906,  .902,  .902}\textbf{6.67} \\
    \bottomrule
    \end{tabular}%
  \caption{Performance comparison of LLMs trained with \textbf{PPO} using different datasets on AlignBench. The model names include two possible suffixes: ``Open'' and ``TaP.'' The ``Open'' suffix indicates that the LLMs were initialized from models trained via supervised fine-tuning on open-source datasets, whereas ``TaP'' denotes initialization from models trained on a dataset constructed by TaP. The evaluation is conducted using \textbf{DeepSeek-V3}, which scores the models’ responses. The ``Fund.'' column denotes Fundamental Language Ability, ``Chi.'' denotes Advanced Chinese Understanding, ``Open.'' denotes Open-ended Questions, ``Writ.'' denotes Writing Ability, ``Role.'' denotes Task-oriented Role Play, ``Pro'' denotes ``Professional Knowledge'', ``Math.'' denotes Mathematics, and ``Logic.'' denotes Logical Reasoning.}
  \label{tab:ppo_alignbench_deepseek}%
\end{table*}%

\end{document}